\documentclass[oneside,10pt]{article}

\usepackage{times,url,mathrsfs,algorithm}
\usepackage{amsmath}
\usepackage{amsfonts}
\usepackage{graphicx}
\usepackage{subfigure}

\newtheorem{lemma}{Lemma}

\begin{document}

\title{Adaptive Base Class Boost for Multi-class Classification}

\author{ Ping Li \\
       Department of Statistical Science\\
       Faculty of Computing and Information Science \\
       Cornell University\\
       Ithaca, NY 14853\\
       pingli@cornell.edu\\\\
       First draft July 2008. \ \  Revised October 2008}
\date{}

\maketitle

\begin{abstract}
We\footnote{The first draft  was submitted to IEEE ICDM on July 07, 2008. Although the submission was not accepted, the author appreciates two of the reviewers for their informative comments. In particular, the author appreciates that one of the reviewer considered this work ''will be one of the state of the art.'' In this revision, we followed that reviewer's suggestion by adding more experiments.    } develop the concept of \textbf{ABC}-Boost (\textbf{A}daptive \textbf{B}ase \textbf{C}lass Boost) for  multi-class classification and  present  \textbf{ABC}-MART, a concrete implementation of \textbf{ABC}-Boost.  The original MART (\textbf{M}ultiple \textbf{A}dditive \textbf{R}egression \textbf{T}rees) algorithm has been very successful in large-scale applications.  For binary classification, ABC-MART recovers MART. For multi-class  classification,  ABC-MART considerably improves MART, as  evaluated on several public data sets.
\end{abstract}

\section{Introduction}

Classification is a basic task in machine learning. A training data set $\{y_i,\mathbf{X}_i\}_{i=1}^N$ consists of $N$ feature vectors (samples) $\mathbf{X}_i$, and $N$ class labels, $y_i$, $i = 1$ to $N$. Here $y_i \in \{0, 1, 2, ..., K-1\}$ and $K$ is the number of classes. The  task is to predict the class labels. This study focuses on multi-class classification ($K\geq 3$).

Many classification algorithms are based on \textbf{boosting}\cite{Article:Schapire_ML90,Article:Freund_95,Article:Freund_JCSS97,Article:FHT_AS00,Article:Friedman_AS01}, which is  regarded one of most significant breakthroughs in machine learning.  MART\cite{Article:Friedman_AS01} (\textbf{M}ultiple \textbf{A}dditive \textbf{R}egression \textbf{T}rees) is a successful boosting algorithm, especially for large-scale applications in industry practice. For example, the regression-based ranking method developed in Yahoo!\cite{Proc:Zhang_COLT06,Proc:GBRank_SIGIR07} used an underlying learning algorithm based on MART. {\em McRank}\cite{Proc:McRank_NIPS07}, the classification-based ranking method, also used MART as the underlying learning procedure.

This study proposes \textbf{ABC}-Boost (\textbf{A}daptive \textbf{B}ase \textbf{C}lass Boost) for  multi-class classification. We present  \textbf{ABC}-MART, a concrete implementation of {ABC}-Boost.
ABC-Boost  is based on the following two key ideas:
\begin{enumerate}
\item For multi-class classification, popular loss functions for $K$ classes usually assume a constraint\cite{Article:FHT_AS00,Article:Friedman_AS01,Article:Zhang_JMLR04}  such that only the values for $K-1$ classes are needed. Therefore, we can choose a \textbf{base class} and derive  algorithms only for  $K-1$ classes.
\item At each boosting step, although the base class is not explicitly trained, it will implicitly benefit from  the training on $K-1$ classes, due to the constraint.  Thus, we \textbf{adaptively} choose the base class which  has the ``worst'' performance.
\end{enumerate}

The idea of assuming a constraint on the loss function and  using a base class  may not be at all surprising. For binary ($K=2$) classification, a ``sum-to-zero'' constraint on the loss function is automatically considered so that we only need to train the algorithm for one (instead of $K=2$) class. For multi-class ($K\geq3$) classification, the sum-to-zero constraint on the loss function is also ubiquitously adopted\cite{Article:FHT_AS00,Article:Friedman_AS01,Article:Zhang_JMLR04}. In particular, the multi-class {\em Logitboost}\cite{Article:FHT_AS00} algorithm was derived by explicitly averaging over $K-1$ base classes.

The loss function adopted in our ABC-MART is the same as in MART\cite{Article:Friedman_AS01} and {\em Logitboost}\cite{Article:FHT_AS00}. All three algorithms assume the ``sum-to-zero'' constraint. However, we obtain different first and second derivatives of the loss function, from MART\cite{Article:Friedman_AS01} and {\em Logitboost}\cite{Article:FHT_AS00}. See Section \ref{sec_ABC} for details.

In terms of implementation, our proposed ABC-MART differs from the original MART algorithm only in a few lines of code. Since MART is known to be a successful algorithm, much of our work is devoted to the empirical comparisons of ABC-MART with MART. Our experiment results on publicly available data sets will demonstrate that ABC-MART could considerably improves MART. Also, ABC-MART reduces both the training and testing time by $1/K$, which may be quite beneficial when $K$ is small.

We notice that data sets in industry applications are often quite large (e.g., several million samples\cite{Proc:McRank_NIPS07}). Publicly available data sets (e.g., UCI repository), however, are mostly small. In our study, the {\em Covertype} data set from the UCI repository is reasonably large with 581,012 observations. \\

We first review the original MART algorithm and functional gradient boosting\cite{Article:Friedman_AS01}.

\section{Review MART and Functional Gradient Boosting}\label{sec_review}

The MART algorithm is the marriage of the functional gradient boosting and regression trees. Given the training data set $\{y_i, \mathbf{x}_{i}\}_{i=1}^N$ and a loss function $L$, \cite{Article:Friedman_AS01} adopted a ``greedy stagewise'' approach to build an additive function $F^{(M)}$, which is a sum of $M$ terms,
\begin{align}\label{eqn_F_M}
F^{(M)}(\mathbf{x}) = \sum_{m=1}^M \rho_m h(\mathbf{x};\mathbf{a}_m),
\end{align}
such that, at each stage $m$, $m = 1$ to $M$,
\begin{align}\label{eqn_F_m_argmin}
\left\{\rho_m, \mathbf{a}_{m}\right\} =  \underset{\rho,\mathbf{a}}{\text{argmin}} \sum_{i=1}^N L\left(y_i, F^{(m-1)}(\mathbf{x}_i;\mathbf{a},\rho)\right).
\end{align}
Here the function $h(\mathbf{x};\mathbf{a})$ is the ``base learner'' or ``weak learner.'' In general, (\ref{eqn_F_m_argmin}) is still a difficult optimization problem. \cite{Article:Friedman_AS01} approximately conducted steepest descent in the function space, by solving a least square problem
\begin{align}
\mathbf{a}_m =  \underset{\mathbf{a},\rho}{\text{argmin}} \sum_{i=1}^N \left[ -g_m(\mathbf{x}_i) - \rho h(\mathbf{x}_i;\mathbf{a})  \right]^2,
\end{align}
where
 \begin{align}
-g_{m}(\mathbf{x}_i) =- \left[ \frac{\partial L\left(y_i, F(\mathbf{x}_i)\right)}{\partial F(\mathbf{x}_i)}\right]_{F(\mathbf(x)) = F^{(m-1)}(\mathbf(x))}
\end{align}
is the negative gradient (the steepest descent direction) in the $N$-dimensional data space at $F^{(m-1)}(\mathbf{x})$. For obtaining another coefficient $\rho_m$, a line search is performed:
\begin{align}
\rho_m = \underset{\rho}{\text{argmin}}\sum_{i=1}^NL\left(y_i, F^{(m-1)}(\mathbf{x}_i) + \rho h(\mathbf{x}_i;\mathbf{a}_m)\right).
\end{align}

A generic ``gradient boosting'' algorithm is described in Alg. \ref{alg_GB}, for any differentiable loss function $L$. For multi-class classification, \cite{Article:Friedman_AS01} proposed MART, which implemented Line 5 in Alg. \ref{alg_GB} by regression trees and Line 7 by a one-step Newton update within each terminal node of the trees.

{\small\begin{algorithm}
1: $F_{\mathbf{x}} = \text{argmin}_\rho \sum_{i=1}^N L(y_i,\rho)$ \\
2: For $m=1$ to $M$ Do\\
3: \hspace{0.2in}    For $k=0$ to $K-1$ Do\\
4: \hspace{0.4in}  $\tilde{y}_i = - \left[ \frac{\partial L\left(y_i, F(\mathbf{x}_i)\right)}{\partial F(\mathbf{x}_i)}\right]_{F(\mathbf(x)) = F^{(m-1)}(\mathbf(x)))}$,  $i = 1$ to $N$. \\
5:  \hspace{0.4in}  $\mathbf{a}_{m} = \text{argmin}_{\mathbf{a},\rho}\sum_{i=1}^N\left[\tilde{y}_i - \rho h(\mathbf{x}_i;\mathbf{a}) \right]^2$ \\
6:   \hspace{0.4in}  $\rho_m = \text{argmin}_\rho\sum_{i=1}^NL\left(y_i, F^{(m-1)}(\mathbf{x}_i) + \rho h(\mathbf{x}_i;\mathbf{a}_m)\right)$  \\
7:  \hspace{0.4in}  $F_{\mathbf{x}} = F_{\mathbf{x}} + \rho_m h(\mathbf{x};\mathbf{a}_m)$\\
8:   \hspace{0.2in} End\\
9: End
\caption{A generic gradient boosting algorithm \cite[Alg. 1]{Article:Friedman_AS01}. }
\label{alg_GB}
\end{algorithm}}

\subsection{The Loss Function and Multiple Logistic Probability Model in MART}

For multi-class classification, MART adopted the following {\em negative multinomial log-likelihood loss}, which is also the loss function in {\em Logitboost}\cite{Article:FHT_AS00}:
\begin{align}\label{eqn_loss}
L = \sum_{i=1}^N L_i = \sum_{i=1}^N \left\{ - \sum_{k=0}^{K-1}r_{i,k}  \log p_{i,k}\right\}
\end{align}
where $r_{i,k} = 1$ if $y_i = k$ and $r_{i,k} = 0$ otherwise. Apparently, $\sum_{k=0}^{K-1} r_{i,k} = 1$ for any $i$.  Here $p_{i,k}$ is the probability that the $i$th observation belongs to class $k$:
\begin{align}
p_{i,k} = \mathbf{Pr}\left( y _i = k | \mathbf{X}_i\right).
\end{align}

MART adopted the following  logistic probability model\cite{Article:Friedman_AS01}
\begin{align}\label{eqn_logistic_F}
F_{i,k} = \log p_{i,k} - \frac{1}{K} \sum_{s=0}^{K-1} \log p_{i,s},
\end{align}
or equivalently\cite{Article:Friedman_AS01},
\begin{align}\label{eqn_logistic_p}
p_{i,k} = \frac{e^{F_{i,k}}}{\sum_{s=0}^{K-1} e^{F_{i,s}}}.
\end{align}


Apparently, the model (\ref{eqn_logistic_F}) implies  $\sum_{k=0}^{K-1} F_{i,k} = 0$, the sum-to-zero constraint. In fact, since $\sum_{k=0}^{K-1}p_{i,k} = 1$, the model only has $K-1$ degrees of freedom. Some constraint on $F_{i,k}$ is necessary in order to obtain a unique solution. For binary ($K=2$) classification, the sum-to-zero constraint is automatically enforced.


\subsection{The Original MART Algorithm}

Alg. \ref{alg_mart} describes the MART algorithm for multi-class classification using  {\em negative multinomial log-likelihood loss} (\ref{eqn_loss}) and multi-class logistic model (\ref{eqn_logistic_p}).

{\small\begin{algorithm}
0: $r_{i,k} = 1$, if $y_{i} = k$, and $r_{i,k} =
0$ otherwise. \\
1: $F_{i,k} = 0$, $k = 0$ to  $K-1$, $i = 1$ to $N$ \\
2: For $m=1$ to $M$ Do\\
3: \hspace{0.2in}    For $k=0$ to $K-1$ Do\\
4: \hspace{0.4in}  $p_{i,k} = \exp(F_{i,k})/\sum_{s=0}^{K-1}
\exp(F_{i,s})$\\
5:  \hspace{0.4in}  $\left\{R_{j,k,m}\right\}_{j=1}^J = J$-terminal node regression tree from $\{r_{i,k} - p_{i,k}, \ \ \mathbf{x}_{i}\}_{i=1}^N$ \\
6:   \hspace{0.4in}  $\beta_{j,k,m} = \frac{K-1}{K}\frac{ \sum_{\mathbf{x}_i \in
  R_{j,k,m}} r_{i,k} - p_{i,k}}{ \sum_{\mathbf{x}_i\in
  R_{j,k,m}}\left(1-p_{i,k}\right)p_{i,k} }$ \\
7:  \hspace{0.4in}  $F_{i,k} = F_{i,k} +
\nu\sum_{j=1}^J\beta_{j,k,m}1_{\mathbf{x}_i\in R_{j,k,m}}$ \\
8:   \hspace{0.2in} End\\
9: End
\caption{MART\cite[Alg. 6]{Article:Friedman_AS01}.  Note that in Line 6, the term {\small$p_{i,k}(1-p_{i,k})$} replaces the equivalent term {\small$|r_{i,k} - p_{i,k}|(1-|r_{i,k}-p_{i,k}|)$} in \cite[Alg. 6]{Article:Friedman_AS01}}.
\label{alg_mart}
\end{algorithm}}

MART follows the generic paradigm of functional gradient boosting Alg. \ref{alg_GB}. At each stage $m$, MART solves the mean square problem (Line 5 in Alg. \ref{alg_GB}) by regression trees.  MART builds $K$ regression trees at each boosting step.\\

We elaborate in more detail on serval key components of MART.

\subsubsection{The Functional Gradient (Pseudo Response)}

MART performs gradient descent in the function space, using the gradient evaluated at the function values. For the $i$th data point, using the {\em negative multinomial log-likelihood loss} (\ref{eqn_loss}), i.e.,
\begin{align}\label{eqn_Li}
L_i\left(F_{i,k}, y_i\right) = -\sum_{k=0}^{K-1} r_{i,k} \log p_{i,k}
\end{align}
and the probability model (\ref{eqn_logistic_p}), \cite{Article:Friedman_AS01} showed
\begin{align}\label{eqn_MART_d1}
\frac{\partial L_i}{\partial F_{i,k}} = - \left(r_{i,k} - p_{i,k}\right).
\end{align}
This explains the term $r_{i,k} - p_{i,k}$ in Line 5 of Alg. \ref{alg_mart}.

\subsubsection{The Second Derivative and One-Step Newton Update}

While only the first derivatives were used for building the structure of the trees, MART used the second derivatives to update the values of the terminal nodes by a one-step Newton procedure. \cite{Article:Friedman_AS01} showed
\begin{align}\label{eqn_MART_d2}
\frac{\partial^2 L_i}{\partial F_{i,k}^2} = p_{i,k}\left(1-p_{i,k}\right),
\end{align}
which explains Line 6 in Alg. \ref{alg_mart}.\\

(\ref{eqn_MART_d1}) and (\ref{eqn_MART_d2}) were also derived in the {\em Logitboost}\cite{Article:FHT_AS00}. However, in this paper we actually obtain different first and second derivatives.

\subsubsection{$K$ Trees for $K$ Classes}

For each $i$, there are $K$ function values, $F_{i,k}$, $k = 0$ to $K-1$, and consequently $K$ gradients.  MART builds $K$ regression tress at each boosting step. Apparently, the constraint $\sum_{k=0}^{K-1} F_{i,k} = 0$  will not hold.

Note that one can actually re-center the $F_{i,k}$ at the end of every boosting step so that the sum-to-zero constraint is satisfied after training. That is, one can insert a line
\begin{align}
F_{i,k} \leftarrow F_{i,k} - \frac{1}{K}\sum_{s=0}^{K-1} F_{i,s},
\end{align}
after Line 8 in Alg. \ref{alg_mart} to make $\sum_{k=0}^{K-1} F_{i,k} = 0$. However, we observe that this re-centering step makes no difference in our experiments. We believe the sum-to-zero constraint should be enforced before the training  instead of after the training, at every boosting step.

The case $K=2$ is an exception. Because the two pseudo responses, $r_{i,k=0} - p_{i,k=0}$ and $r_{i,k=1} - p_{i,k=1}$ are identical with signs flipped (i.e., $\sum_{k=0}^1 F_{i,k} = 0$ automatically holds), there is no need to build $K=2$ tress. In fact, \cite{Article:Friedman_AS01} presented the binary classification algorithm separately \cite[Alg. 5]{Article:Friedman_AS01}, which is the same as \cite[Alg. 6]{Article:Friedman_AS01} by letting $K=2$, although the presentations were somewhat different.

Line 6 of Alg. \ref{alg_mart} contains a factor $\frac{K-1}{K}$. For binary classification, it is clear that the factor $\frac{1}{2}$ comes from the mathematical derivation. For $K\geq3$, we believe this factor in a way approximates the constraint $\sum_{k=0}^{K-1} F_{i,k} = 0$. Because MART builds $K$ trees while there are only $K-1$ degrees of freedom, a factor $\frac{K-1}{K}$ may help reduce the influence.

\subsubsection{Three Main Parameters: $J$, $\nu$, and $M$}

Practitioners like MART partly because this great algorithm has only a few parameters, which are not very sensitive as long as they fall in some ``reasonable'' range. It is often fairly easy to identify the (close to) optimal parameters with limited tuning. This is a huge advantage, especially for large data sets.

The number of terminal nodes, $J$,  determines the capacity of the base learner. MART suggested $J=6$ often might be  a good choice.

The shrinkage parameter, $\nu$, should be large enough to make a sufficient progress at each boosting step and  small enough to avoid over-fitting. Also, a very small $\nu$ may require a large number of boosting steps, which may be a practical concern for real-world applications if the training and/or testing is time-consuming.  \cite{Article:Friedman_AS01} suggested $\nu\leq 0.1$.

The number of boosting steps, $M$, in a sense is largely determined  by the computing time one can afford. A commonly-regarded merit of boosting is that over-fitting can be largely avoided for reasonable $J$ and $\nu$ and hence one might simply let $M$ be as large as possible. However, for small data sets, the training loss (\ref{eqn_loss}) may reach the machine accuracy before $M$ can be too large.

\section{ABC-Boost and ABC-MART}\label{sec_ABC}

We re-iterate the two key components in developing  ABC-Boost (Adaptive Base Class Boost):
\begin{enumerate}
\item Using the popular constraint in the loss function, we can choose a \textbf{base class} and derive the boosting algorithm  for only $K-1$ classes.
\item At each boosting step, we can \textbf{adaptively} choose the base class which has the ``worst'' performance.
\end{enumerate}

ABC-MART is a concrete implementation of ABC-Boost by using the {\em negative multinomial log-likelihood loss} (\ref{eqn_loss}), the  multi-class logistic model (\ref{eqn_logistic_p}), and the paradigm of functional gradient boosting.  Apparently, there are other possible implementations of ABC-Boost. For example, one can  implement an ``ABC-Logitboost.'' This study focuses on ABC-MART. Since MART has been proved to be successful in large-scale industry applications, demonstrating that ABC-MART may considerably improve MART will be appealing.

\subsection{The Multi-class Logistic Model with a Fixed  Base}

We first derive some basic formulas needed for developing ABC-MART. Without loss of generality, we assume class 0 is the base class.

Lemma \ref{lem_derivative_p} provides the first and second derivatives of the class probabilities $p_{i,k}$ under the multi-class logistic model (\ref{eqn_logistic_p}) and the sum-to-zero constraint $\sum_{k=0}^{K-1} F_{i,k} = 0$\cite{Article:FHT_AS00,Article:Friedman_AS01,Article:Zhang_JMLR04}.
\begin{lemma}\label{lem_derivative_p}
\begin{align}
&\frac{\partial p_{i,k}}{\partial F_{i,k}} = p_{i,k}\left(1+p_{i,0} - p_{i,k}\right), \ \ k \neq 0\\
&\frac{\partial p_{i,k}}{\partial F_{i,s}} = p_{i,k}\left(p_{i,0} - p_{i,s}\right), \ \ k \neq s\neq 0\\
&\frac{\partial p_{i,0}}{\partial F_{i,k}} = p_{i,0}\left(-1+p_{i,0} - p_{i,k}\right), \ \ k \neq 0
\end{align}
\textbf{Proof:}
Note that $F_{i,0} = -\sum_{k=1}^{K-1} F_{i,k}$. Hence
\begin{align}\notag
&p_{i,k} = \frac{e^{F_{i,k}}}{\sum_{s=0}^{K-1} e^{F_{i,s}}} = \frac{e^{F_{i,k}}}{\sum_{s=1}^{K-1} e^{F_{i,s}} + e^{\sum_{s=1}^{K-1}-F_{i,s}}}   \\\notag
&\frac{\partial p_{i,k}}{\partial F_{i,k}}  = \frac{e^{F_{i,k}}}{\sum_{s=0}^{K-1} e^{F_{i,s}}} - \frac{e^{F_{i,k}}\left(e^{F_{i,k}} - e^{-F_{i,0}}\right) }{ \left(\sum_{s=0}^{K-1} e^{F_{i,s}}\right)^2}\\\notag
&\hspace{0.4in} = p_{i,k} \left(1+p_{i,0} - p_{i,k}\right).
\end{align}
The other derivatives can be obtained similarly. $\Box$
\end{lemma}

Lemma \ref{lem_derivative_p} helps derive the derivatives of the loss function, presented in Lemma \ref{lem_derivative_L}.
\begin{lemma}\label{lem_derivative_L}
\begin{align}
&\frac{\partial L_i}{\partial F_{i,k}}  = \left(r_{i,0} - p_{i,0}\right) - \left(r_{i,k} - p_{i,k}\right),\\
&\frac{\partial^2 L_i}{\partial F_{i,k}^2} = p_{i,0}(1-p_{i,0}) + p_{i,k}(1-p_{i,k}) + 2p_{i,0}p_{i,k}.
\end{align}
\textbf{Proof:}\vspace{-0.1in}
\begin{align}\notag
&L_i = -\sum_{s=1,s\neq k}^{K-1} r_{i,s} \log p_{i,s} - r_{i,k}\log p_{i,k} - r_{i,0} \log p_{i,0}.
\end{align}
It first derivative is
\begin{align}
&\frac{\partial L_i}{\partial F_{i,k}} = - \sum_{s=1,s\neq k}^{K-1} \frac{r_{i,s}}{p_{i,s}} \frac{\partial p_{i,s}}{F_{i,k}} -  \frac{r_{i,k}}{p_{i,k}} \frac{\partial p_{i,k}}{F_{i,k}}- \frac{r_{i,0}}{p_{i,0}} \frac{\partial p_{i,0}}{F_{i,k}}\\\notag
=& \sum_{s=1,s\neq k}^{K-1}- r_{i,s}\left(p_{i,0} - p_{i,k}\right)- r_{i,k}\left(1+p_{i,0} - p_{i,k}\right)- r_{i,0}\left(-1+p_{i,0} - p_{i,k}\right)\\\notag
=&- \sum_{s=0}^{K-1} r_{i,s}\left(p_{i,0} - p_{i,k}\right)+r_{i,0}-r_{i,k}\\\notag
=&\left(r_{i,0} - p_{i,0}\right) - \left(r_{i,k} - p_{i,k}\right).
\end{align}

And the second derivative is
\begin{align}\notag
&\frac{\partial^2 L_i}{\partial F_{i,k}^2} = -\frac{\partial p_{i,0}}{\partial F_{i,k}} + \frac{\partial p_{i,k}}{\partial F_{i,k}}\\\notag
=&-p_{i,0}\left(-1+p_{i,0} - p_{i,k}\right)+p_{i,k}\left(1+p_{i,0} - p_{i,k}\right)\\\notag
=&p_{i,0}(1-p_{i,0}) + p_{i,k}(1-p_{i,k}) + 2p_{i,0}p_{i,k}. \ \Box
\end{align}\\
\end{lemma}

Note the the first and second derivatives we derives differ from (\ref{eqn_MART_d1}) and (\ref{eqn_MART_d2}), which are the derivatives used in {\em Logitboost}\cite{Article:FHT_AS00} and MART\cite{Article:Friedman_AS01}.

\subsection{ABC-MART}

Alg. \ref{alg_ABC-MART} provides the pseudo code for ABC-MART. Compared with  MART (Alg. \ref{alg_mart}), we use different gradients to build the trees and different second derivatives to update the values of terminal nodes. In addition, there is a procedure (Line 10) for selecting the base class, denoted by $b$, which has the ``worst'' (i.e., largest) training loss (\ref{eqn_loss}).   At each boosting step, we only need to build $K-1$ trees because the constraint $\sum_{k=0}^{K-1} F_{i,k}=0$ is enforced.

{\begin{algorithm}[h]
0: $r_{i,k} = 1$, if $y_{i} = k$, $r_{i,k} =
0$ otherwise. Choose a random base $b$\\
1: $F_{i,k} = 0$,\ \  $p_{i,k} = \frac{1}{K}$, \ \ \ $k = 0$ to  $K-1$, \ $i = 1$ to $N$ \\
2: For $m=1$ to $M$ Do\\
3: \hspace{0.2in}    For $k=0$ to $K-1$, $k\neq b$, Do\\
4:  \hspace{0.4in}  $\left\{R_{j,k,m}\right\}_{j=1}^J = J$-terminal
node regression tree from  $\{-(r_{i,b} - p_{i,b}) +  (r_{i,k} - p_{i,k}), \ \ \mathbf{x}_{i}\}_{i=1}^N$ \\\\
5:   \hspace{0.3in}  $\beta_{j,k,m} = \frac{ \sum_{\mathbf{x}_i \in
  R_{j,k,m}} -(r_{i,b} - p_{i,b}) + (r_{i,k} - p_{i,k})  }{ \sum_{\mathbf{x}_i\in
  R_{j,k,m}} p_{i,b}(1-p_{i,b})+ p_{i,k}\left(1-p_{i,k}\right) + 2p_{i,b}p_{i,k} }$ \\\\
6:  \hspace{0.4in}  $F_{i,k} = F_{i,k} +
\nu\sum_{j=1}^J\beta_{j,k,m}1_{\mathbf{x}_i\in R_{j,k,m}}$ \\
7:   \hspace{0.2in} End\\
8: \hspace{0.2in} $F_{i,b} = - \sum_{k\neq b} F_{i,k}$, \  $i = 1$ to $N$. \\
9: \hspace{0.2in}  $p_{i,k} = \exp(F_{i,k})/\sum_{s=0}^{K-1}\exp(F_{i,s})$, \ \ \ $k = 0$ to  $K-1$, \ $i = 1$ to $N$ \\
10: \hspace{0.2in} $b = \underset{k}{\text{argmax}} \  \ L^{(k)}$, \ \ where\  \  $L^{(k)} = \sum_{i=1}^N -\log\left(p_{i,k}\right)1_{y_i = k}, \hspace{0.05in} k = 0, 1, ..., K-1$.\\
11: End
\caption{ABC-MART. }
\label{alg_ABC-MART}
\end{algorithm}}


\subsection{ABC-MART Recovers MART when $K = 2$}

Note that the factor $\frac{K-1}{K}$ does not appear in ABC-MART (Alg. \ref{alg_ABC-MART}). Interestingly, when $K=2$,  ABC-MART recovers MART.

For example, consider $K = 2$, $r_{i,0} = 1$, $r_{i,1} = 0$, then
\begin{align}\notag
&\frac{\partial L_i}{\partial F_{i,1} } = (1-p_{i,0}) - (0-p_{i,1}) = 2p_{i,1},\\\notag
&\frac{\partial^2 L_i}{\partial F_{i,1}^2 } = 4p_{i,0}p_{i,1}.
\end{align}
In other words, the first (second) derivative is twice (four times) of the first (second) derivative in MART. Using the one-step Newton update (Line 6 in Alg. \ref{alg_ABC-MART}), the factor $\frac{1}{2}$ (which appeared in MART) is  recovered. Note that scaling the first derivatives does not affect the tree structures.

\subsection{MART Approximately ``Averages'' All Base Classes when $K\geq3$}

In a sense, MART did consider the averaging affect from all $K-1$ base classes.

For the first derivatives, the following equality holds,
\begin{align}
\sum_{b\neq k}\left\{ -(r_{i,b} - p_{i,b}) + (r_{i,k} - p_{i,k})\right\} = K (r_{i,k} - p_{i,k}),
\end{align}
because
\begin{align}\notag
&\sum_{b\neq k}\left\{ -(r_{i,b} - p_{i,b}) + (r_{i,k} - p_{i,k})\right\}\\\notag
=& -\sum_{b\neq k}r_{i,b} + \sum_{b\neq k}p_{i,b} + (K-1)(r_{i,k} - p_{i,k})\\\notag
=&-1+r_{i,k} + 1-p_{i,k} + (K-1)(r_{i,k} - p_{i,k}) \hspace{0.5in} \text{Recall :} \ \sum_{k=0}^{K-1}r_{i,k} = 1, \ \ \sum_{k=0}^{K-1} p_{i,k} = 1\\\notag
=&K(r_{i,k} - p_{i,k}).
\end{align}
In other words, the gradient used in MART is the averaged gradient of ABC-MART.

Next, we can show that, for the second derivatives,
\begin{align}
\sum_{b\neq k}\left\{ (1 - p_{i,b})p_{i,b} + (1 - p_{i,k})p_{i,k} + 2p_{i,b}p_{i,k}\right\} \approx (K+2) (1 - p_{i,k})p_{i,k},
\end{align}
with equality holding only when $K =2$, because
\begin{align}\notag
&\sum_{b\neq k}\left\{ (1 - p_{i,b})p_{i,b} + (1 - p_{i,k})p_{i,k} + 2p_{i,b}p_{i,k}\right\}\\\notag
=&(K-1)(1 - p_{i,k})p_{i,k} + 2p_{i,k}\sum_{b\neq k} p_{i,b} +  \sum_{b\neq k} (1 - p_{i,b})p_{i,b}\\\notag
=&(K+1)(1 - p_{i,k})p_{i,k}  +  \sum_{b\neq k} p_{i,b} - \sum_{b\neq k}p^2_{i,b}\\\notag
\approx &(K+1)(1 - p_{i,k})p_{i,k}  +  \sum_{b\neq k} p_{i,b} - \left(\sum_{b\neq k} p_{i,b}\right)^2\\\notag
=&(K+1)(1 - p_{i,k})p_{i,k} + (1-p_{i,k})p_{i,k}\\\notag
=& (K+2) (r_{i,k} - p_{i,k}).
\end{align}

Thus, even the second derivative used in MART may be approximately viewed as the averaged second derivatives in ABC-MART.
It appears the factor $\frac{K-1}{K}$ in MART may be reasonably replaced by $\frac{K}{K+2}$ (both equal $\frac{1}{2}$ when $K=2$).
This, of course, will not  make a real difference because the constant (either $\frac{K-1}{K}$ or $\frac{K}{K+2}$)  can be absorbed into the shrinkage factor $\nu$.

\section{Evaluations}\label{sec_evaluations}

The goal of the evaluation study is to compare  ABC-MART (Alg. \ref{alg_ABC-MART}) with MART (Alg. \ref{alg_mart}) for multi-class classification.  The experiments were conducted on one fairly large data set ({\em Covertype}) plus five small data sets ({\em Letter, Pendigits, Zipcode, Optdigits}, and {\em Isolet}); see Table \ref{tab_data}.\footnote{ All data sets are publicly available. The {\em Zipcode} data set is downloaded from \url{http://www-stat.stanford.edu/~tibs/ElemStatLearn/data.html} and all other data sets can be found from the UCI repository.}

\begin{table}[h]
\caption{Our experiments were based on six publicly available data sets. We randomly split the {\em Covertype} data set into a training set and test set. For all other data sets, we used the standard (default) training and test sets.
 }
\begin{center}{
\begin{tabular}{l r r r r}
\hline \hline\\
Data set &$K$ & \# training samples & \# test samples &\# features\\
\hline
Covertype &7 & 290506 & 290506 & 54\\
Letter &26   & 16000   &4000 &16\\
Pendigits &10 &7494   &3498 &16\\
Zipcode  &10  &7291   &2007 &256\\
Optdigits &10  &3823   &1797 &64\\
Isolet &26   & 6218   &1559 &617\\
\hline\hline
\end{tabular}
}
\end{center}
\label{tab_data}
\end{table}

In general, a comprehensive and fair comparison of two classification algorithms is a non-trivial task. In our case, however, the comparison task appears quite easy  because ABC-MART and MART differ only in a few lines of code and their underlying base learners can be completely identical.

Ideally, we hope that ABC-MART will improve MART, for every  set of reasonable parameters, $J$ and $\nu$. For the five small data sets, we experiment with every combination of number of terminal nodes $J$ and shrinkage $\nu$, where $J$ and $\nu$ are chosen from
\begin{align}\notag
J \in  \{4,\ 6,\ 8,\ 10,\ 12,\ 14,\ 16\},\hspace{0.2in}
\nu \in \{0.04,\ 0.06,\ 0.08,\ 0.1\}.
\end{align}

For the five small data sets, we let the number of boosting steps $M=10000$. However, the experiments usually terminated well before $M=10000$ because the machine accuracy was reached.

For the {\em Covertype} data set, since it is fairly large, we only experimented with $J = 6, 10, 20$ and $\nu = 0.04, 0.1$ and we limited $M$ to be 16000, 11500, 6000, for $J= 6, 10, 20$, respectively.

\subsection{Summary of Experiment Results}

The test misclassification error is  a direct measure of performance. MART and ABC-MART output  $K$ class probabilities $p_{i,k}$, $k = 0$ to $K-1$, for each observation $i$. To obtain the class labels, we adopt the commonly used rule
\begin{align}
k^\prime = \underset{k} {\text{argmax}} \  p_{i,k}.
\end{align}

We define $R_{err}$, the ``relative improvement of test mis-classification errors,'' as
\begin{align}
R_{err} = \frac{\text{mis-classification errors of MART} - \text{mis-classification errors of ABC-MART}}
{\text{mis-classification errors of MART}}\times 100 \ (\%).
\end{align}

Since we experimented with a series of parameters, $J$, $\nu$, and $M$, we report the overall ``best'' (i.e., smallest) mis-classification errors in Table \ref{tab_summary}. Later, we will also report the more detailed mis-classification errors for every combination of $J$ and $\nu$, in Sections \ref{sec_Covertype} to \ref{sec_Isolet}.

\begin{table}[h]
\caption{Summary of test mis-classification errors.   }
\begin{center}{
\begin{tabular}{l r r r l}
\hline \hline\\
Data set &MART & ABC-MART & $R_{err}$ (\%)  &$P$-value\\
\hline
Covertype &11133 & 10203 & 8.4 & $4.4\times 10^{-11}$\\
Letter &135   & 111   &17.8 &0.060\\
Pendigits &123 &104   &15.5 &0.100\\
Zipcode  &111  &98   &11.7 &0.178\\
Optdigits &56  &41   &26.8 &0.061\\
Isolet &84   & 69   &17.9 &0.107\\
\hline\hline
\end{tabular}
}
\end{center}
\label{tab_summary}
\end{table}

Table \ref{tab_summary} also provides the ``$P$-value'' of the one-sided $t$-test. The idea is to model the test error rate (i.e., the test mis-classification errors divided by the number of test samples) as a binomial probability and then conduct the $t$-test using the normal approximation of the difference of two binomial probabilities. We can see that for the {\em Covertype} data set, the improvement of ABC-MART over MART is highly significant (the $P$-value is nearly zero)  under this test. For the five small  data sets, the $P$-values are also reasonably small.

We shall mention that this $t$-test is  very stringent when the error rate is small. In fact, we do not often see papers which calculated the $P$-values when comparing different classification algorithms.\\

Next, we present the detailed experiment results on the six data sets.

\subsection{Experiments on the {\em Covertype} Data Set }\label{sec_Covertype}

\begin{figure}[h]
\begin{center}
\mbox{\includegraphics[width=2.0in]{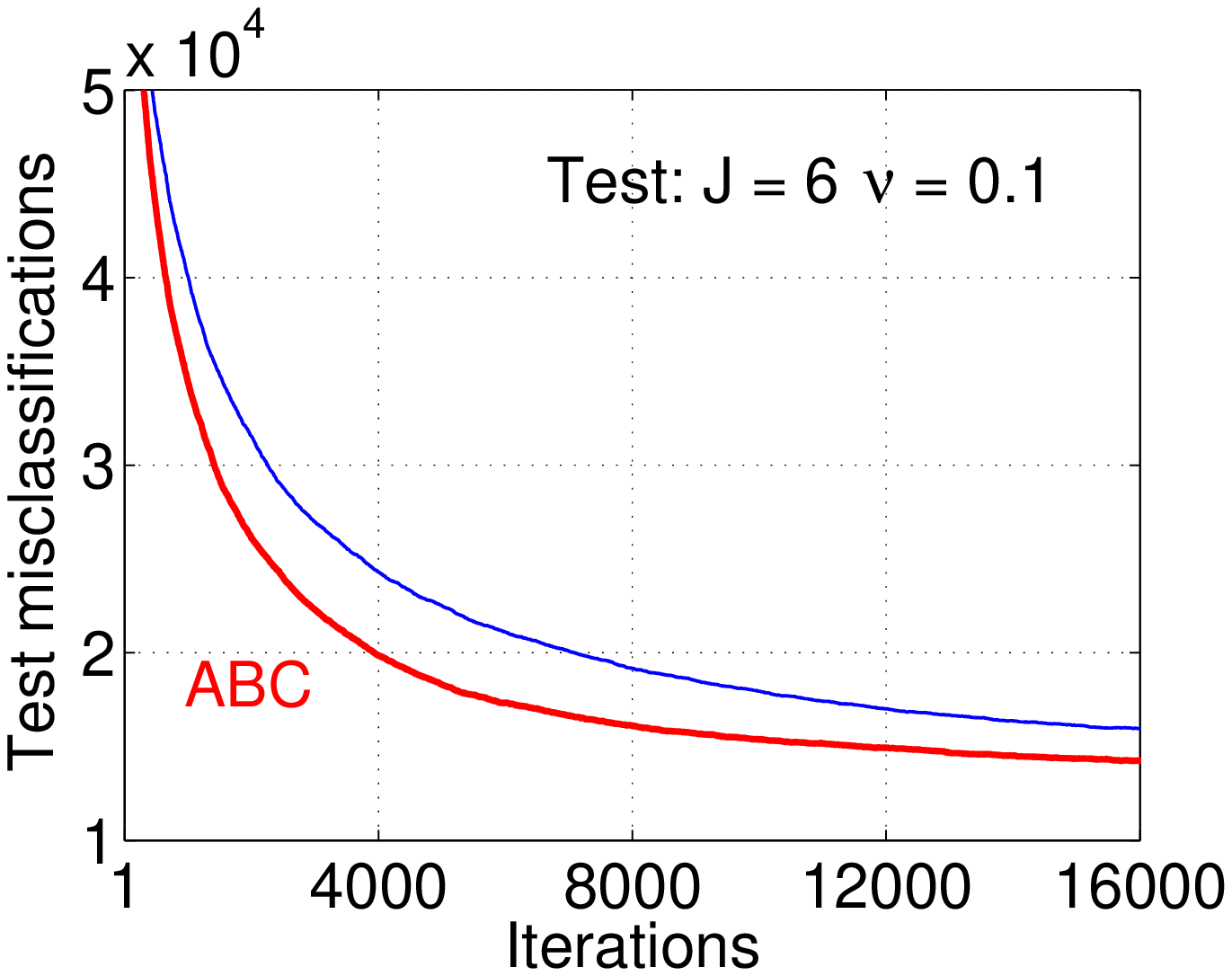}\hspace{0.1in}
{\includegraphics[width=2.0in]{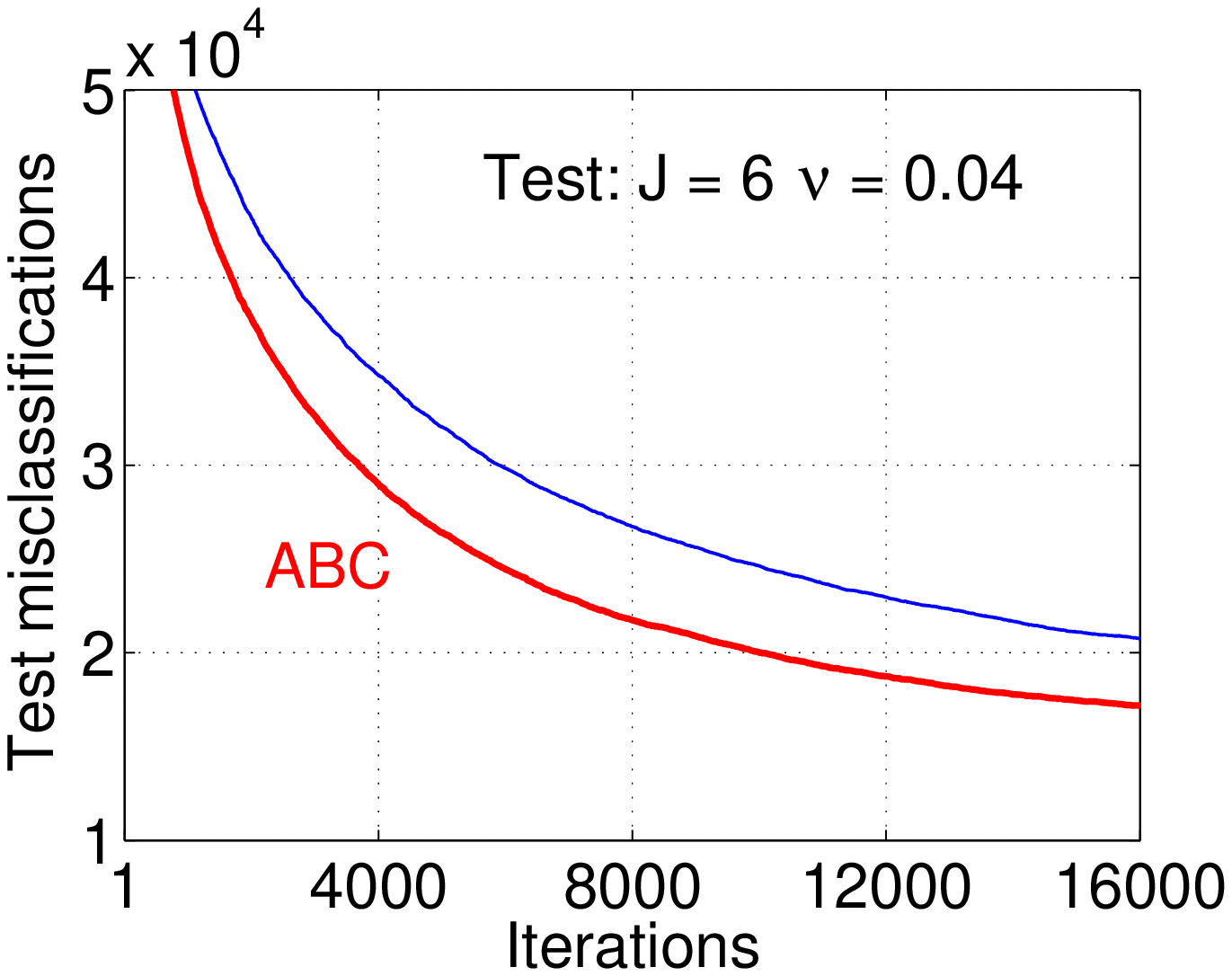}}}
\mbox{\includegraphics[width=2.0in]{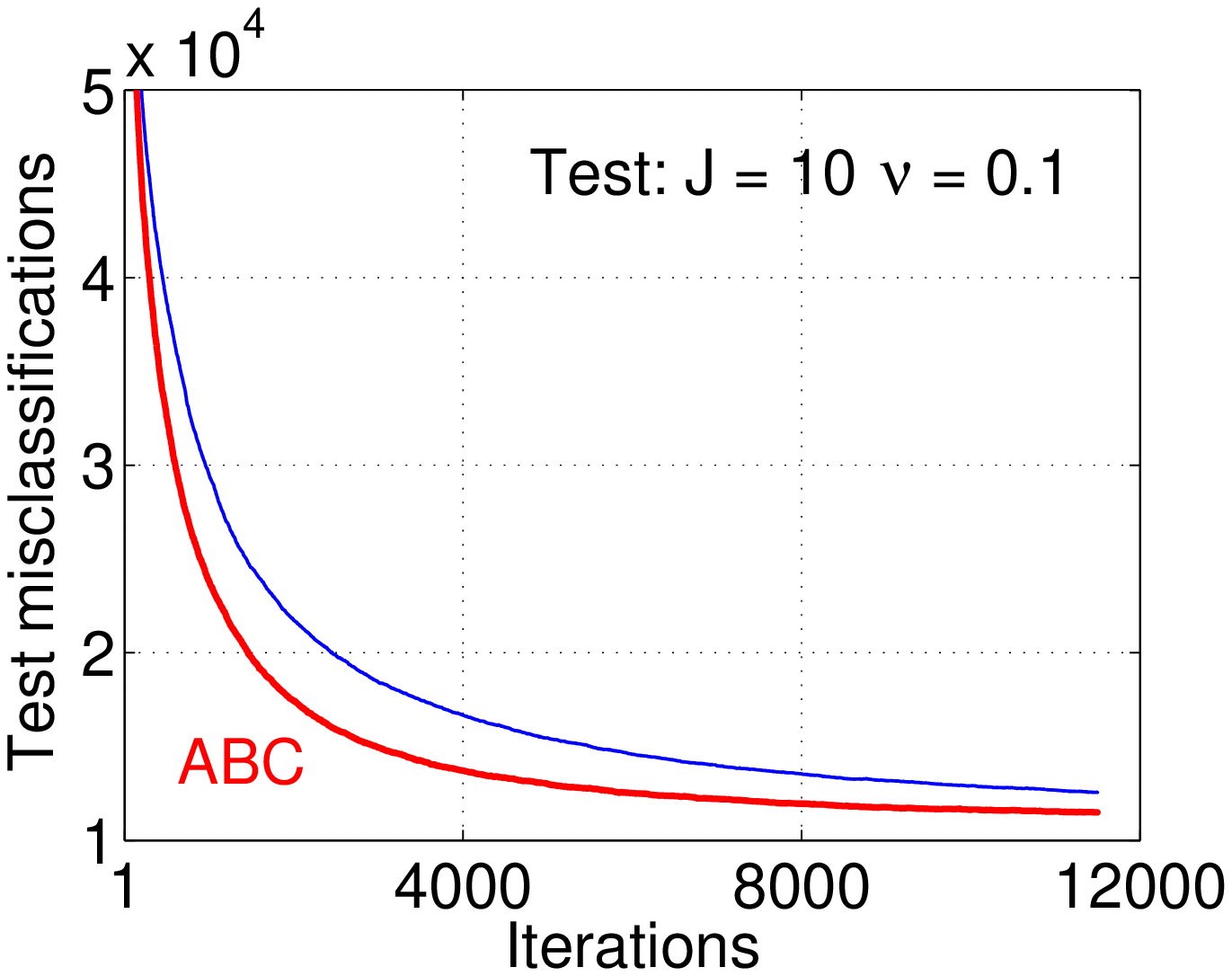}\hspace{0.1in}
{\includegraphics[width=2.0in]{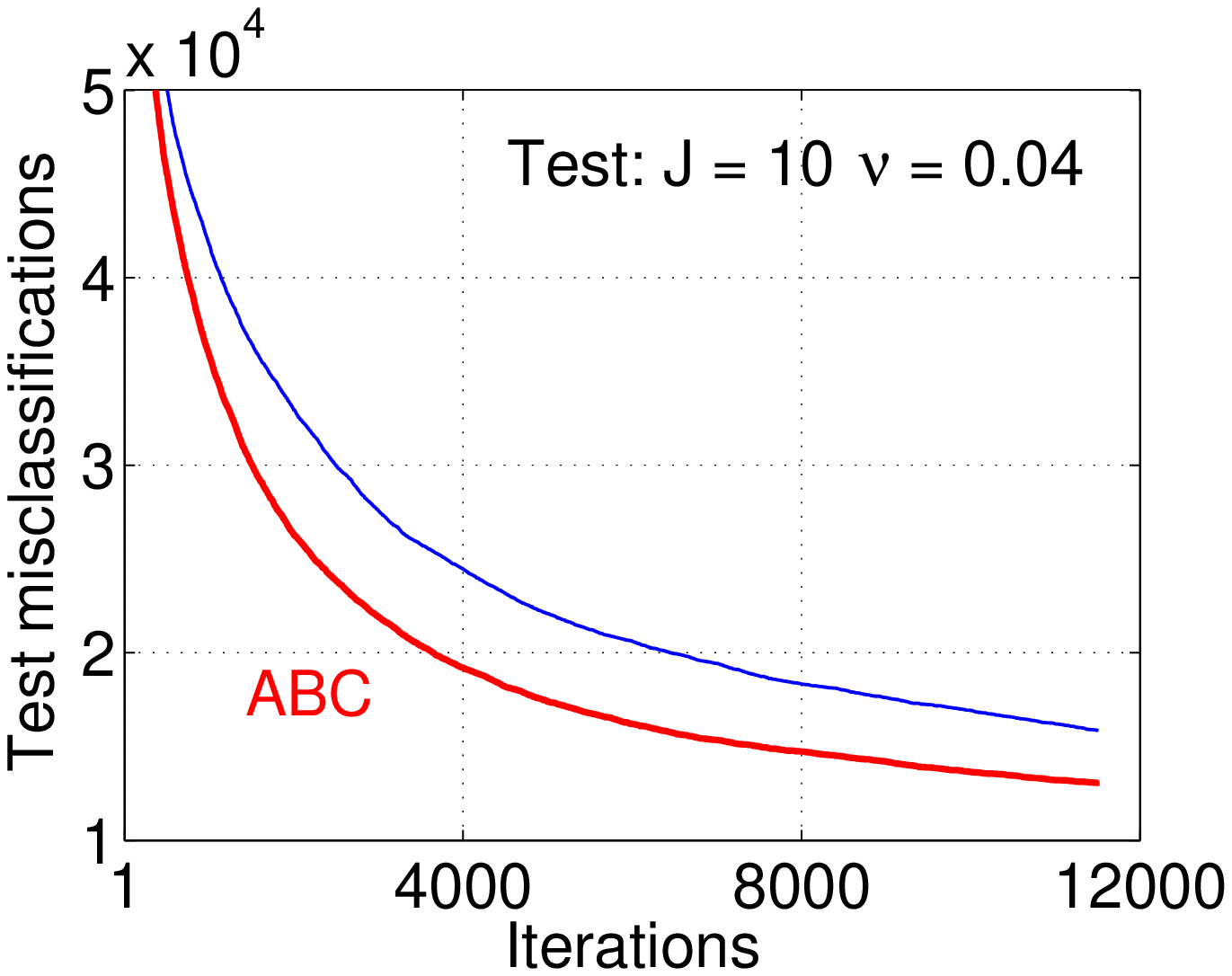}}}
\mbox{\includegraphics[width=2.0in]{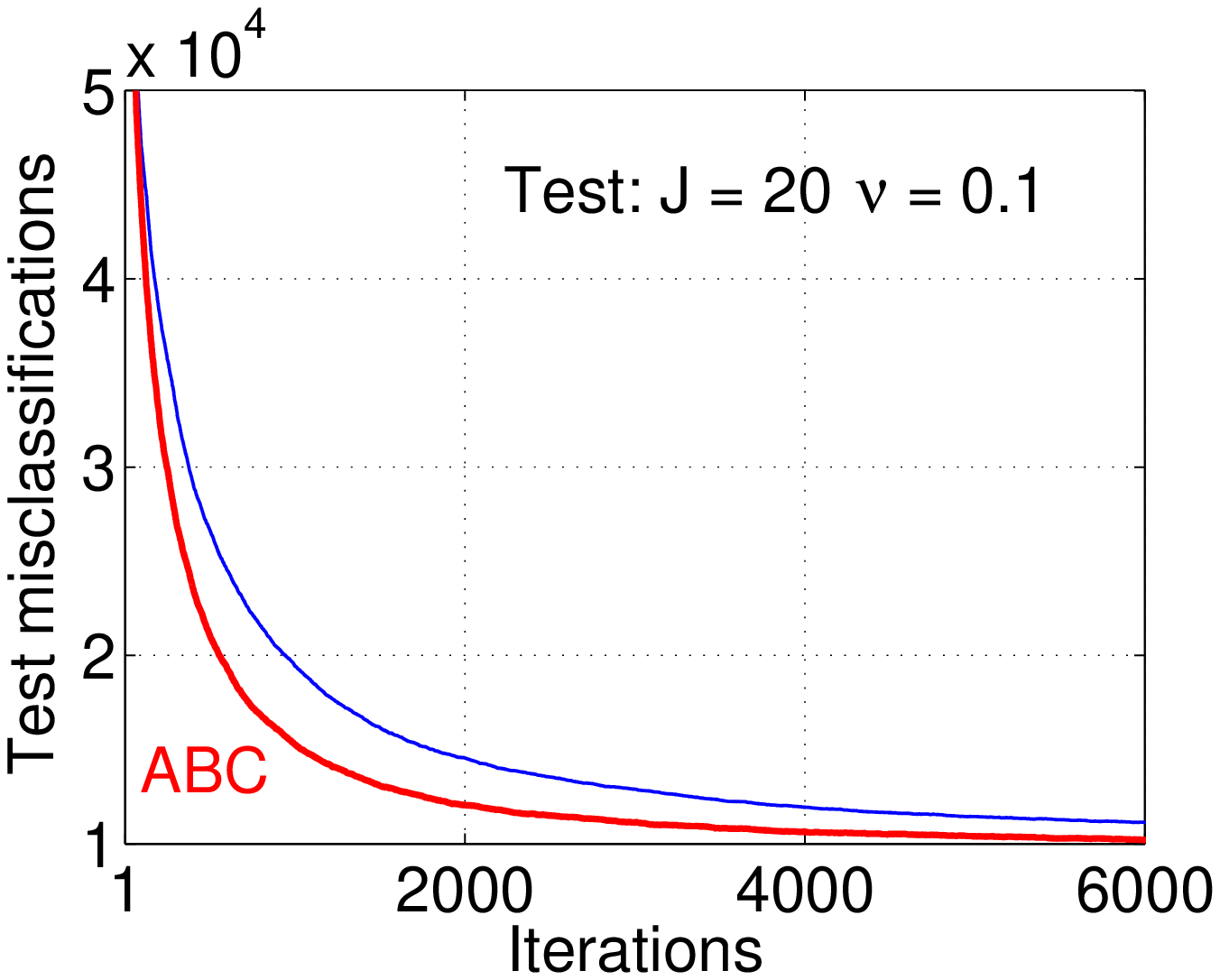}\hspace{0.1in}
{\includegraphics[width=2.0in]{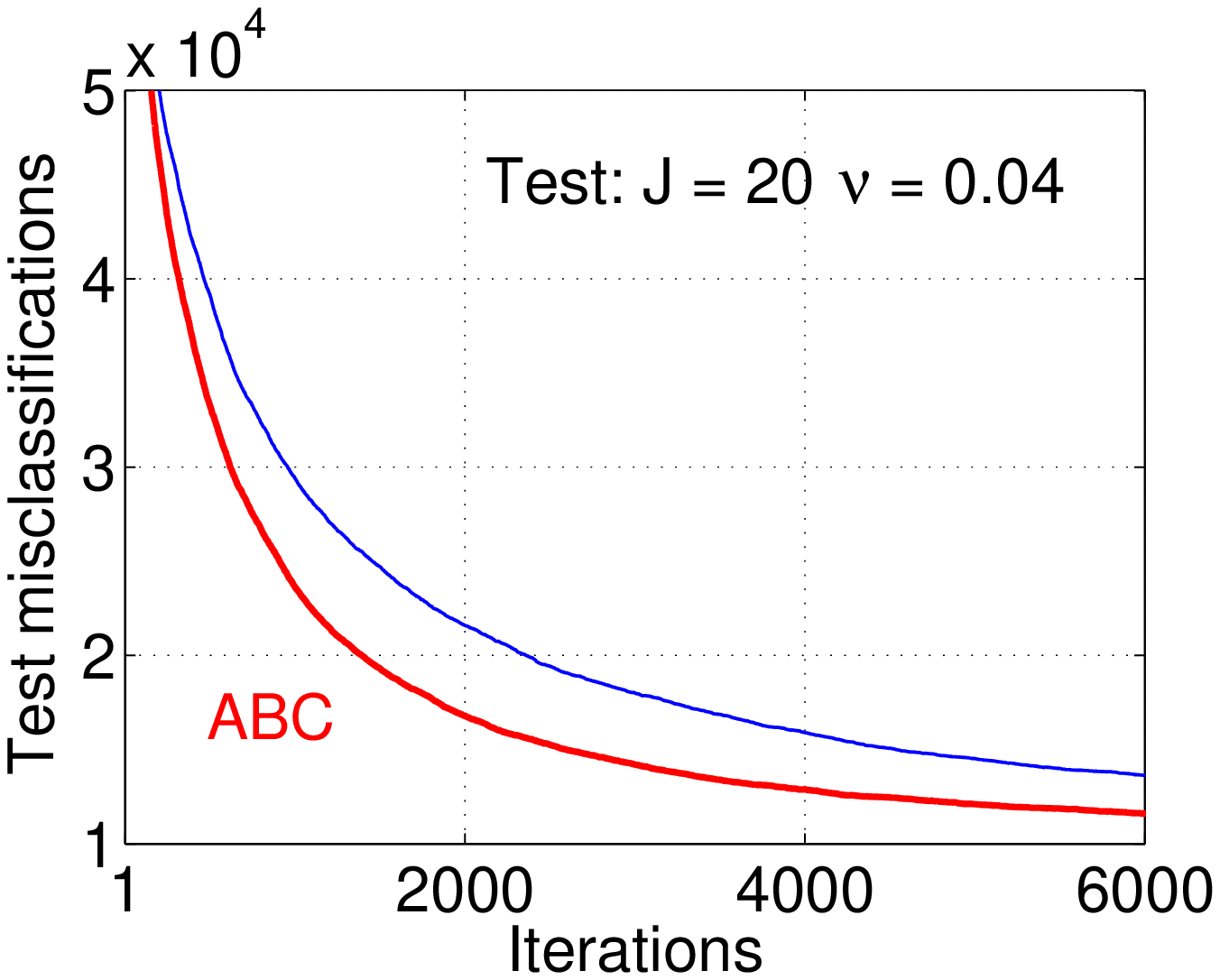}}}
\end{center}
\vspace{-0.20in}
\caption{\textbf{\em Covertype}. The test mis-classification errors. }\label{fig_CovertypeTest}
\end{figure}

Table \ref{tab_Covertype} summarizes the test mis-classification errors along with the relative improvements ($R_{err}$), for every combination of $J\in \{ 6, 10, 20\}$ and $\nu \in \{0.04, 0.1\}$. For each $J$ and $\nu$, the smallest test mis-classification errors, separately for ABC-MART and MART, are the lowest points in the curves in Figure \ref{fig_CovertypeTest}.

\begin{table}[h]
\caption{\textbf{\em Covertype}. The test mis-classification errors. The corresponding relative improvements ($R_{err}$, $\%$) of ABC-MART are included in parentheses.}
\begin{center}
\mbox{
\subtable[MART]{\begin{tabular}{l l l }
\hline \hline
  &$\nu = 0.04$ &$\nu=0.1$ \\
\hline
$J = 6$  &20756    &15959\\
$J=10$   &15862   &12553\\
$J=20$    &13630   &11133\\
\hline\hline
\end{tabular}}\hspace{0.4in}

\subtable[ABC-MART]{\begin{tabular}{l l l}
\hline \hline
  &$\nu = 0.04$ &$\nu=0.1$ \\
\hline
$J=6$ &17185 (17.2)  & 14230 (10.8)\\
$J=10$  &13064\   (17.6)  &11487 \ (8.5)\\
$J=20$  &11595\   (14.9)  &10203 \ (8.4)\\
\hline\hline
\end{tabular}}
}
\end{center}
\label{tab_Covertype}
\end{table}

\begin{figure}[h]
\begin{center}
\mbox{\includegraphics[width=2.0in]{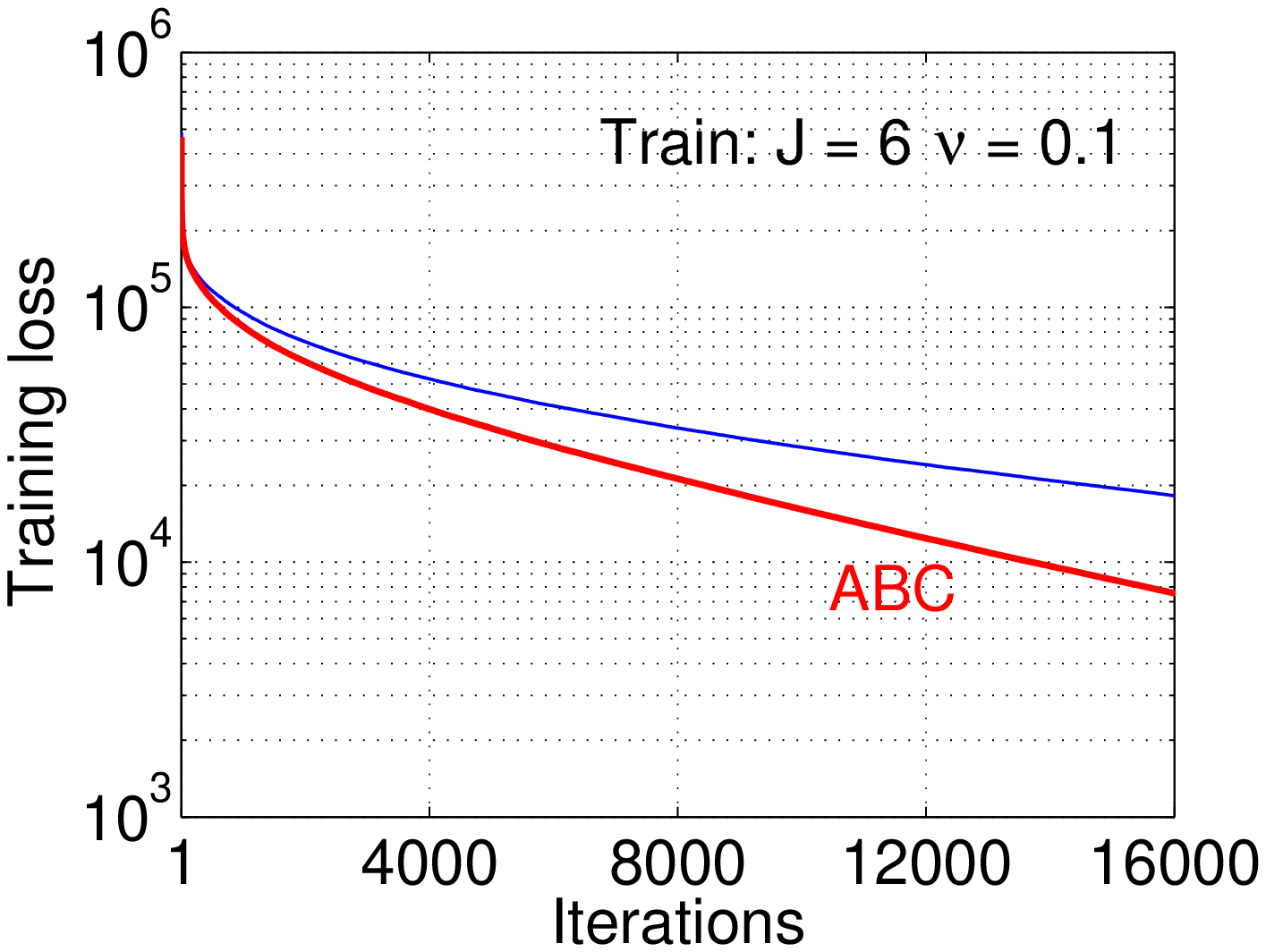}\hspace{0.1in}
{\includegraphics[width=2.0in]{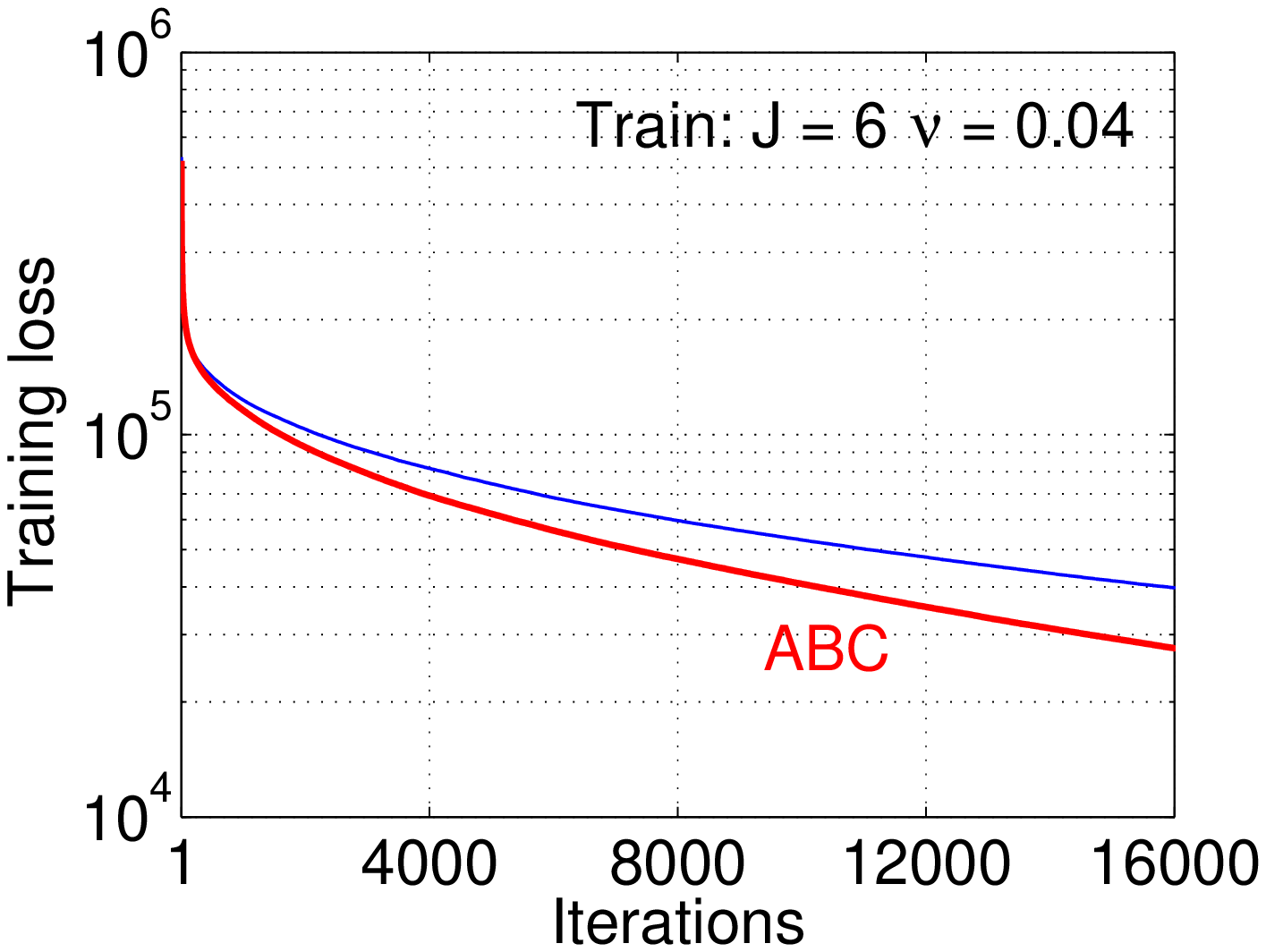}}}
\mbox{\includegraphics[width=2.0in]{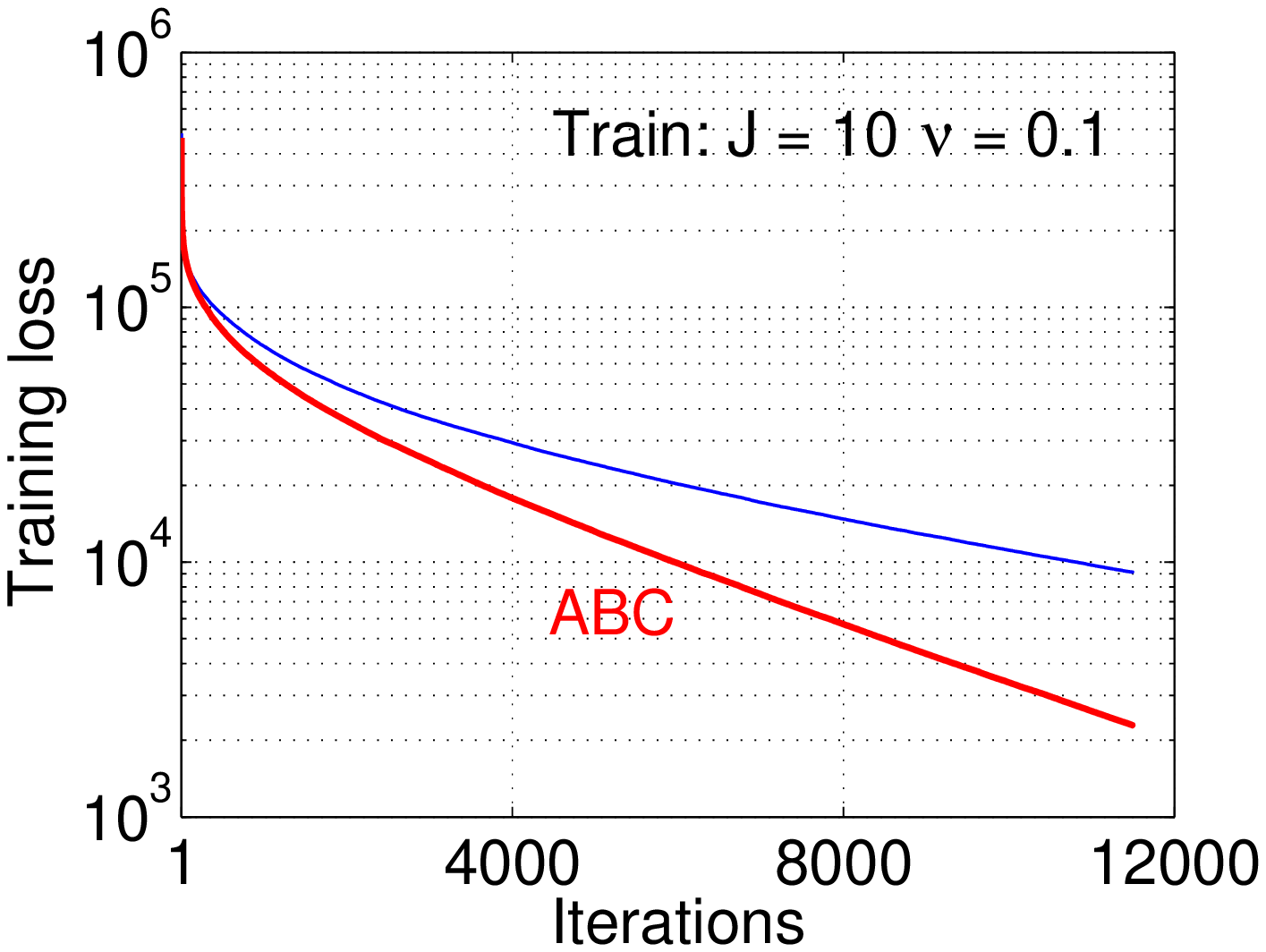}\hspace{0.1in}
{\includegraphics[width=2.0in]{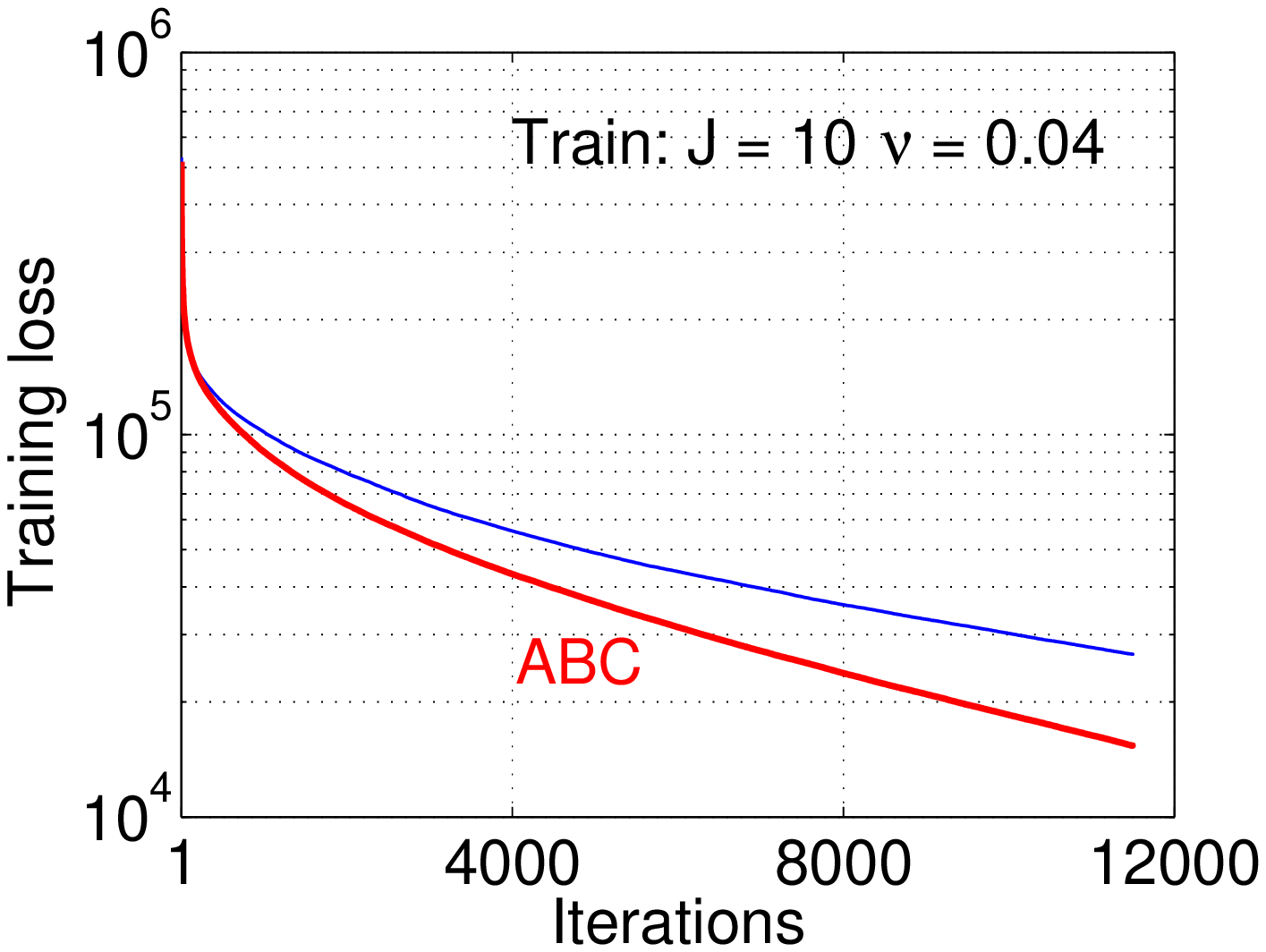}}}
\mbox{\includegraphics[width=2.0in]{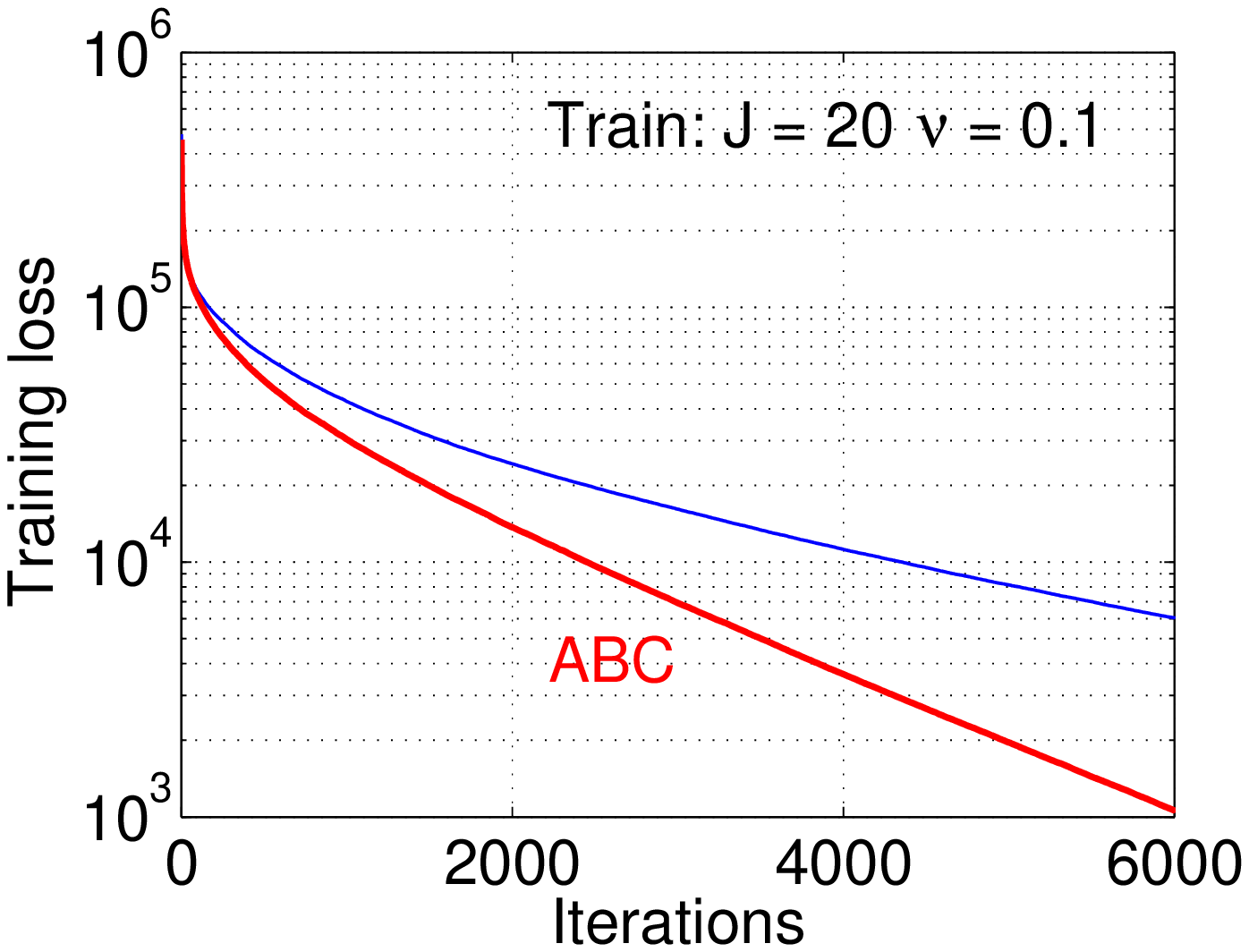}\hspace{0.1in}
{\includegraphics[width=2.0in]{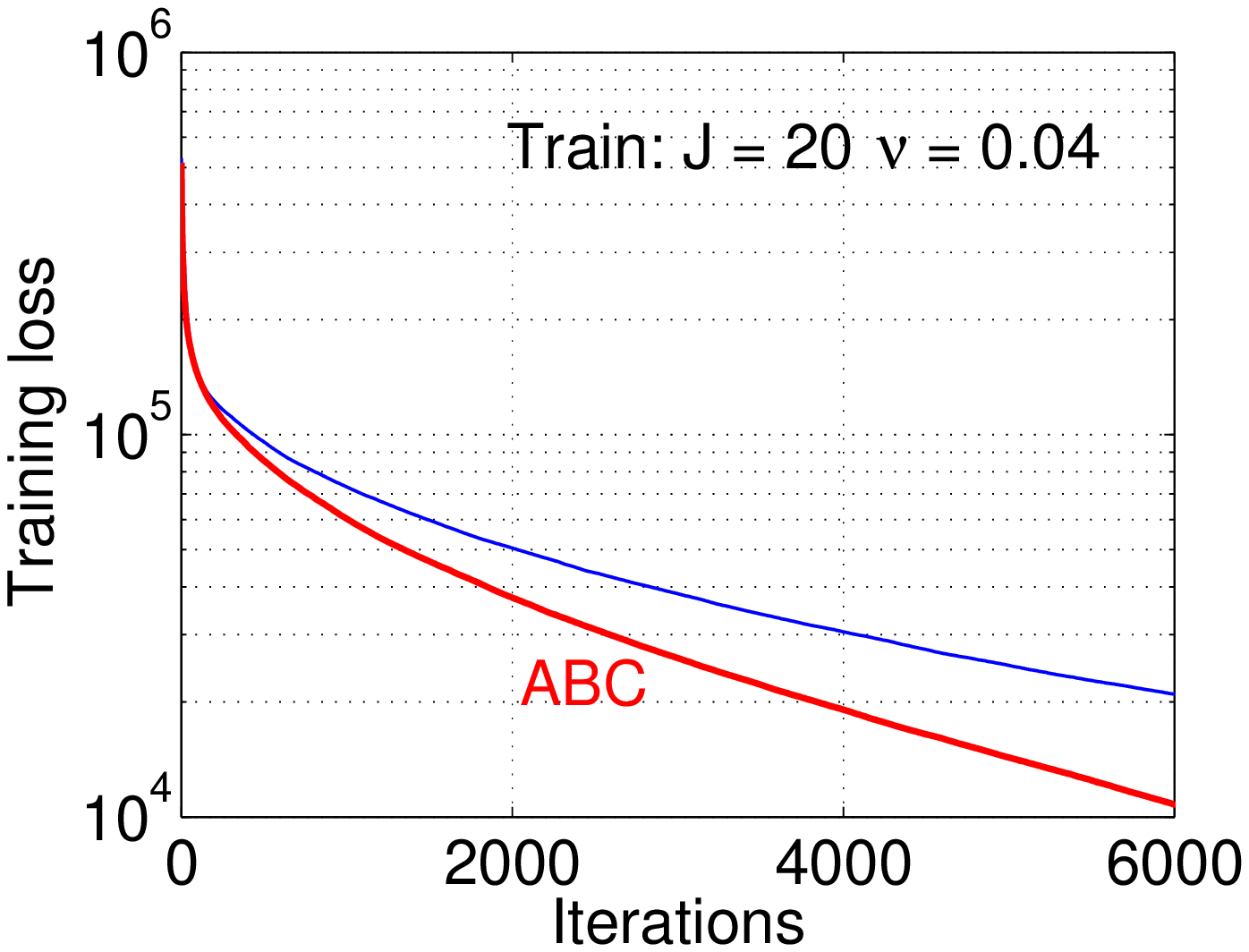}}}
\end{center}
\vspace{-0.2in}
\caption{\textbf{\em Covertype}. The training loss, i.e., (\ref{eqn_loss}). The curves labeled ``ABC'' correspond to ABC-MART. }\label{fig_CovertypeTrain}
\end{figure}

\begin{figure}[h]
\begin{center}
\mbox{\includegraphics[width=2.0in]{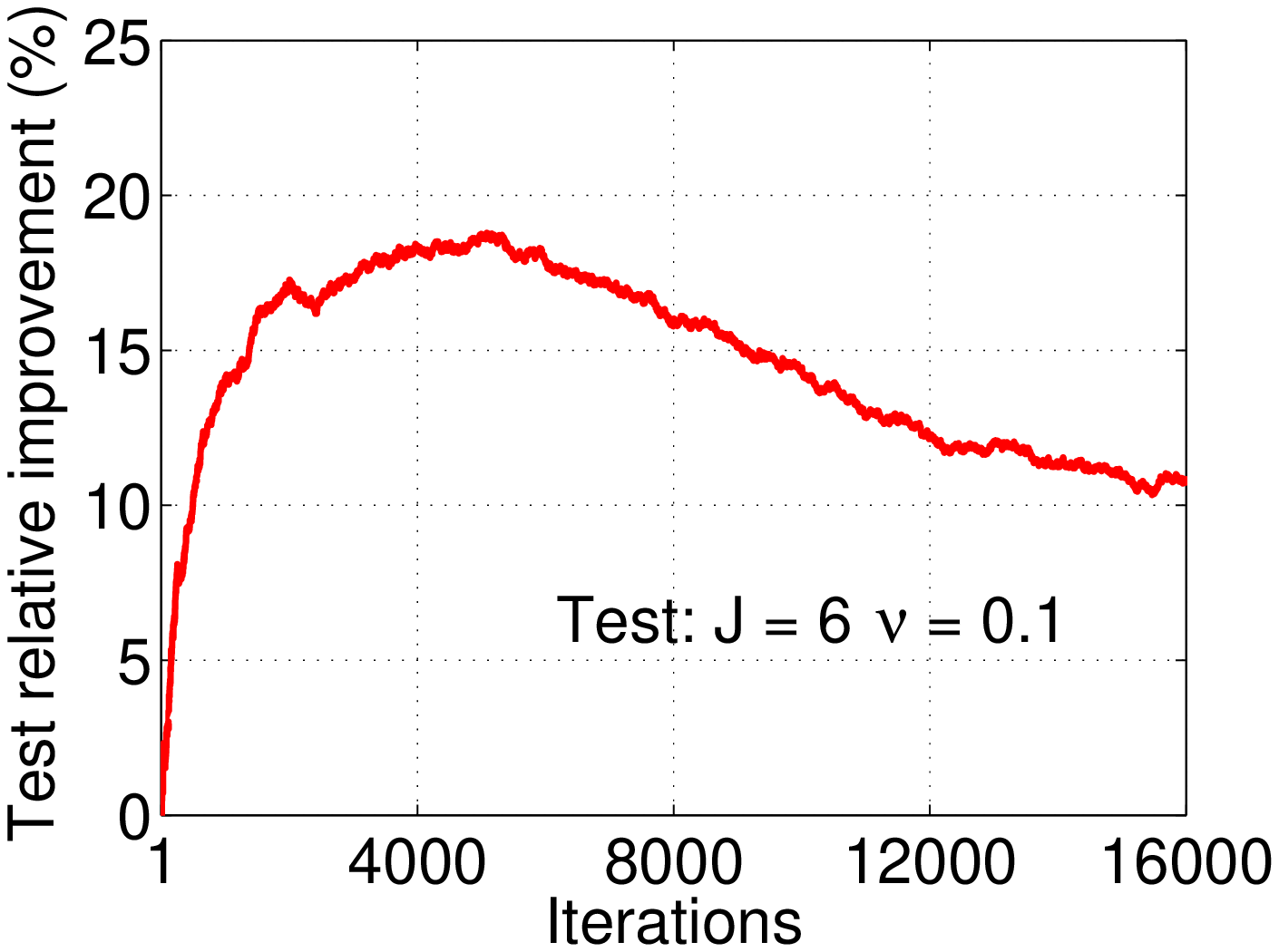}\hspace{0.1in}
{\includegraphics[width=2.0in]{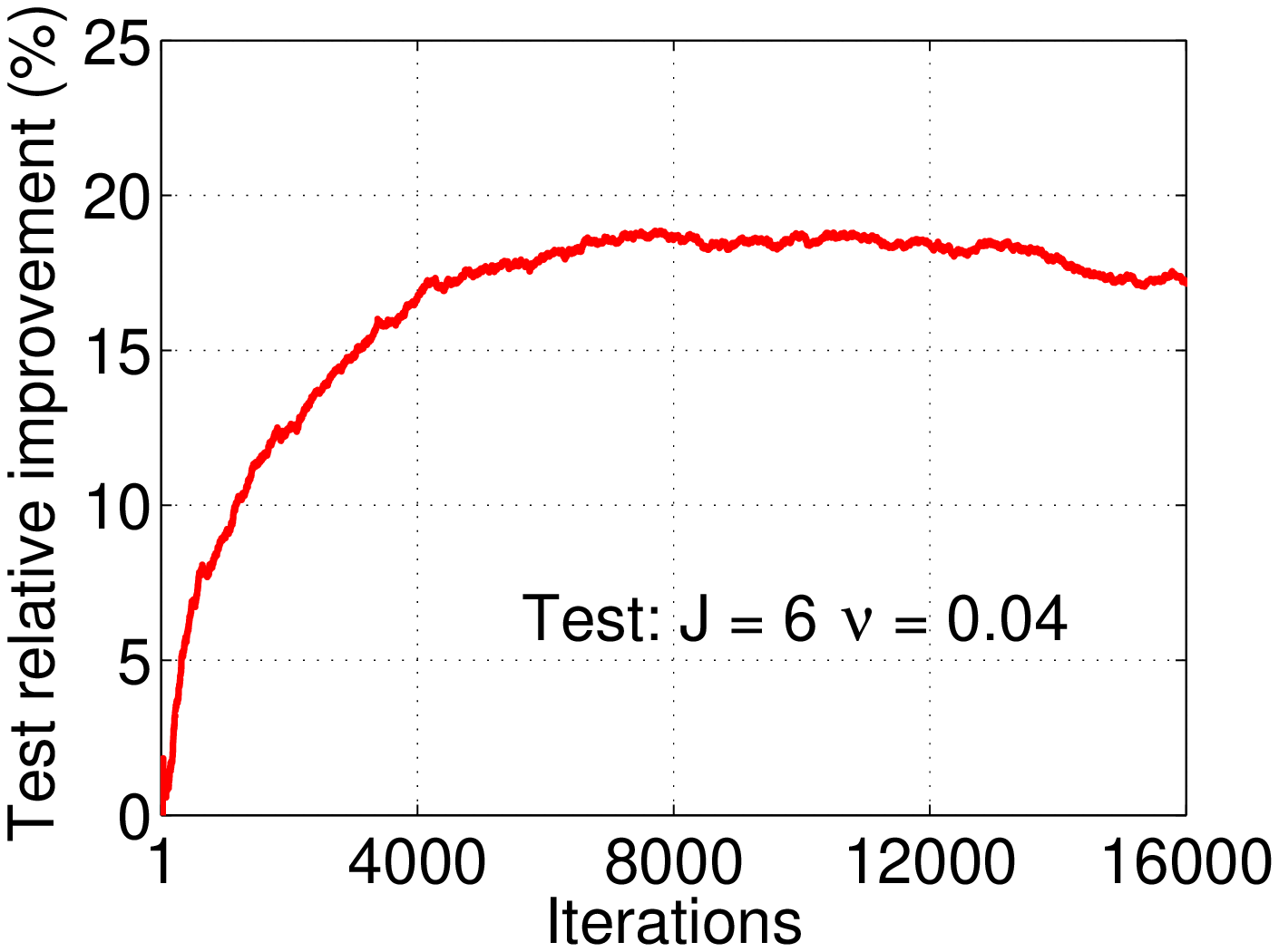}}}
\mbox{\includegraphics[width=2.0in]{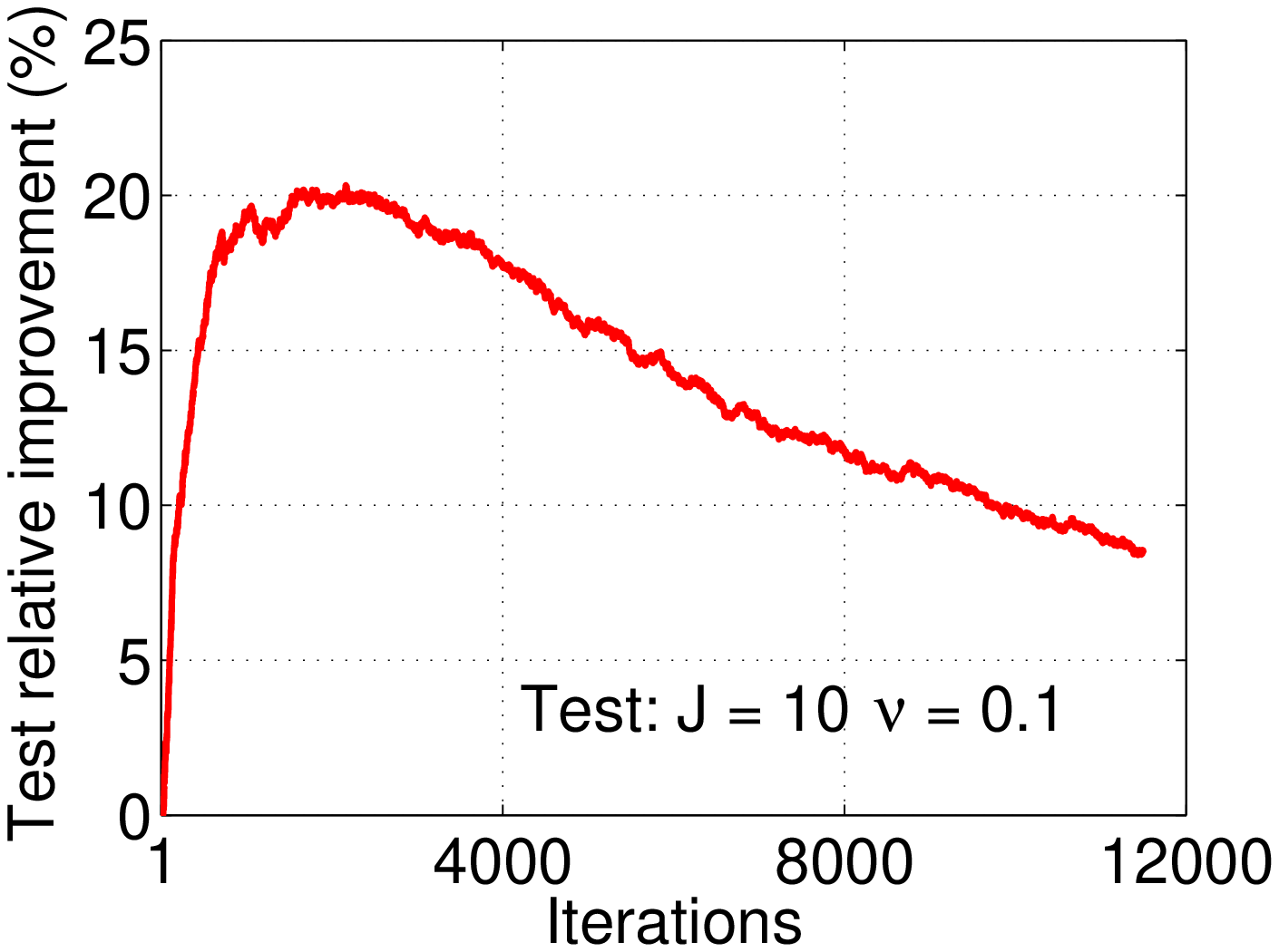}\hspace{0.1in}
{\includegraphics[width=2.0in]{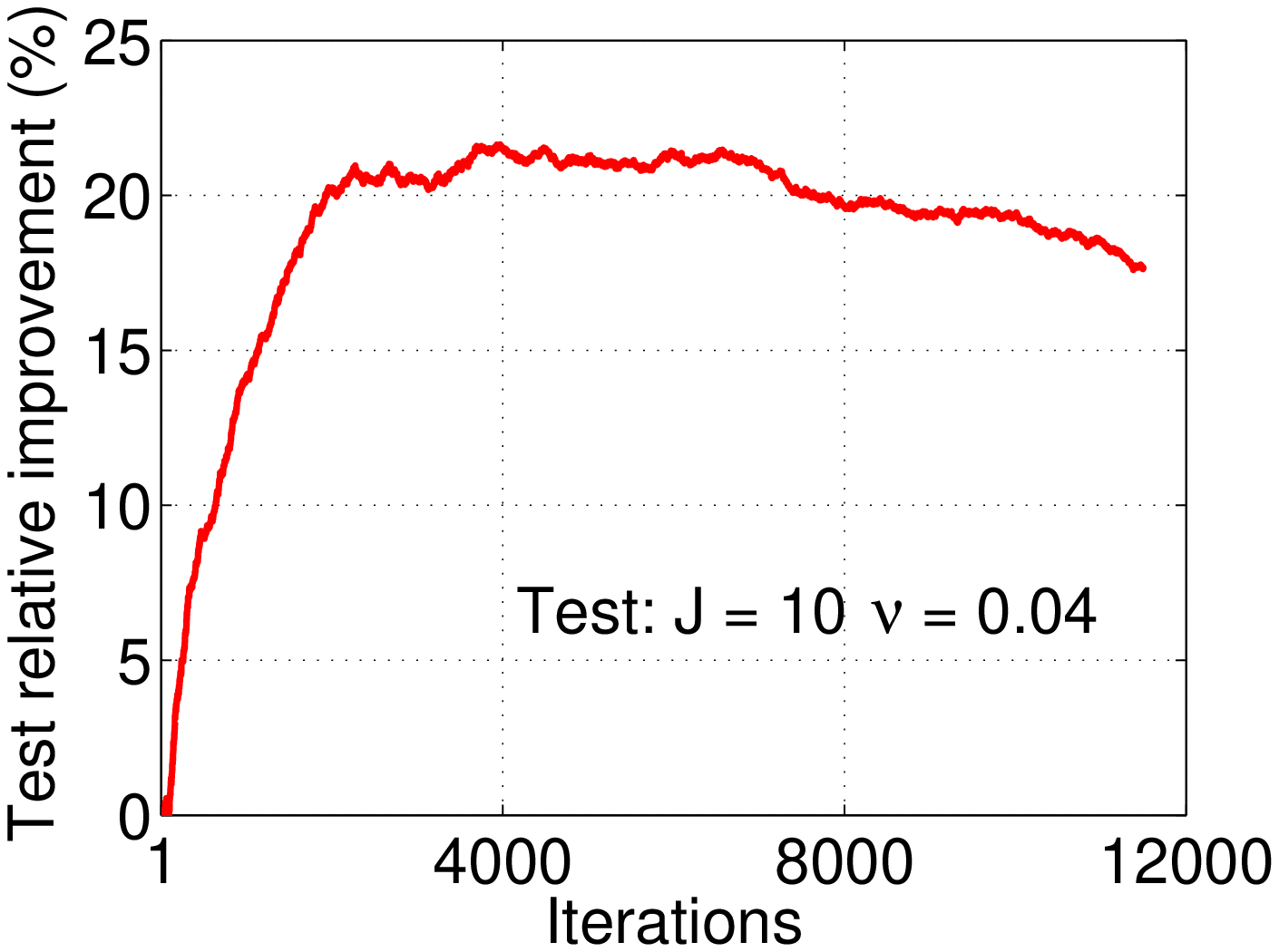}}}
\mbox{\includegraphics[width=2.0in]{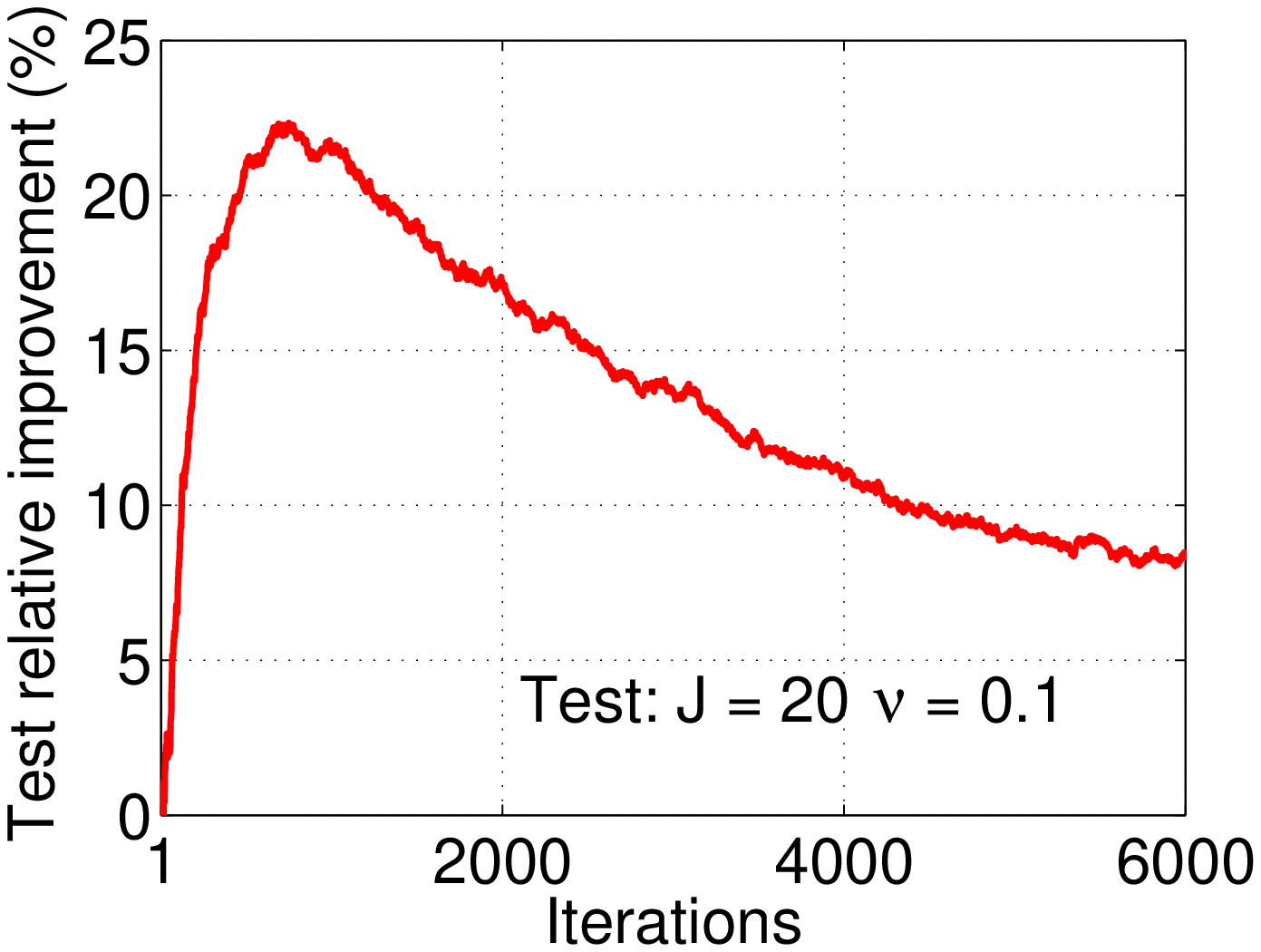}\hspace{0.1in}
{\includegraphics[width=2.0in]{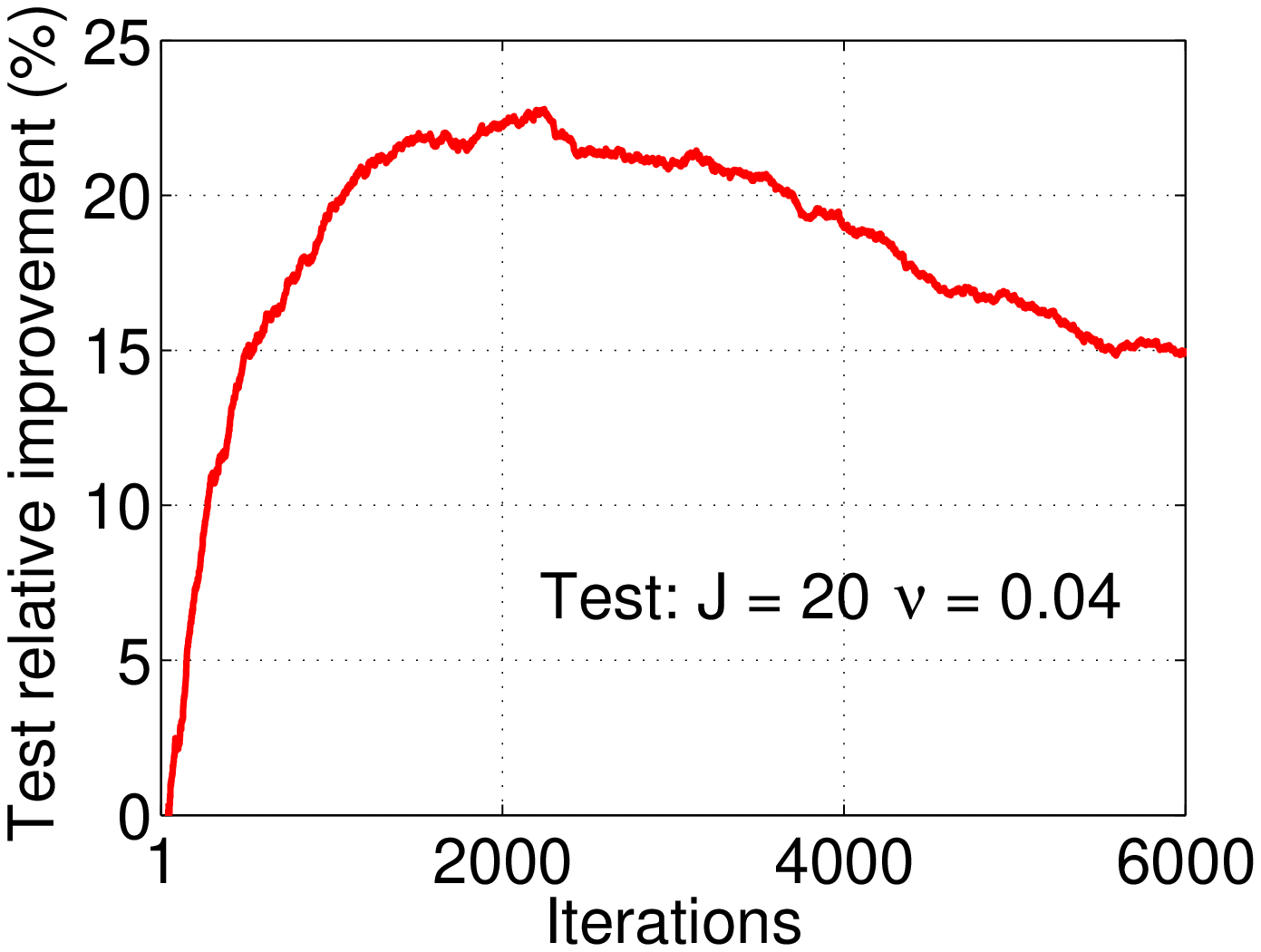}}}
\end{center}
\vspace{-0.2in}
\caption{\textbf{\em Covertype}. The relative improvements, $R_{err}$ ($\%$).  }\label{fig_CovertypeImprove}
\end{figure}

To report the experiments in a more informative manner,  Figure \ref{fig_CovertypeTrain}, Figure \ref{fig_CovertypeTest}, and Figure \ref{fig_CovertypeImprove}, respectively, present the training loss, the test mis-classification errors, and the relative improvements, for the complete history of $M$ boosting steps (iterations).

Figure \ref{fig_CovertypeTrain} indicates that ABC-MART reduces the training loss (\ref{eqn_loss}) considerably and consistently faster than MART. Figure \ref{fig_CovertypeTest} demonstrates that ABC-MART exhibits considerably and consistently smaller test mis-classification errors than MART.

Figure \ref{fig_CovertypeImprove} illustrates that the relative improvements of ABC-MART over MART, in terms of the mis-classification errors, may be considerably larger than the numbers reported in Table \ref{tab_Covertype} if we stop the training earlier. This phenomenon may be quite beneficial for real-world large-scale applications when either the training or test time is part of the performance measure. For example, real-time applications (e.g., search engines) may not be able to afford to use models with a very large number of boosting steps.

\clearpage

The next five subsections are respectively devoted to presenting the experiment results of five small data sets. The results exhibit similar characteristics across data sets.

\subsection{Experiments on the {\em  Letter} Data Set}

Table \ref{tab_Letter} summarizes the test mis-classification errors along with the relative improvements for every combination of $J\in \{ 4,\  6, \ 8, \ 10, \ 12, \ 14, 16\}$ and $\nu \in \{0.04, \ 0.06, \ 0.08,\ 0.1\}$.

\begin{table}[h]
\caption{\textbf{\em Letter}. The test mis-classification errors. The corresponding relative improvements ($R_{err}$, $\%$) of ABC-MART are included in parentheses. }
\begin{center}
{
\subtable[MART]{\begin{tabular}{l l l l l }
\hline \hline
  &$\nu = 0.04$ &$\nu=0.06$ &$\nu=0.08$ &$\nu=0.1$ \\
\hline
$J=4$   &176   &178   &177   &173\\
$J=6$   &154   &160   &156   &158\\
$J=8$   &151   &145   &151   &154\\
$J=10$    &141   &141   &147   &144\\
$J=12$      &150   &144   &144   &140\\
$J=14$     &143   &147   &144   &146\\
$J=16$     &138   &147   &142   &135
\\\hline\hline
\end{tabular}}

\subtable[ABC-MART]{\begin{tabular}{l l l l l }
\hline \hline
  &$\nu = 0.04$ &$\nu=0.06$ &$\nu=0.08$ &$\nu=0.1$ \\
\hline
$J=4$   &155 \ (11.9)   &143\  (19.7)  & 148\  (16.4)   &144\ (16.8)\\
$J=6$   &140 \ (9.1)   &141 \ (11.9)  &130 \ (16.7)   &121\  (23.4)\\
$J=8$    &132 \ (12.6)   &126 \ (13.1)  & 124\ (17.9)   &117\ (24.0)\\
$J=10$    & 131 \ (7.1)  & 119\ (15.6) &  116 \ (21.1) & 115\ (20.1)\\
$J=12$   &125 \ (16.7)  &124\ (13.9)   &119\ (17.4)   &119\ (15.0)\\
$J=14$   &  117 \ (18.2)   &118 \ (19.7) &  123\ (14.6)   &112\ (23.3)\\
$J=16$   & 117  \ (15.2)  &113\ (23.1)  & 113\ (20.4)   &111\ (17.8)
\\\hline\hline
\end{tabular}}
}
\end{center}
\label{tab_Letter}
\end{table}

\begin{figure}[h]
\begin{center}
\mbox{\includegraphics[width=2.0in]{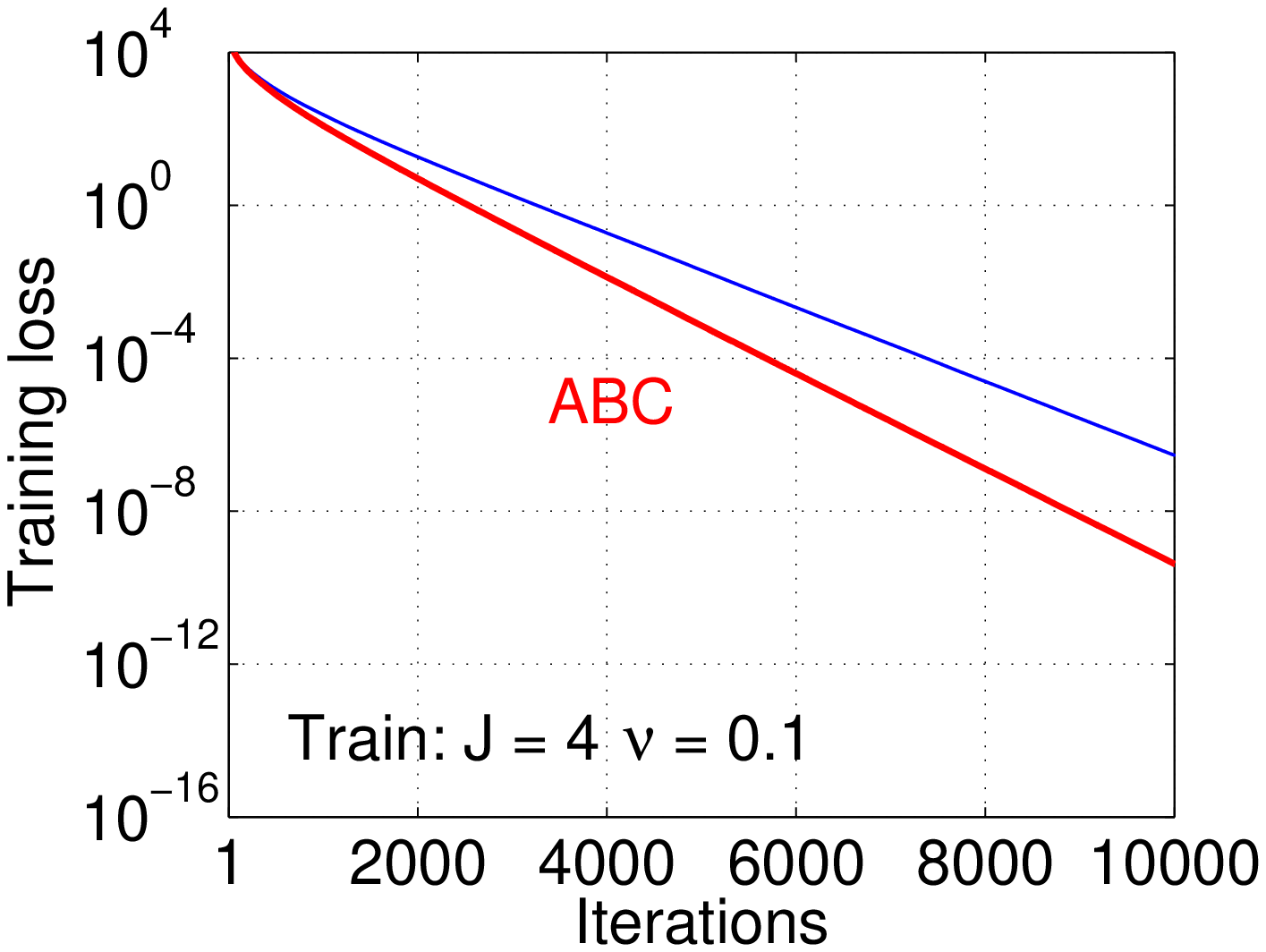}\hspace{0.1in}
{\includegraphics[width=2.0in]{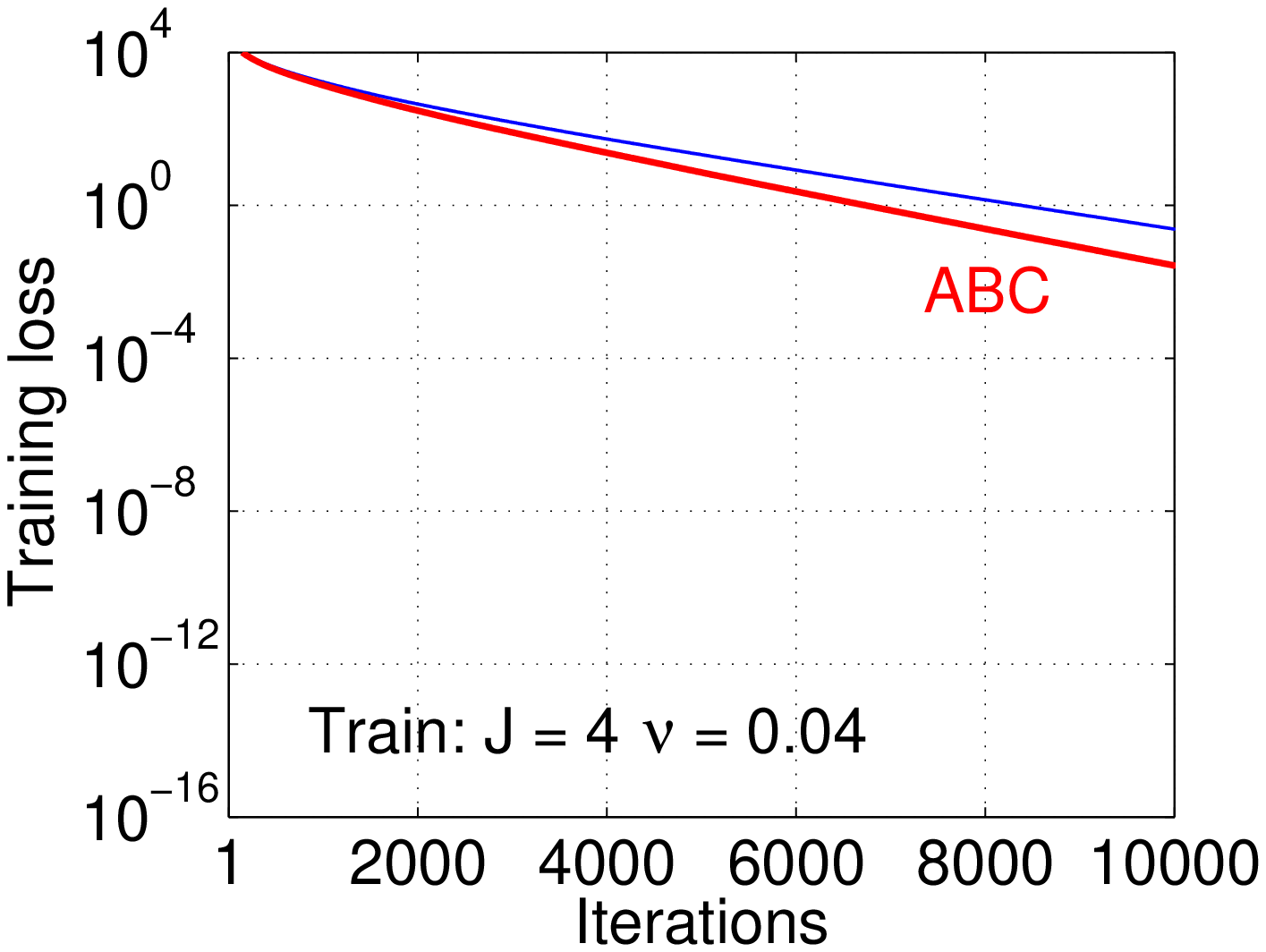}}}
\mbox{\includegraphics[width=2.0in]{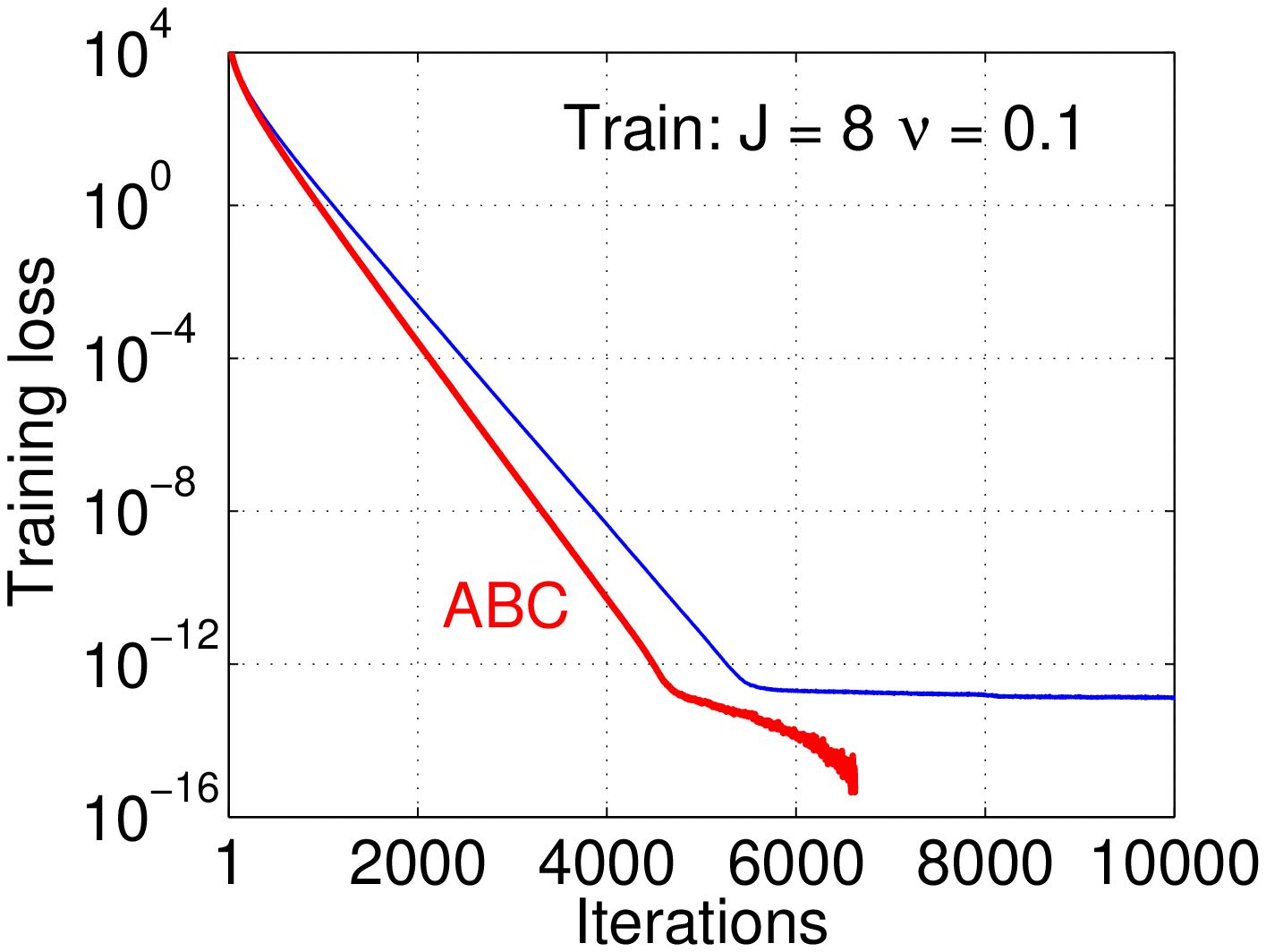}\hspace{0.1in}
{\includegraphics[width=2.0in]{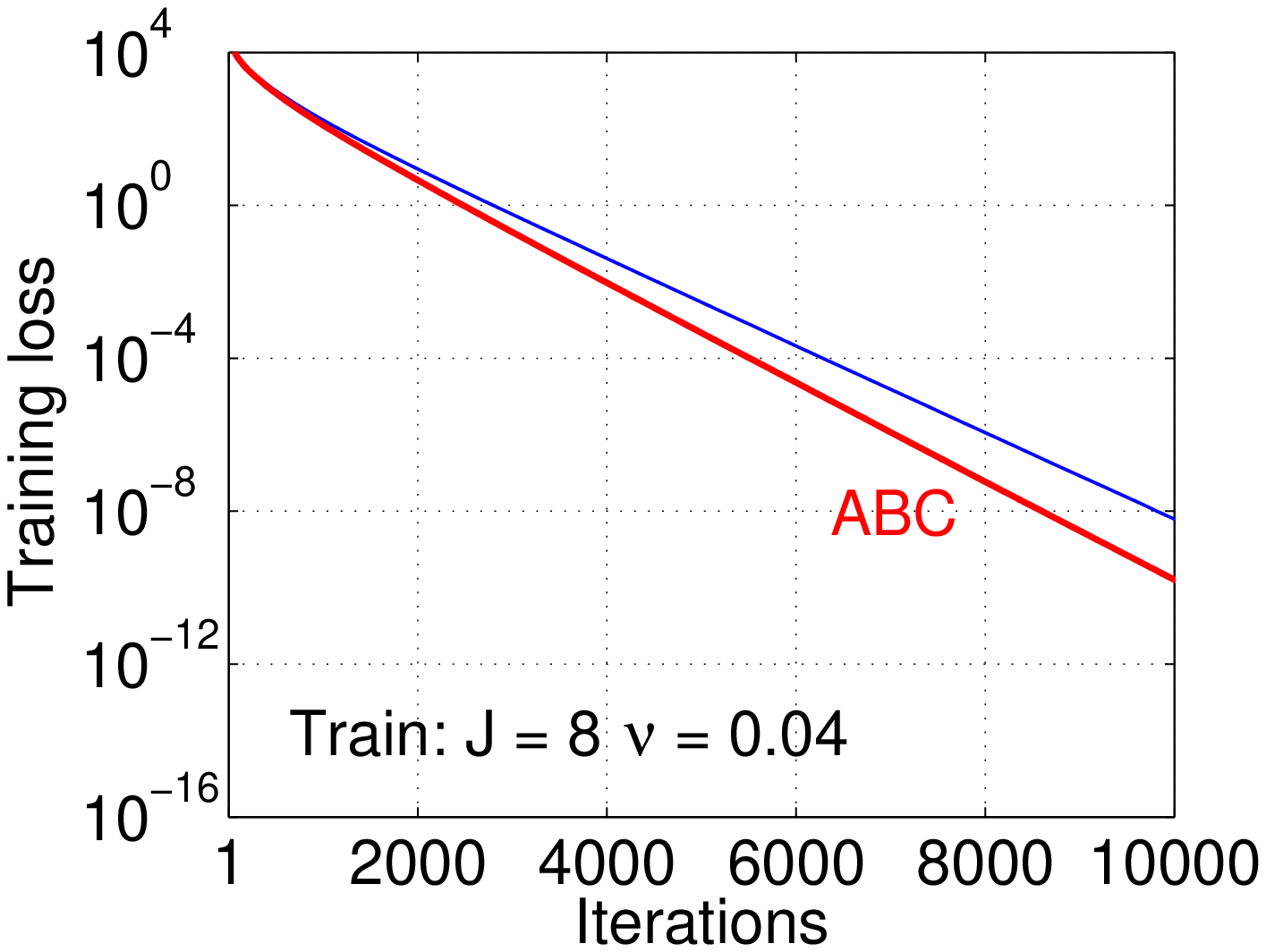}}}
\mbox{\includegraphics[width=2.0in]{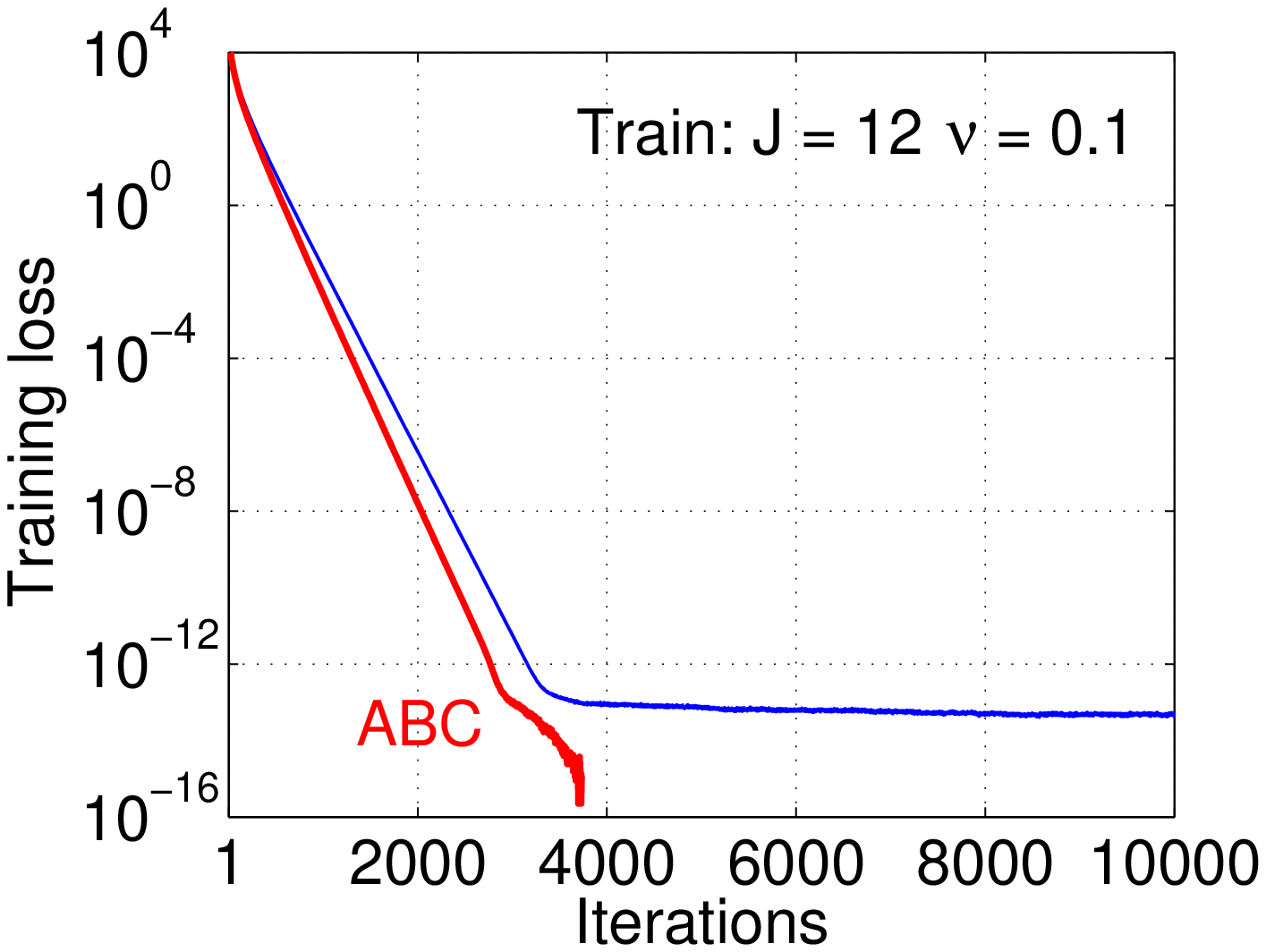}\hspace{0.1in}
{\includegraphics[width=2.0in]{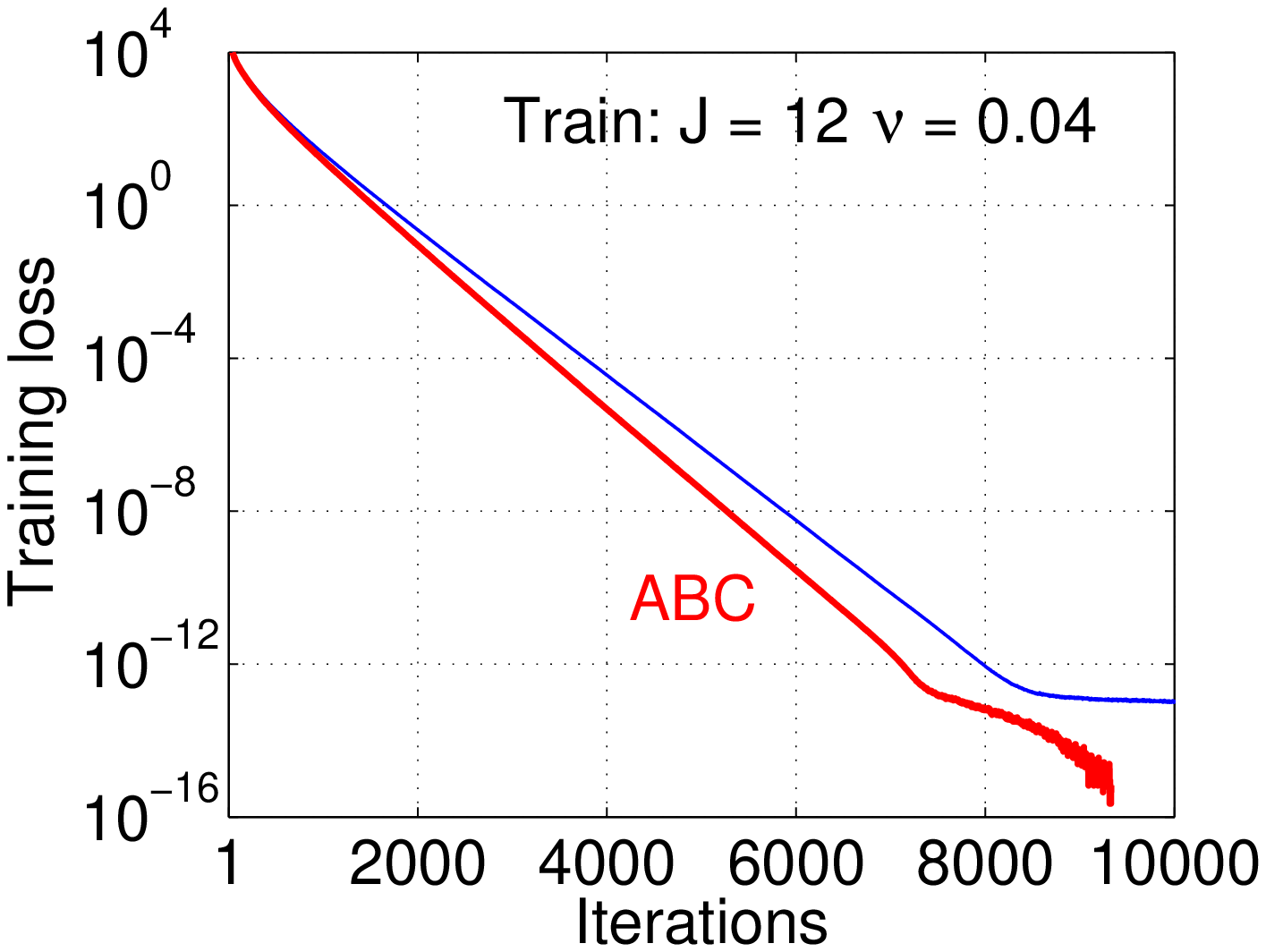}}}
\mbox{\includegraphics[width=2.0in]{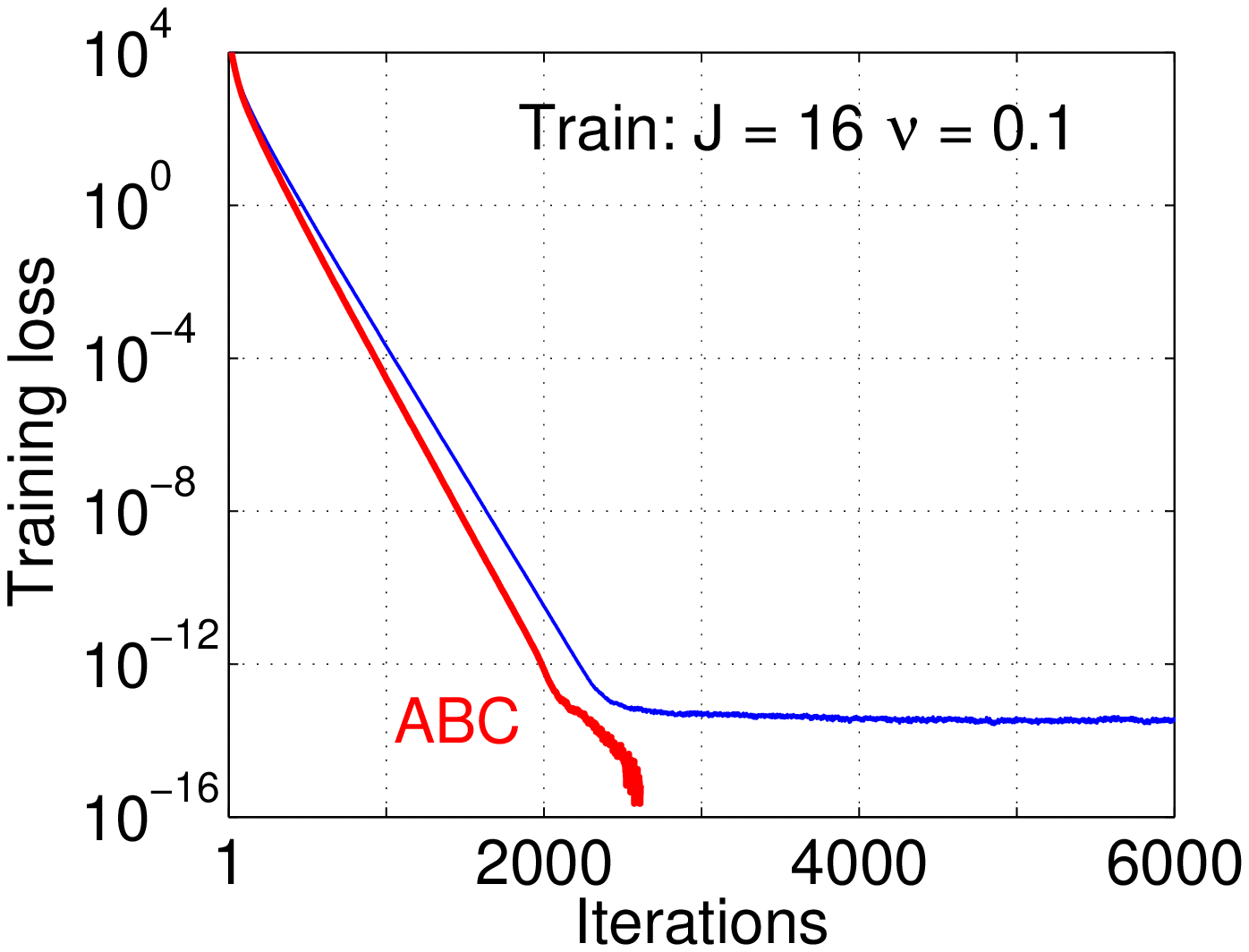}\hspace{0.1in}
{\includegraphics[width=2.0in]{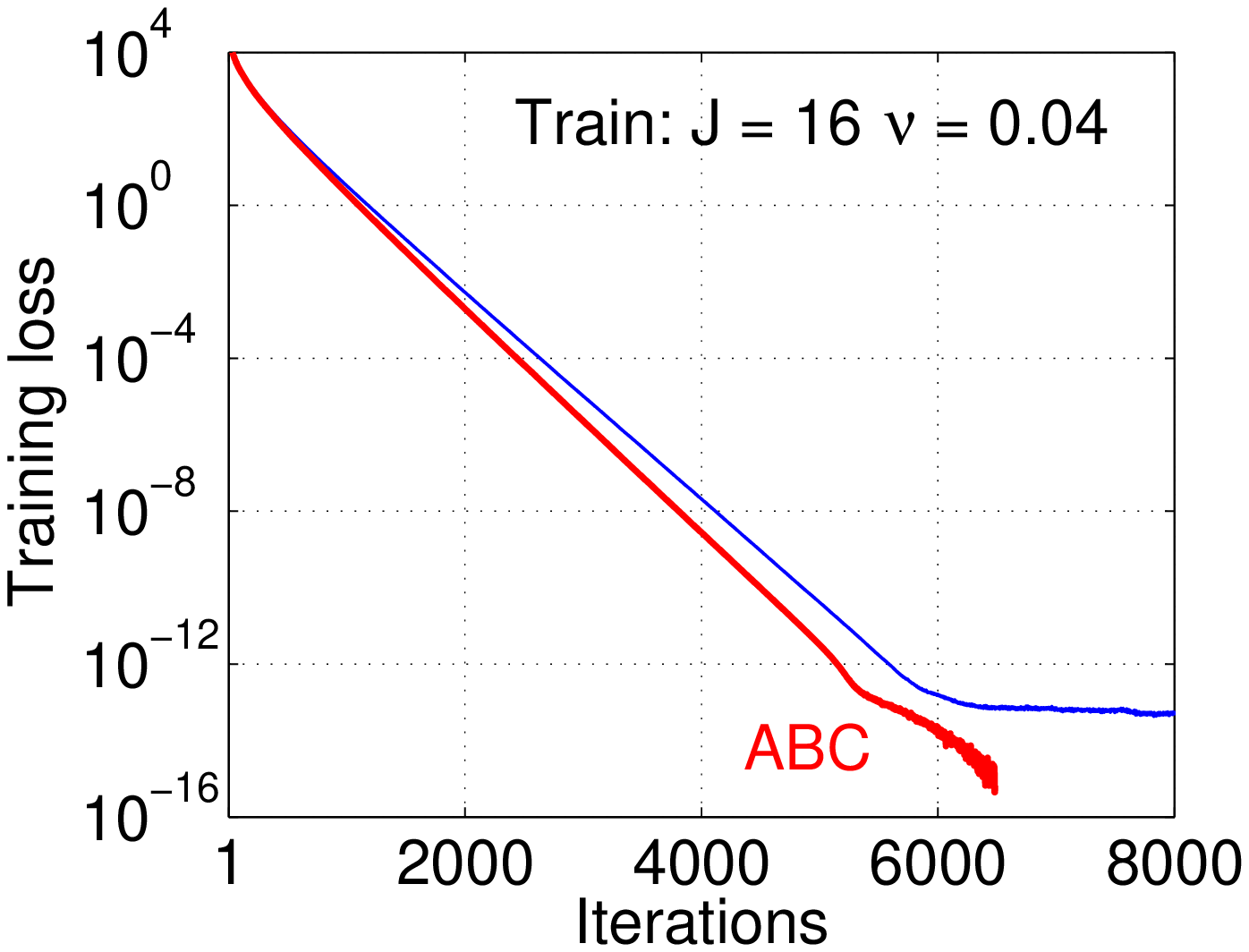}}}
\end{center}
\vspace{-0.2in}
\caption{\textbf{\em Letter}. The training loss, i.e., (\ref{eqn_loss}). The curves labeled ``ABC'' correspond to ABC-MART. }\label{fig_LetterTrain}
\end{figure}

Figure \ref{fig_LetterTrain} again indicates that ABC-MART reduces the training loss (\ref{eqn_loss}) considerably and consistently faster than MART. Since this is a small data set, the training loss could approach zero even when the number of boosting steps is not too large.

\begin{figure}[h]
\begin{center}
\mbox{\includegraphics[width=2.0in]{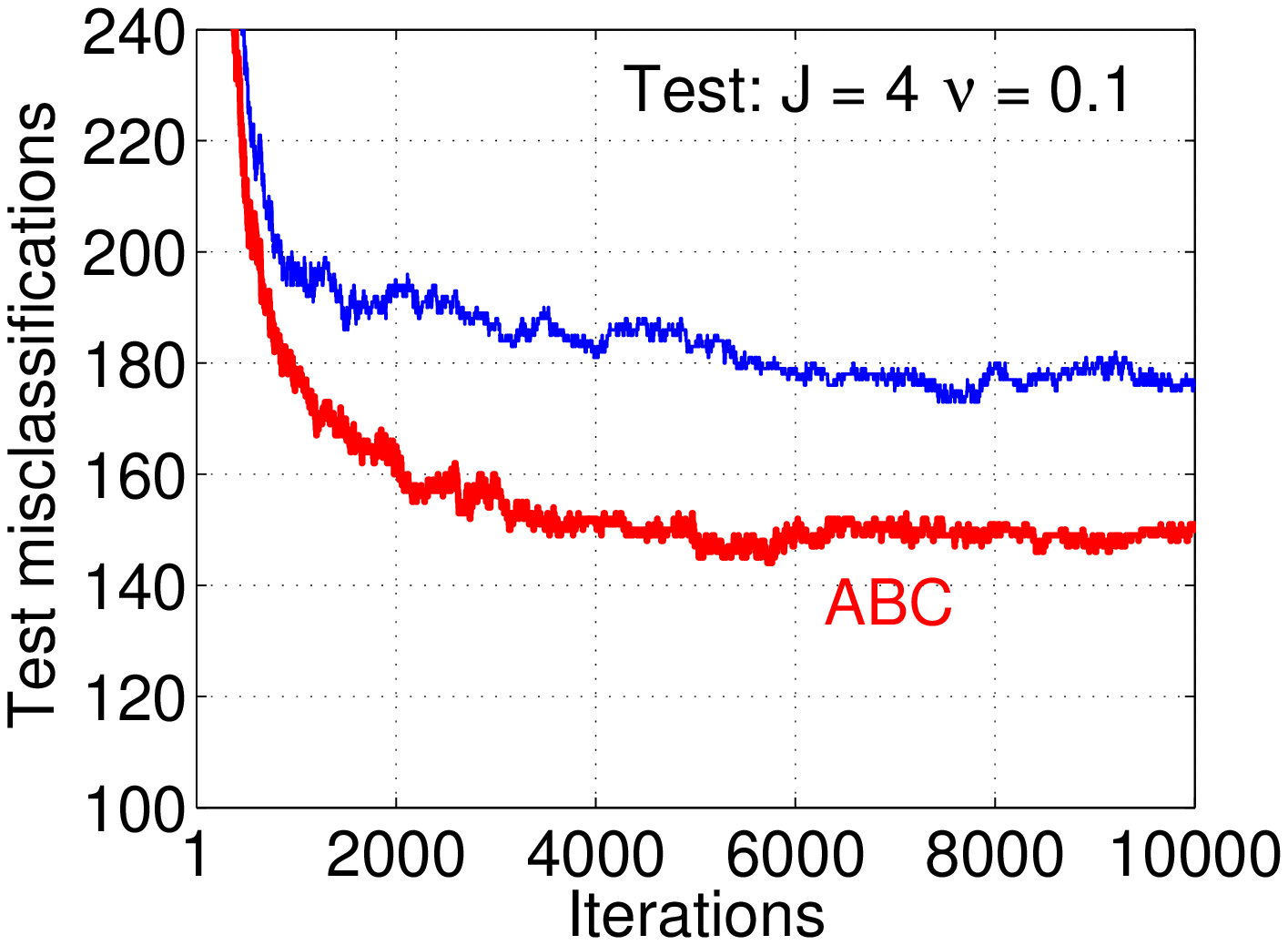}\hspace{0.1in}
{\includegraphics[width=2.0in]{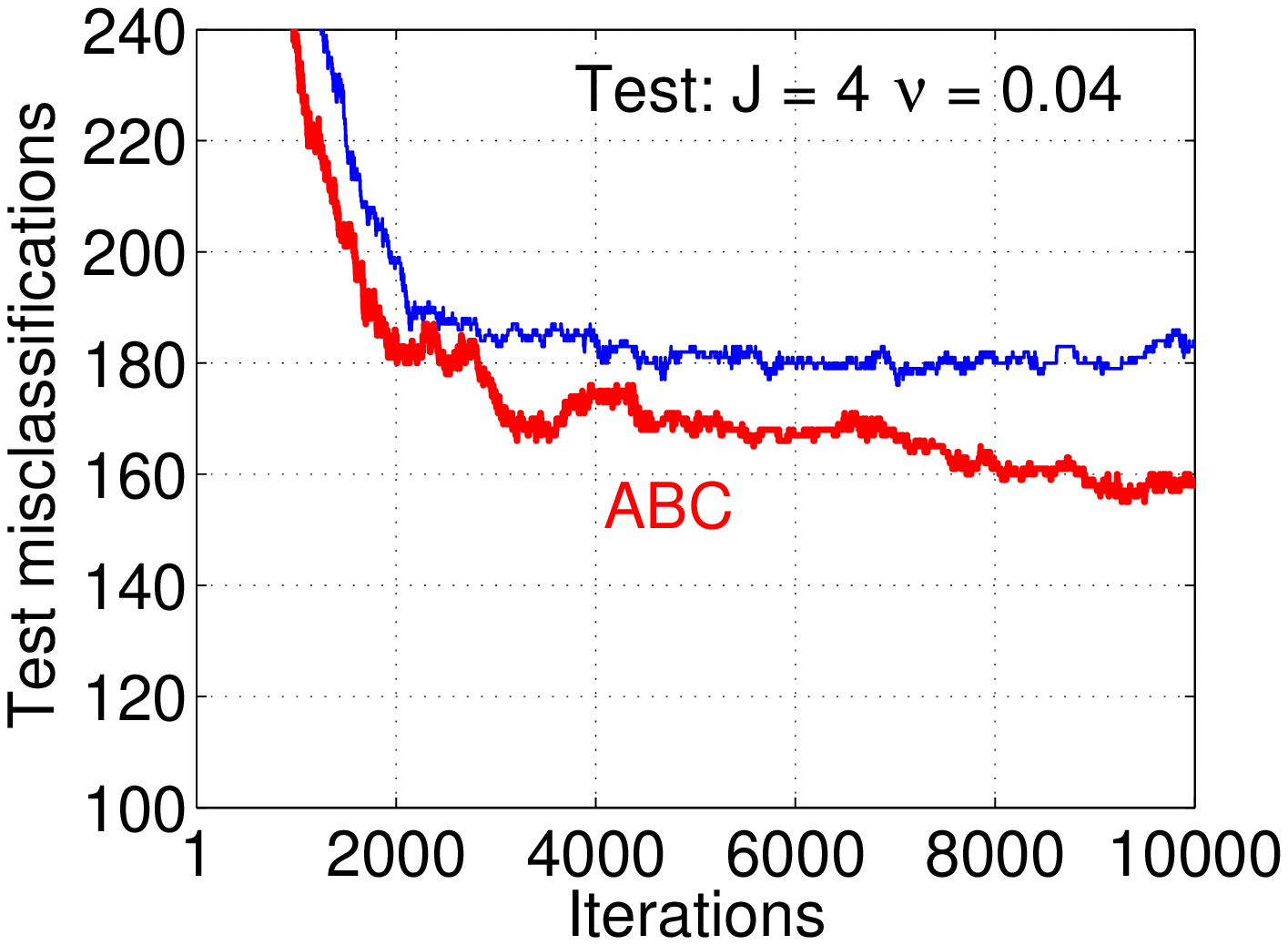}}}
\mbox{\includegraphics[width=2.0in]{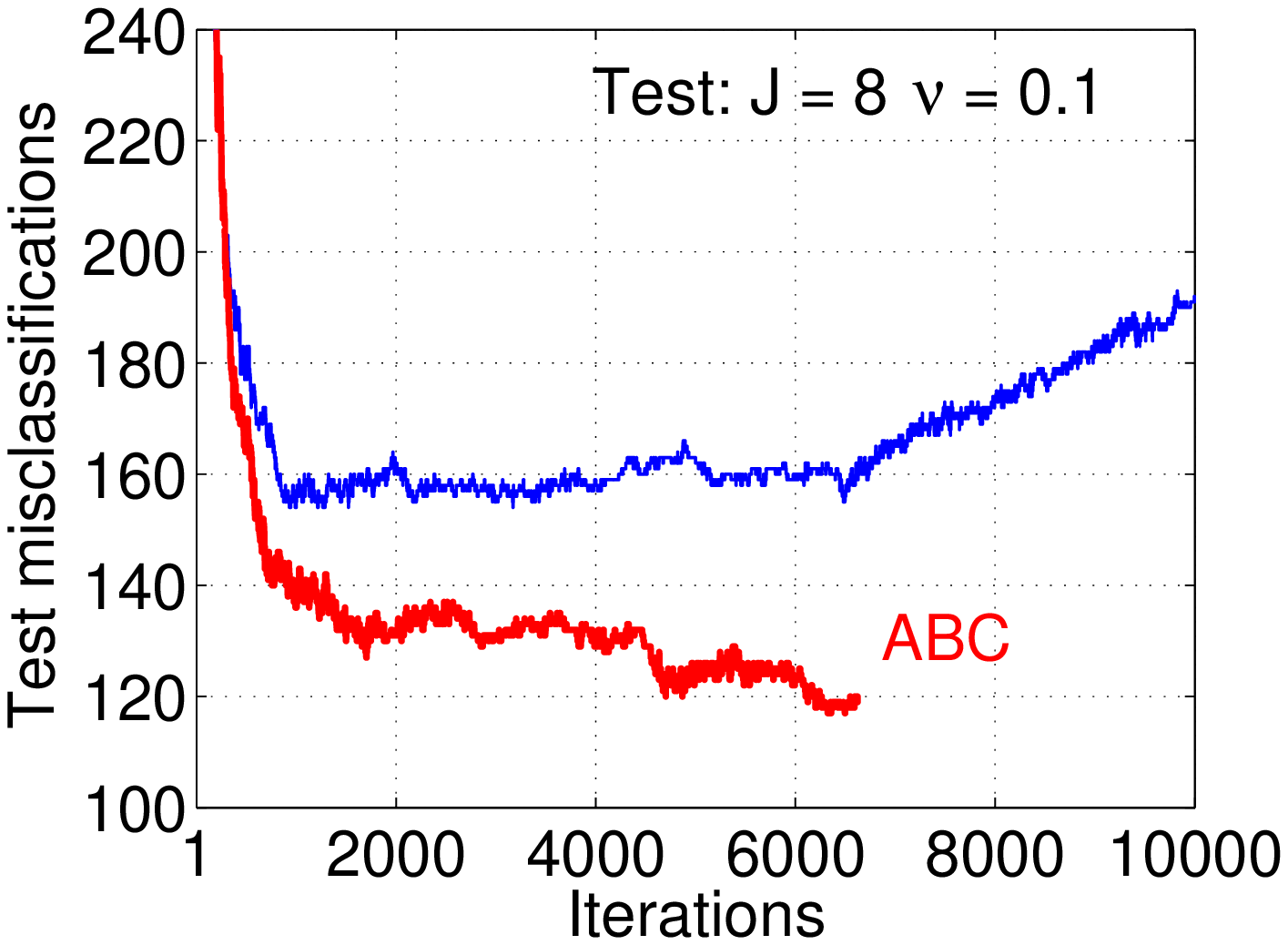}\hspace{0.1in}
{\includegraphics[width=2.0in]{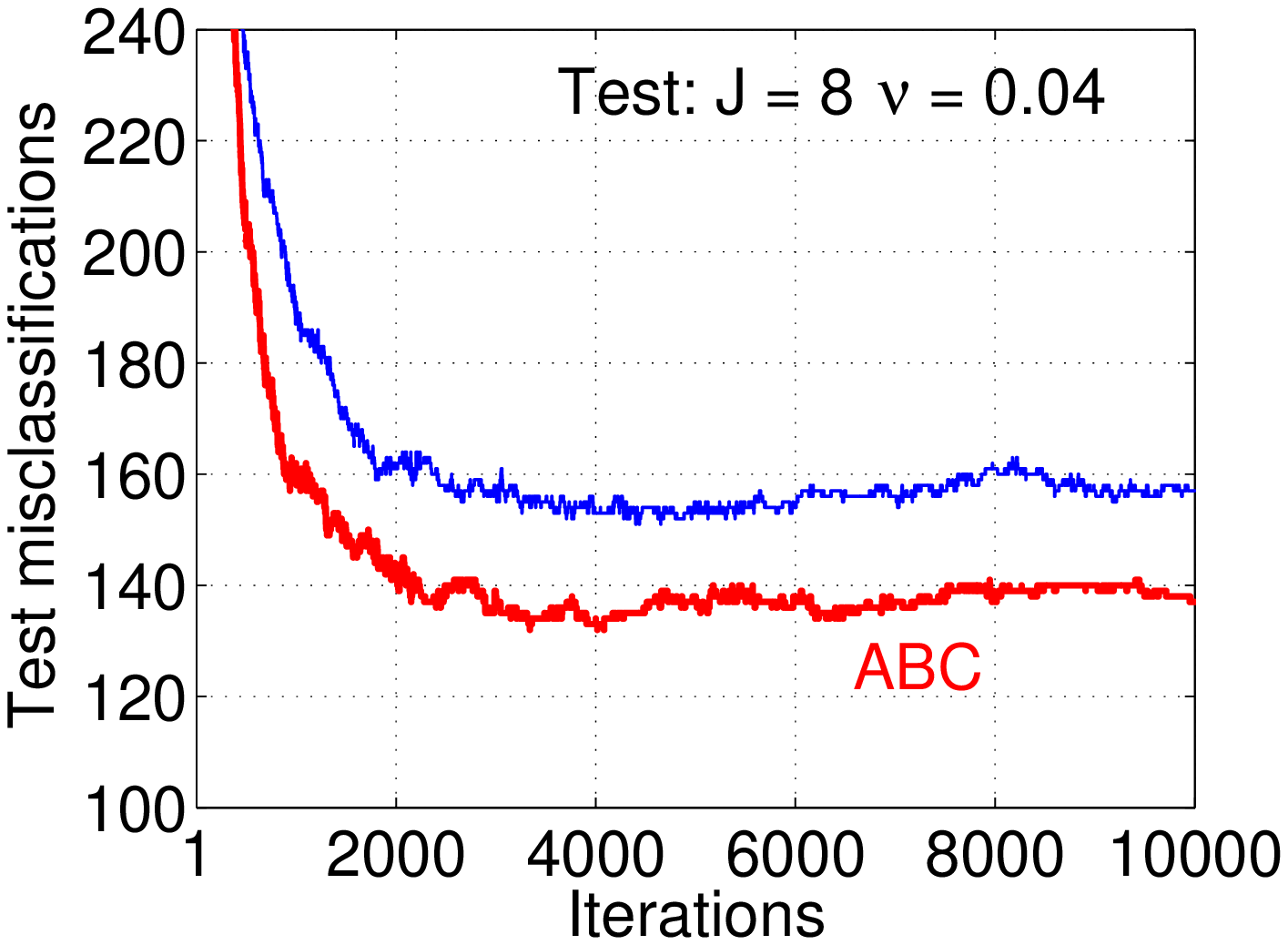}}}
\mbox{\includegraphics[width=2.0in]{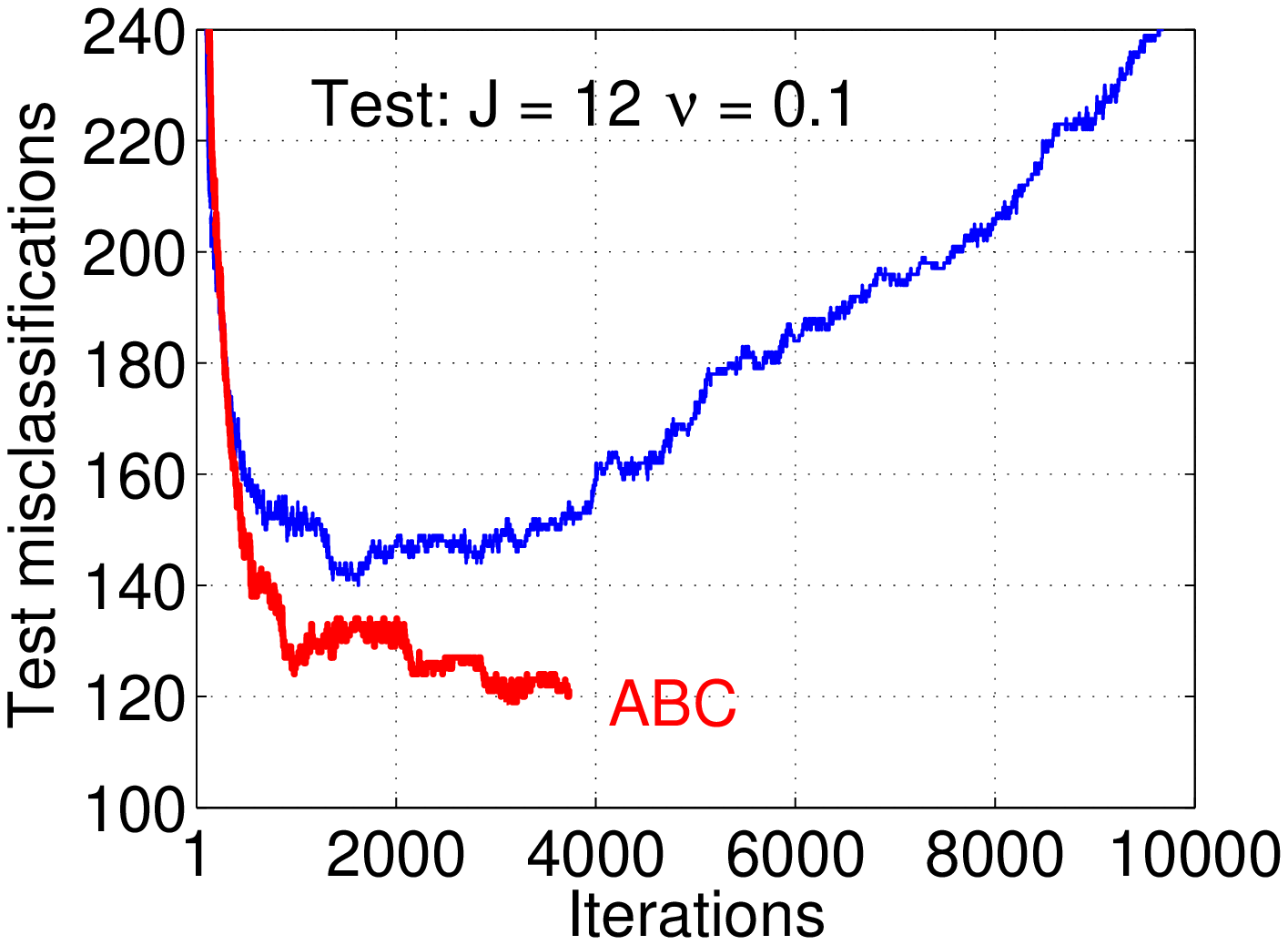}\hspace{0.1in}
{\includegraphics[width=2.0in]{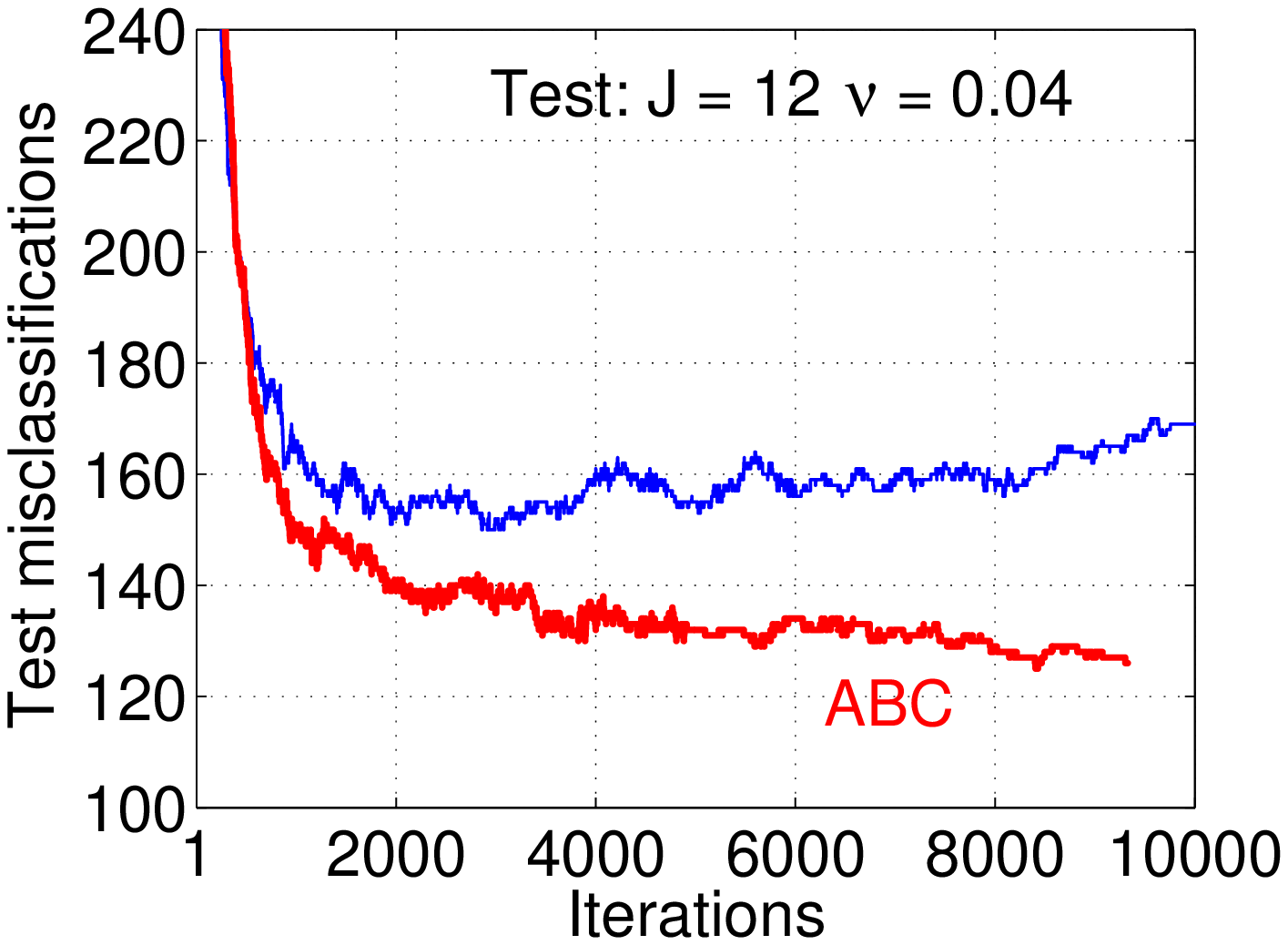}}}
\mbox{\includegraphics[width=2.0in]{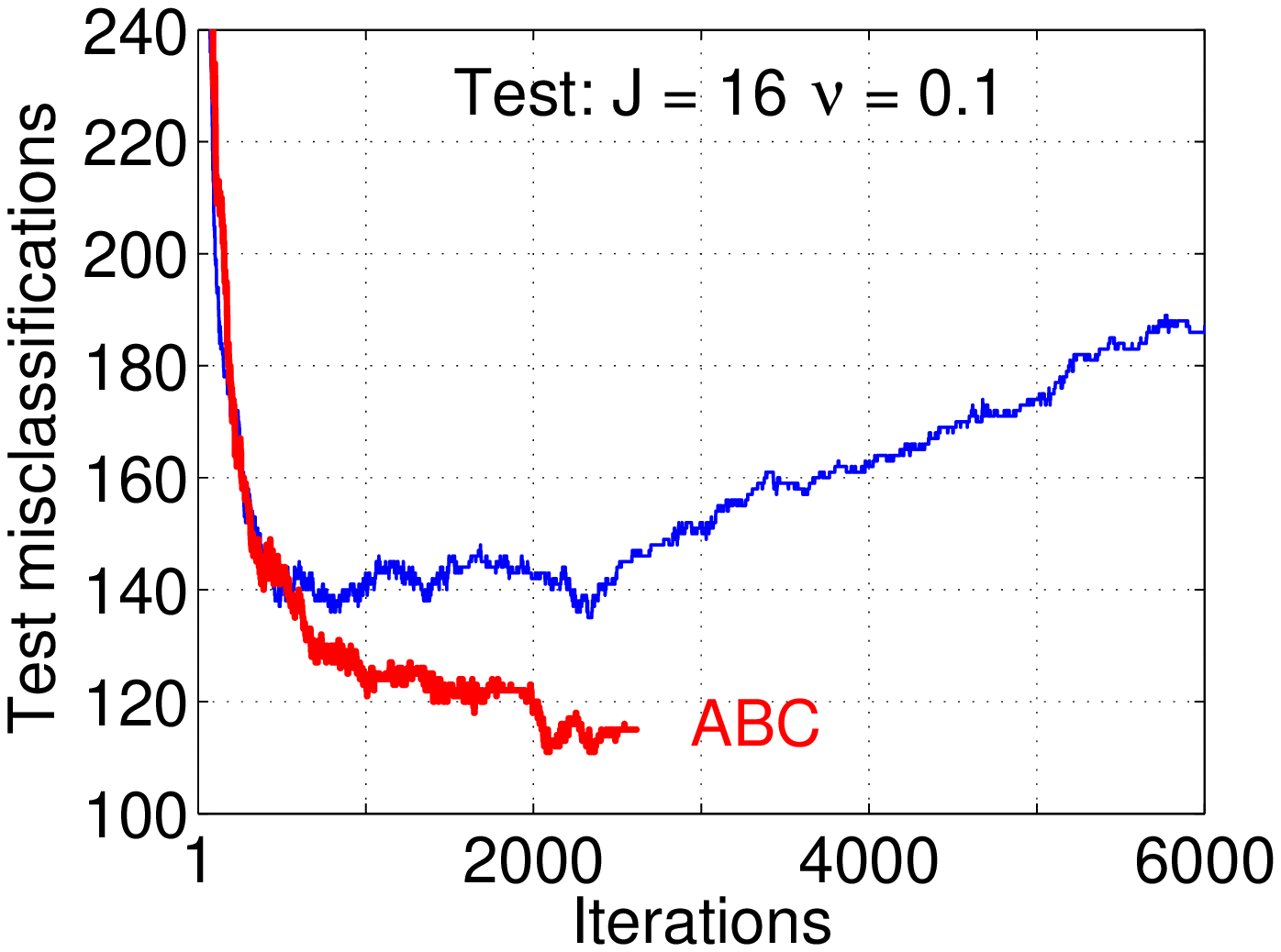}\hspace{0.1in}
{\includegraphics[width=2.0in]{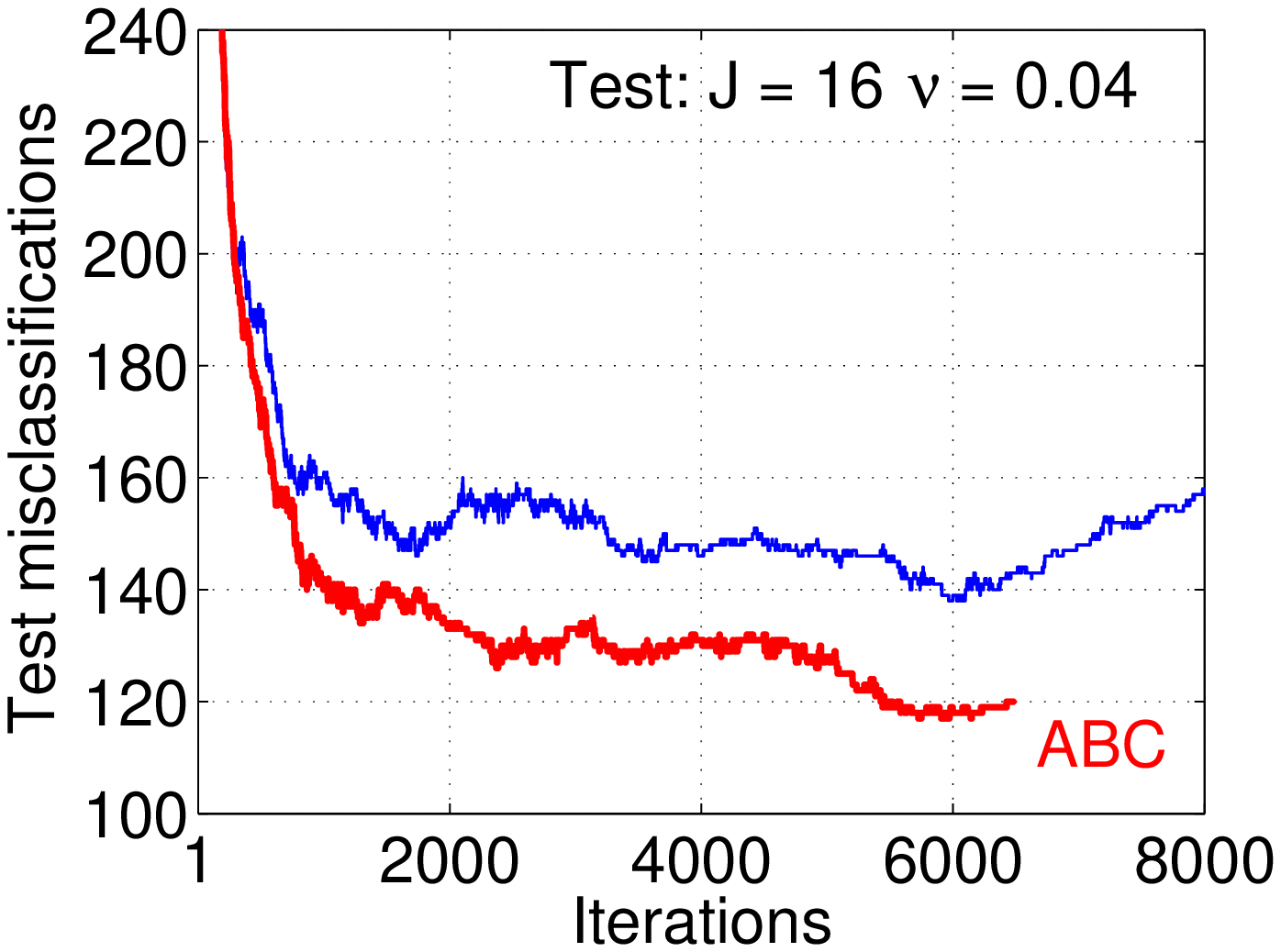}}}
\end{center}
\vspace{-0.2in}
\caption{\textbf{\em Letter}. The test mis-classification errors.}\label{fig_LetterTest}
\end{figure}

Figure \ref{fig_LetterTest} again demonstrates  that ABC-MART exhibits considerably and consistently smaller test mis-classification errors than MART.

\clearpage

\subsection{Experiments on the {\em Pendigits} Data Set}

\begin{table}[h]
\caption{\textbf{\em Pendigits}. The test mis-classification errors. The corresponding relative improvements ($R_{err}$, $\%$) of ABC-MART are included in parentheses.}
\begin{center}
{
\subtable[MART]{\begin{tabular}{l l l l l }
\hline \hline
  &$\nu = 0.04$ &$\nu=0.06$ &$\nu=0.08$ &$\nu=0.1$ \\
\hline
$J=4$   &144   &145  & 142  &148\\
$J=6$    &137   &134  & 136 &  135\\
$J=8$    &133   &134  & 132 &  128\\
$J=10$    &123  & 131  & 127 &  130\\
$J=12$    &135  & 136  & 134 &  133\\
$J=14$     &129  & 131  & 130 &  133\\
$J=16$    &132  & 129  & 130 &  134\\
\hline\hline
\end{tabular}}

\subtable[ABC-MART]{\begin{tabular}{l l l l l }
\hline \hline
  &$\nu = 0.04$ &$\nu=0.06$ &$\nu=0.08$ &$\nu=0.1$ \\
\hline
$J=4$  &112\   (22.2)  &109 \ (24.8)  &110 \  (22.5)  &112 \  (24.3)\\
$J=6$  &114\   (16.8)  &114 \ (14.9)  &111 \  (18.4)  &109 \  (19.3)\\
$J=8$  &112\   (15.8)  &107 \ (20.1)  &111 \  (15.9)  &105 \  (18.0)\\
$J=10$ &105\   (14.6)  &105 \ (19.8)  &107 \  (15.7)  &109 \  (16.2)\\
$J=12$ &104\   (23.0)  &106 \ (22.1)  &104 \  (22.4)  &105 \  (21.1)\\
$J=14$ &107\   (17.1)  &106 \ (19.1)  &104 \  (20.0)  &107 \  (19.5)\\
$J=16$ &106\   (19.7)  &107 \ (17.1)  &106 \  (18.5)  &107 \  (20.1)\\
\hline\hline
\end{tabular}}
}
\end{center}
\label{tab_Pendigits}
\end{table}

\begin{figure}[h]
\begin{center}
\mbox{\includegraphics[width=2.0in]{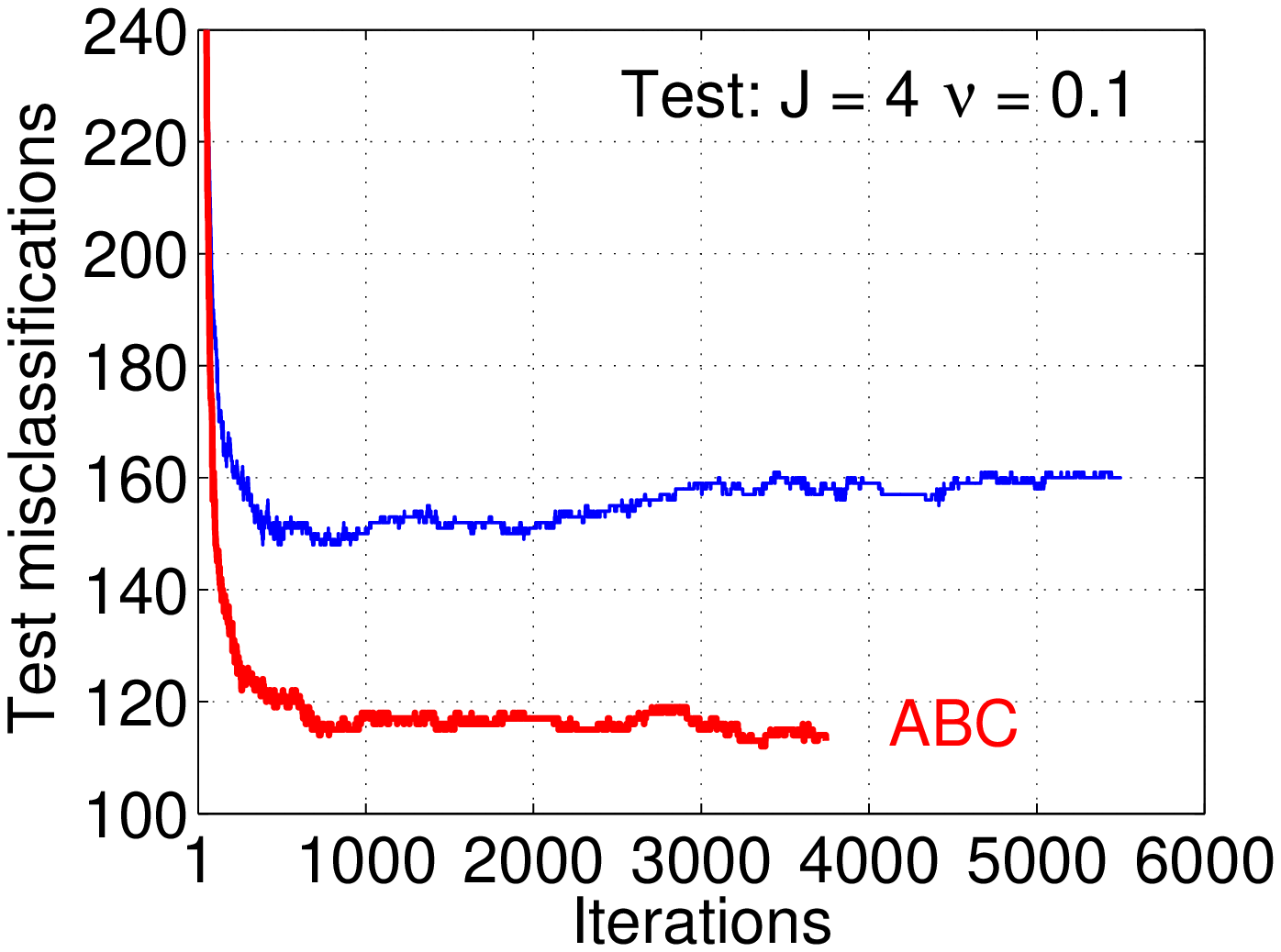}\hspace{0.1in}
{\includegraphics[width=2.0in]{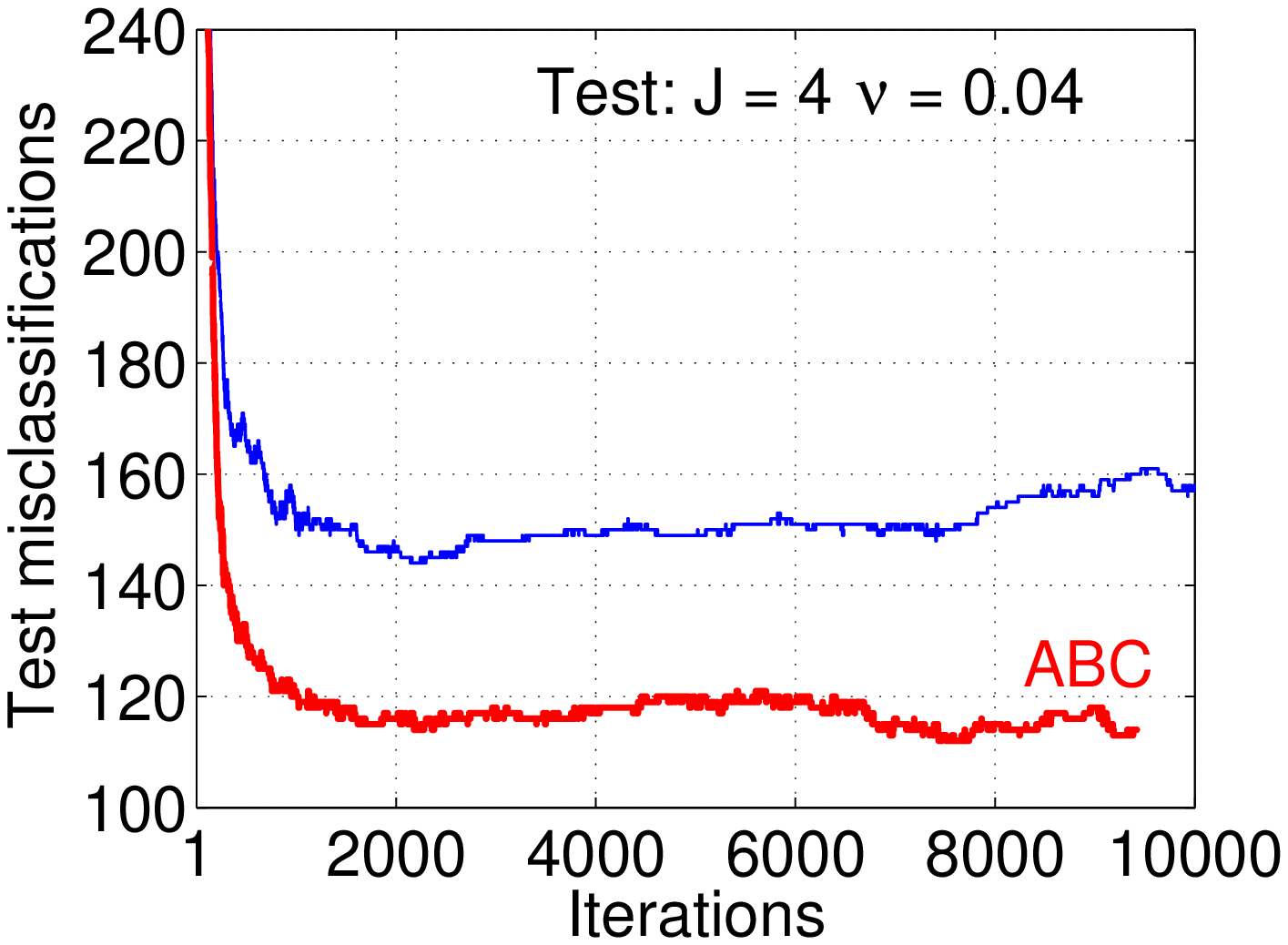}}}
\mbox{\includegraphics[width=2.0in]{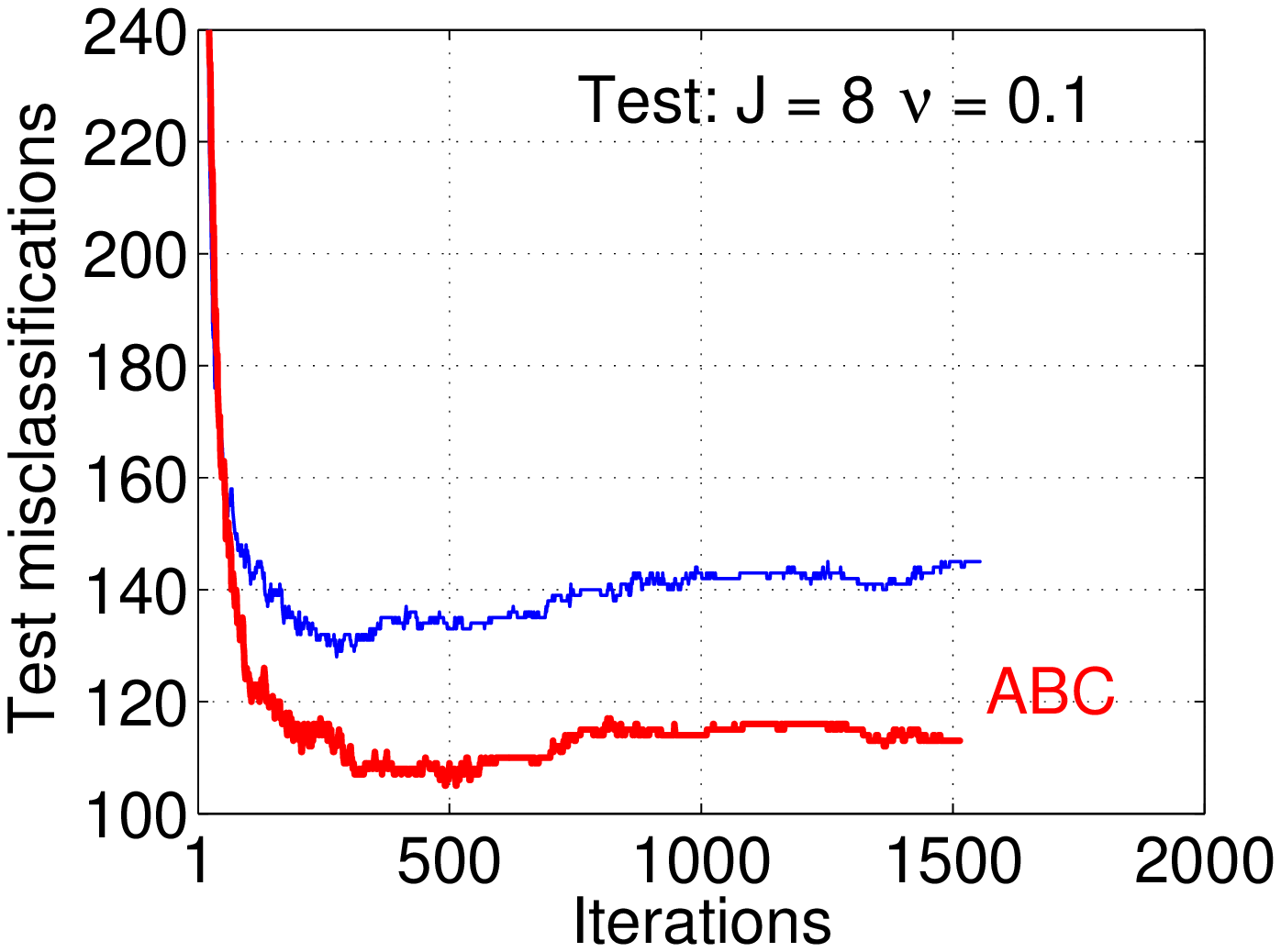}\hspace{0.1in}
{\includegraphics[width=2.0in]{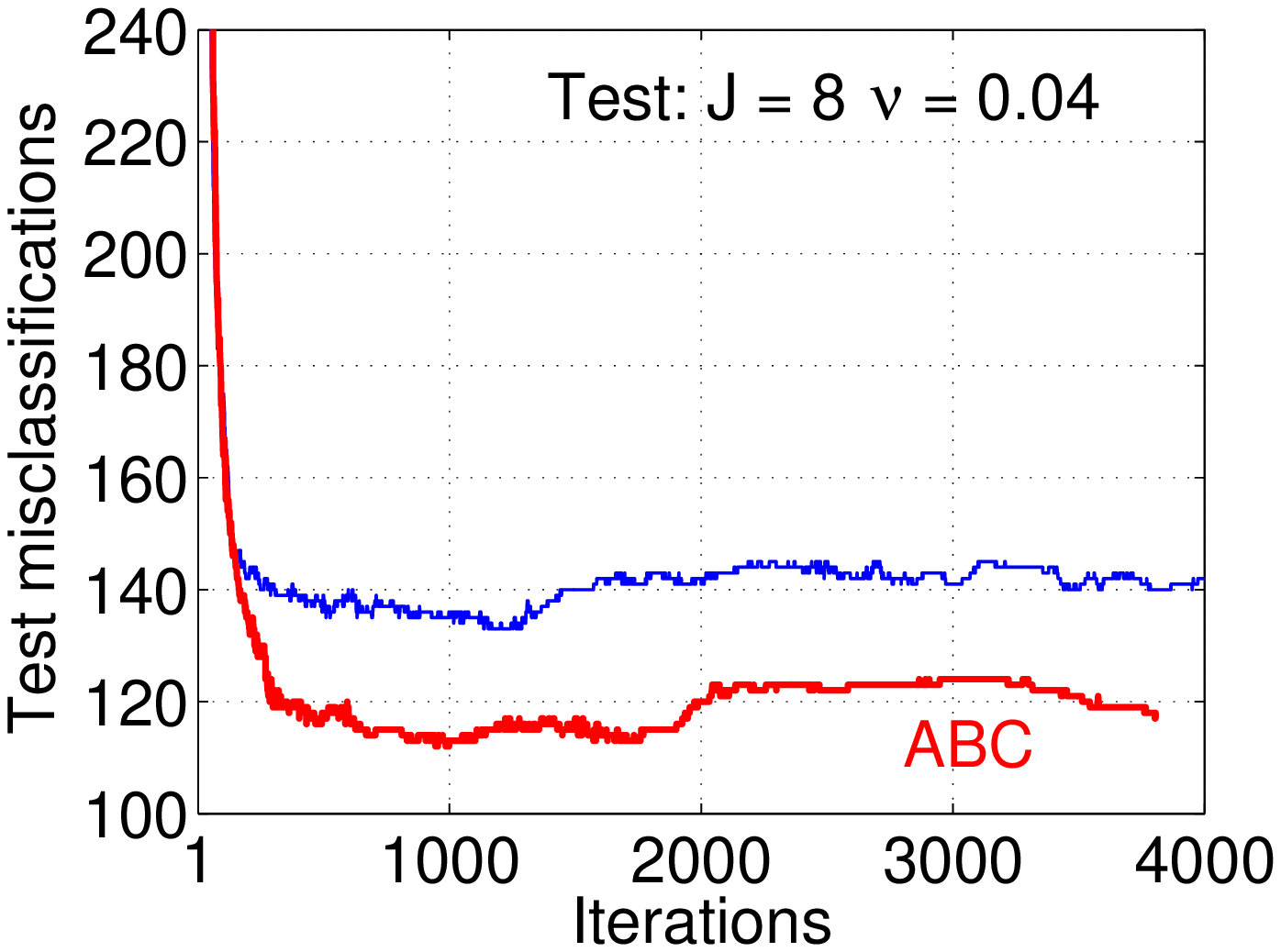}}}
\mbox{\includegraphics[width=2.0in]{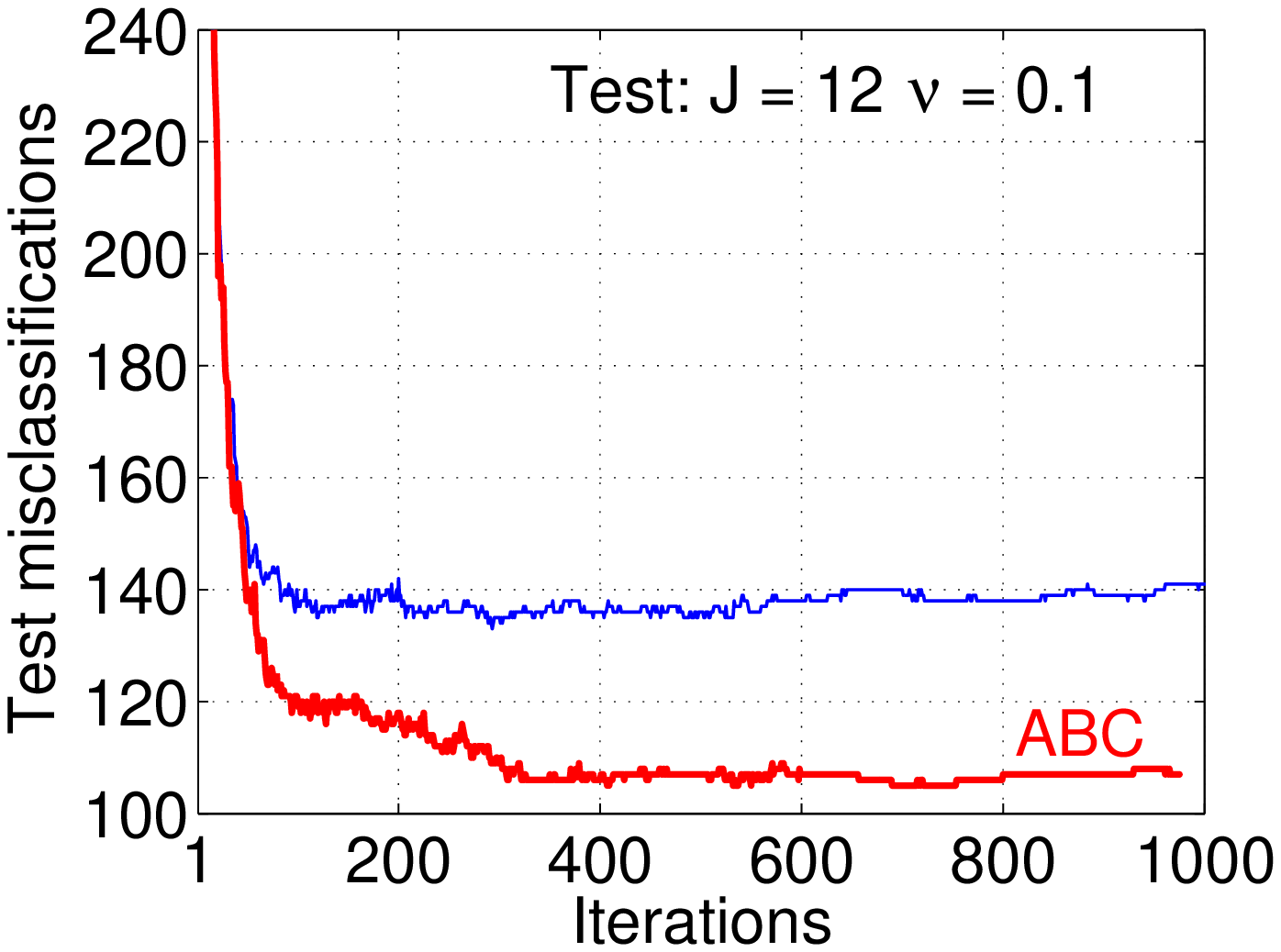}\hspace{0.1in}
{\includegraphics[width=2.0in]{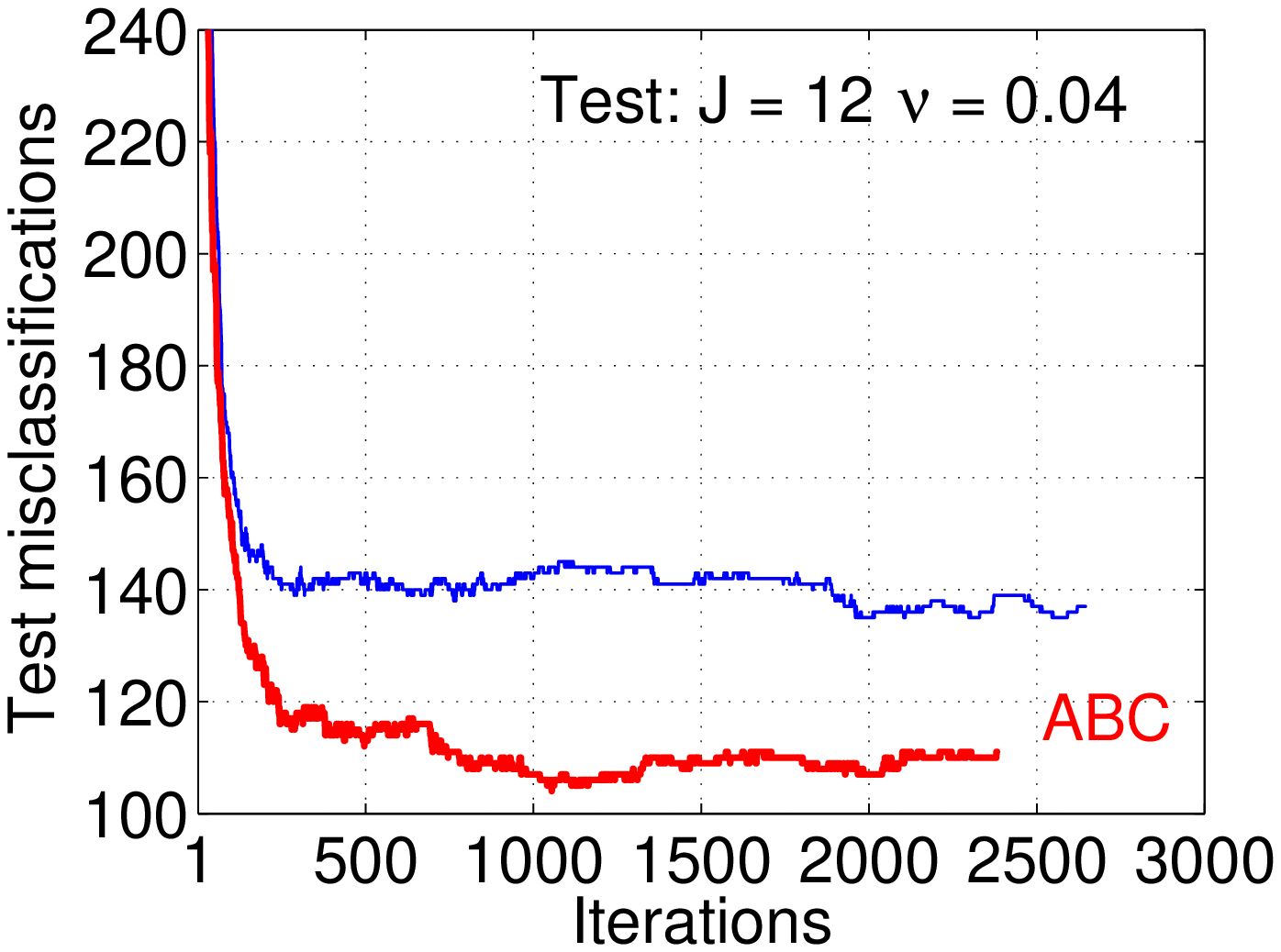}}}
\mbox{\includegraphics[width=2.0in]{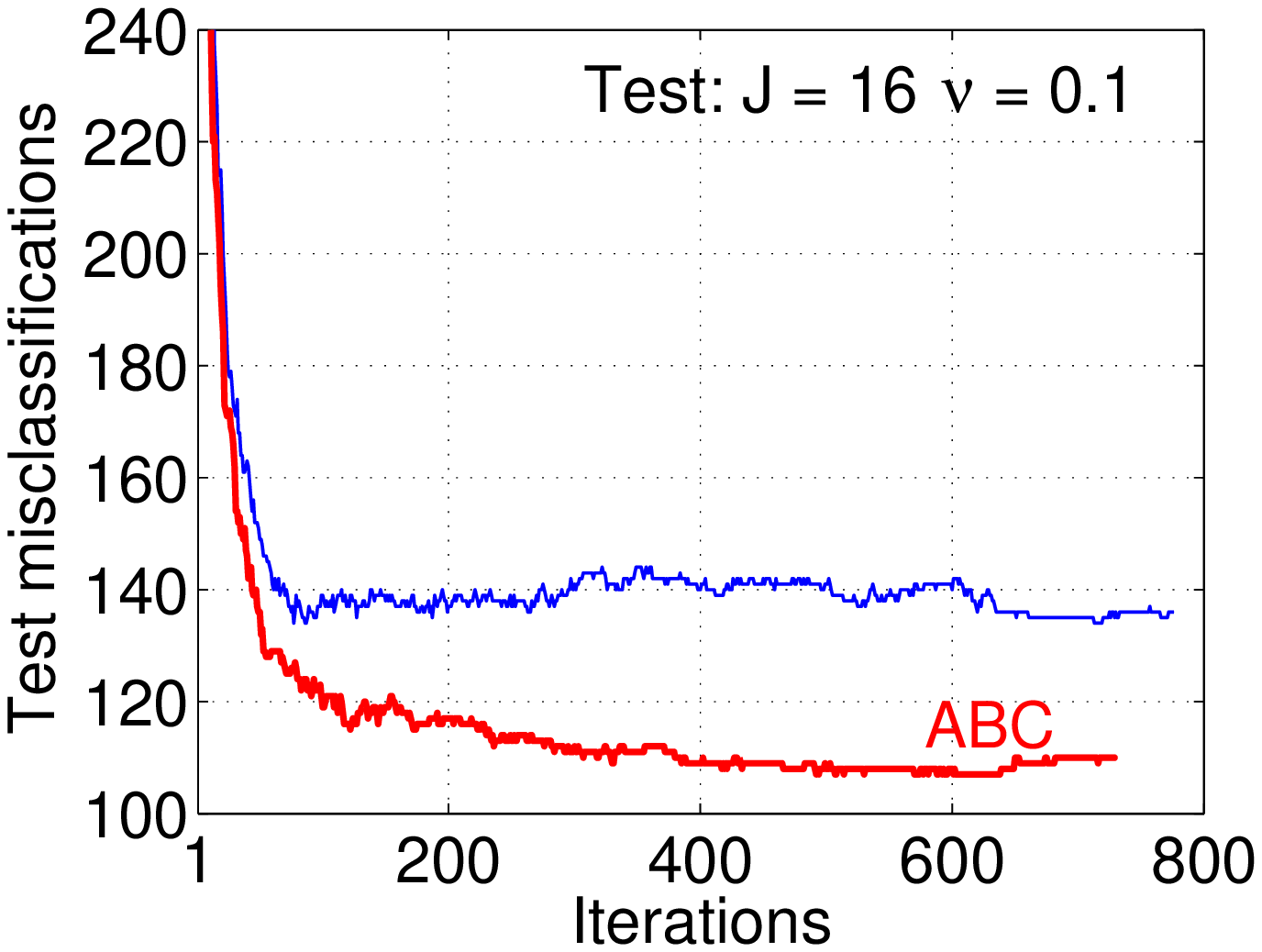}\hspace{0.1in}
{\includegraphics[width=2.0in]{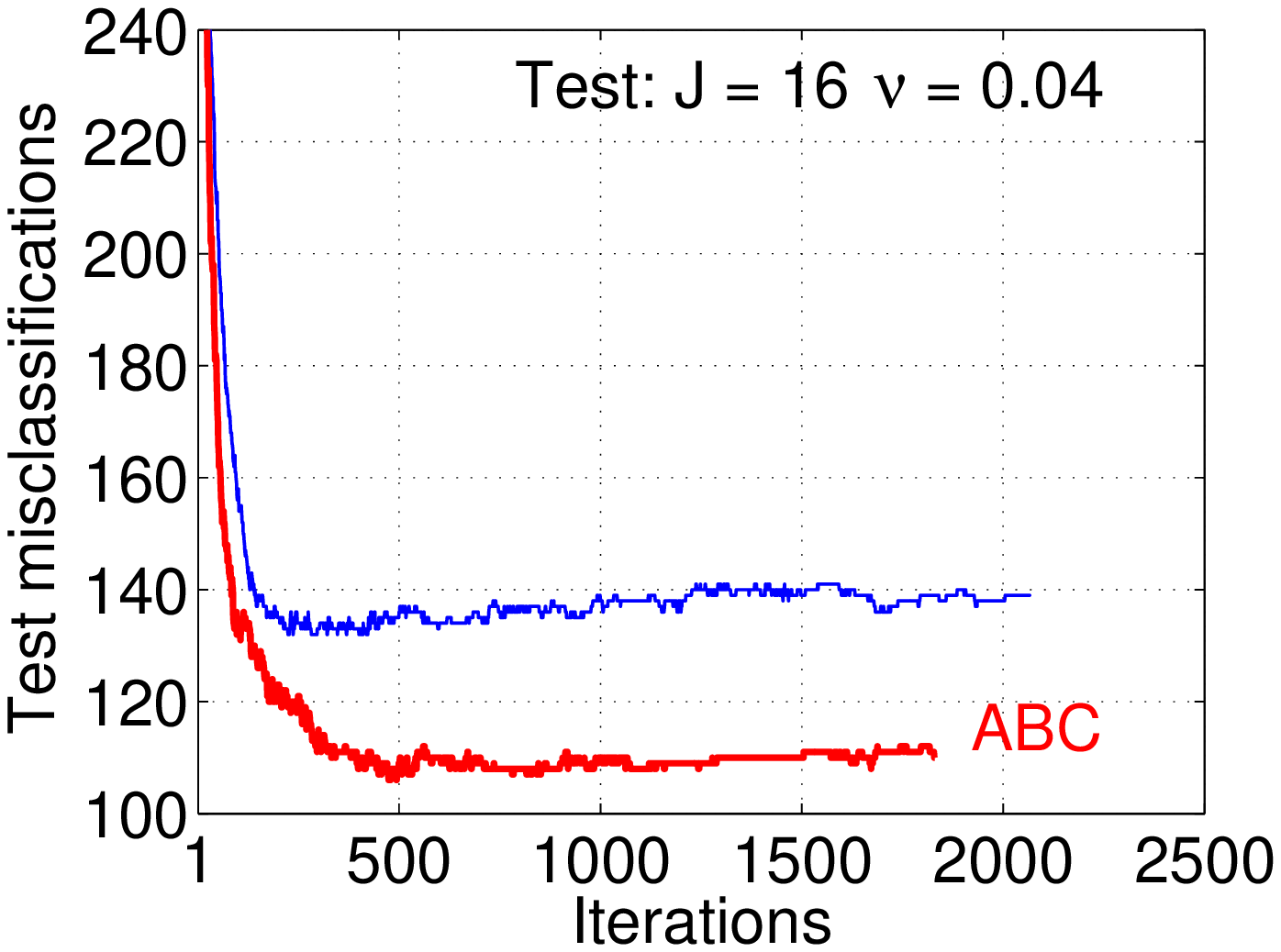}}}
\end{center}
\vspace{-0.2in}
\caption{\textbf{\em Pendigits}. The test mis-classification errors. }\label{fig_PendigitsTest}
\end{figure}

\clearpage

\subsection{Experiments on the {\em Zipcode} Data Set}

\begin{table}[h]
\caption{\textbf{\em Zipcode}. The test mis-classification errors.}
\begin{center}
{
\subtable[MART]{\begin{tabular}{l l l l l }
\hline \hline
  &$\nu = 0.04$ &$\nu=0.06$ &$\nu=0.08$ &$\nu=0.1$ \\
\hline
$J=4$   &130   &126   &129  & 127\\
$J=6$   &123  & 122  & 123  & 126\\
$J=8$   &120  & 122  & 121 &  123\\
$J=10$  & 118  & 118  & 119 &  118\\
$J=12$  & 117  & 117  & 116  & 118\\
$J=14$  &  119 &  120  & 119 &  117\\
$J=16$  & 118  & 111  & 116 &  116\\
\hline\hline
\end{tabular}}

\subtable[ABC-MART]{\begin{tabular}{l l l l l }
\hline \hline
  &$\nu = 0.04$ &$\nu=0.06$ &$\nu=0.08$ &$\nu=0.1$ \\
\hline
$J=4$  &116  \ (10.8)  &114\   (9.5)  &116\   (10.1)  &111  \ (12.6)\\
$J=6$  &110\   (10.6)  &111\   (9.0)  &109\   (11.4)  &108  \ (14.3)\\
$J=8$   &109\    (9.2)  &102\  (16.4)  &111\    (8.3)  &106 \  (13.8)\\
$J=10$  &106\   (10.2)  &103\  (12.7)  &103\   (13.4)  &104 \  (11.9)\\
$J=12$  &105\   (10.3)  &102\  (12.8)  &103\   (11.2) &  98 \ \ \ (16.9)\\
$J=14$  &106\   (10.9)  &104\   (13.3)  &103\  (13.4)  &105 \  (10.3)\\
$J=16$  &104\   (11.9)  &103\   (7.2)  &102\  (12.1)  &104  \ (10.3)\\
\hline\hline
\end{tabular}}
}
\end{center}
\label{tab_Zipcode}
\end{table}

\begin{figure}[h]
\begin{center}
\mbox{\includegraphics[width=2.0in]{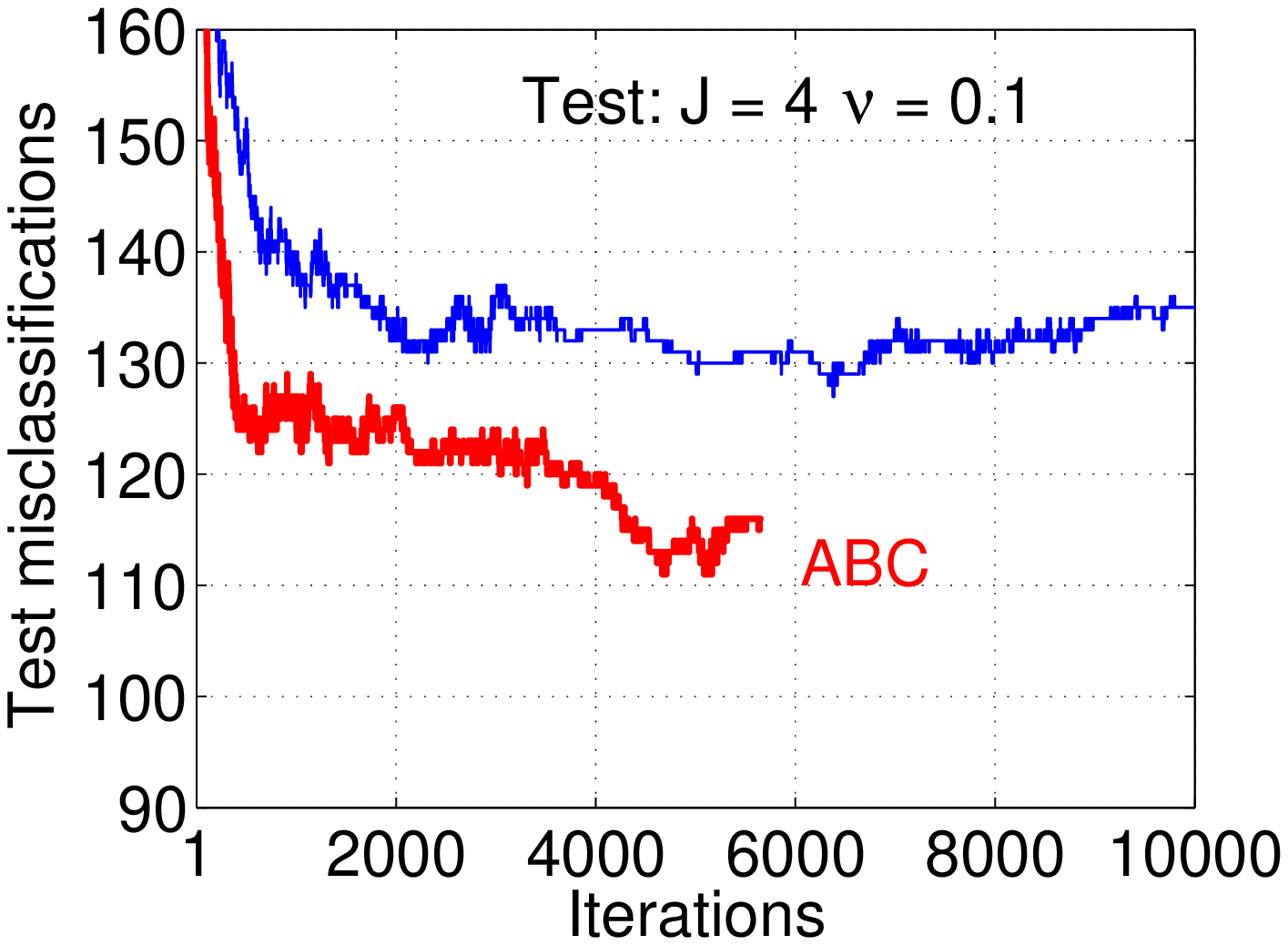}\hspace{0.1in}
{\includegraphics[width=2.0in]{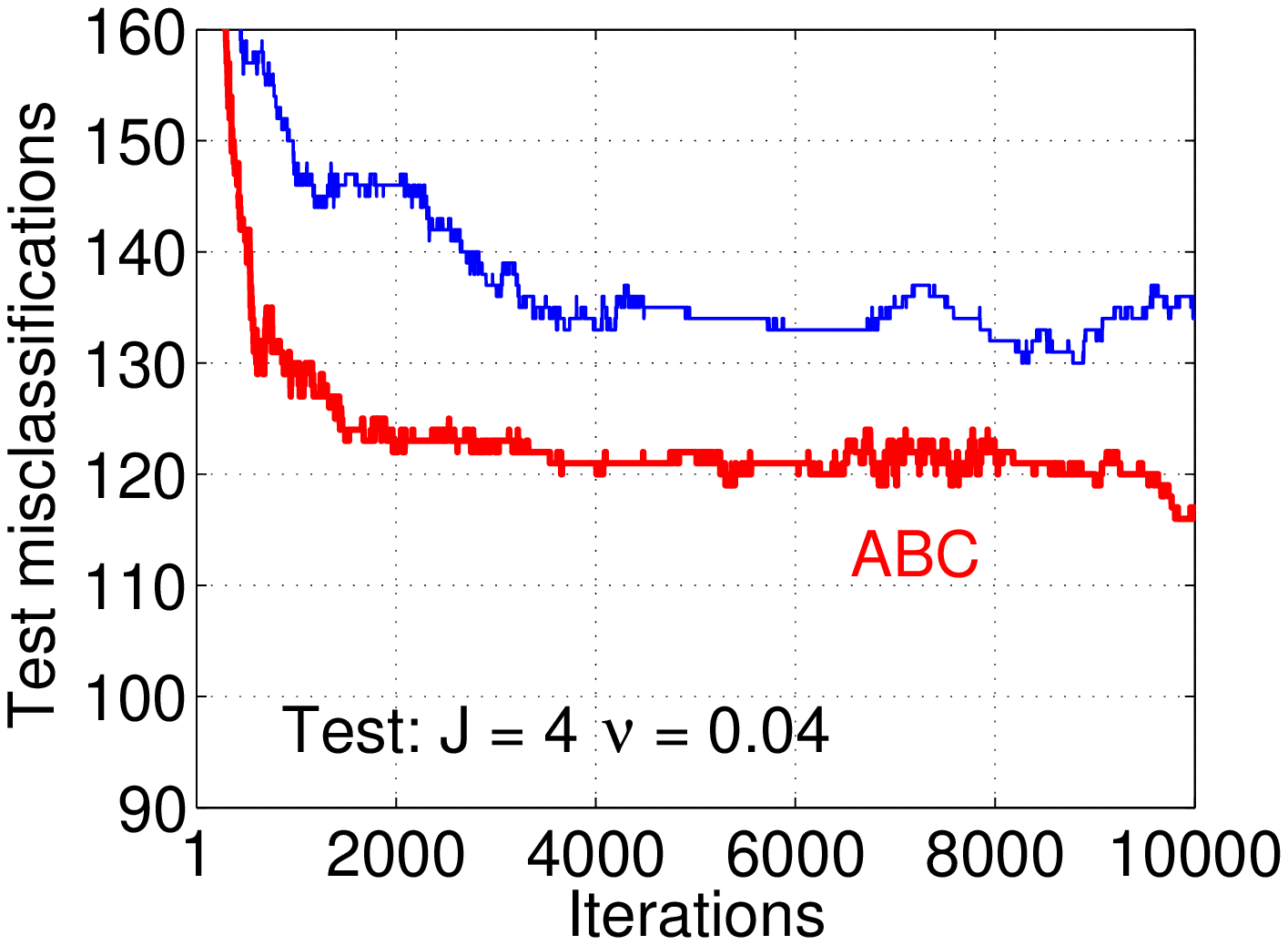}}}
\mbox{\includegraphics[width=2.0in]{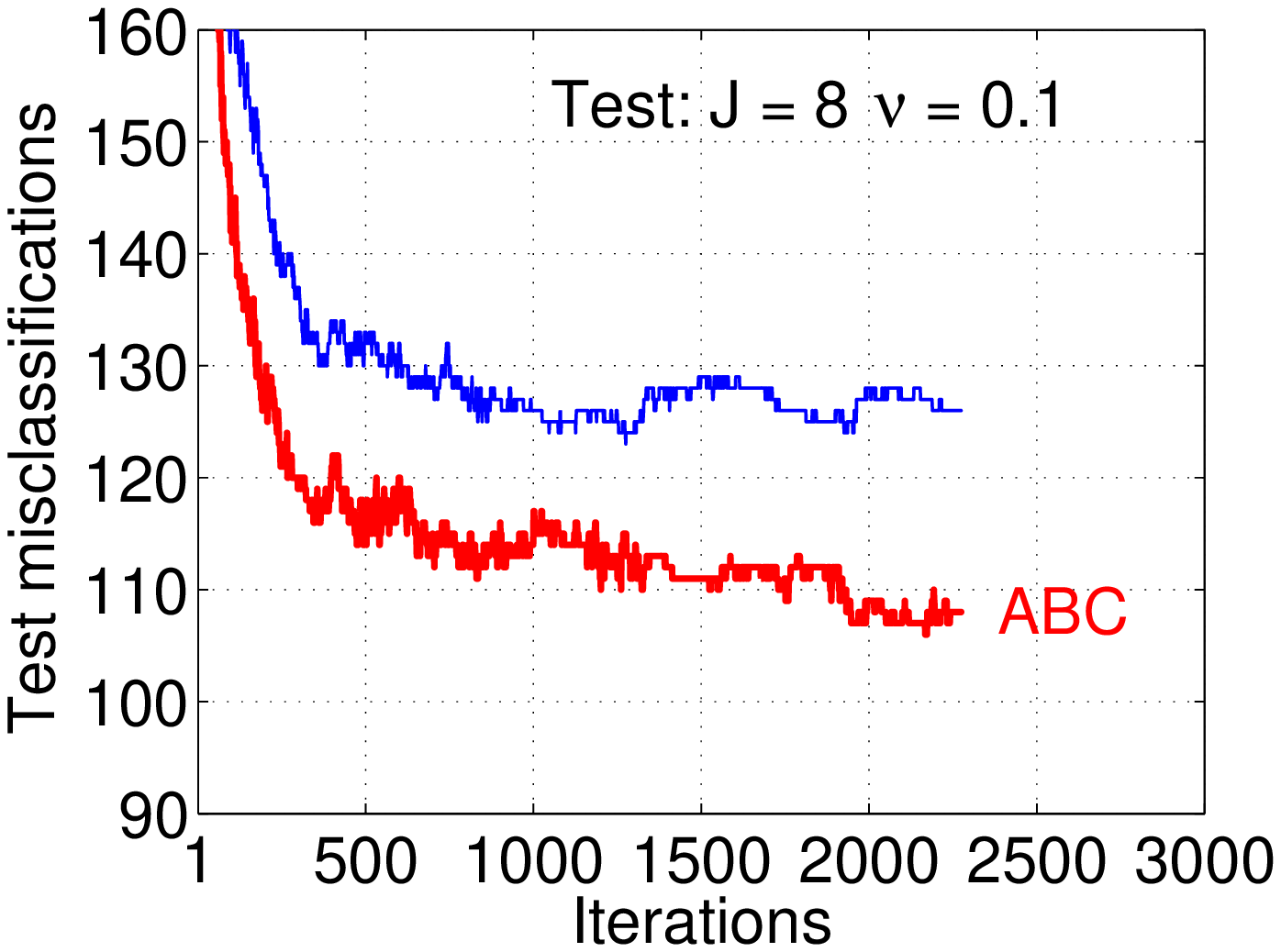}\hspace{0.1in}
{\includegraphics[width=2.0in]{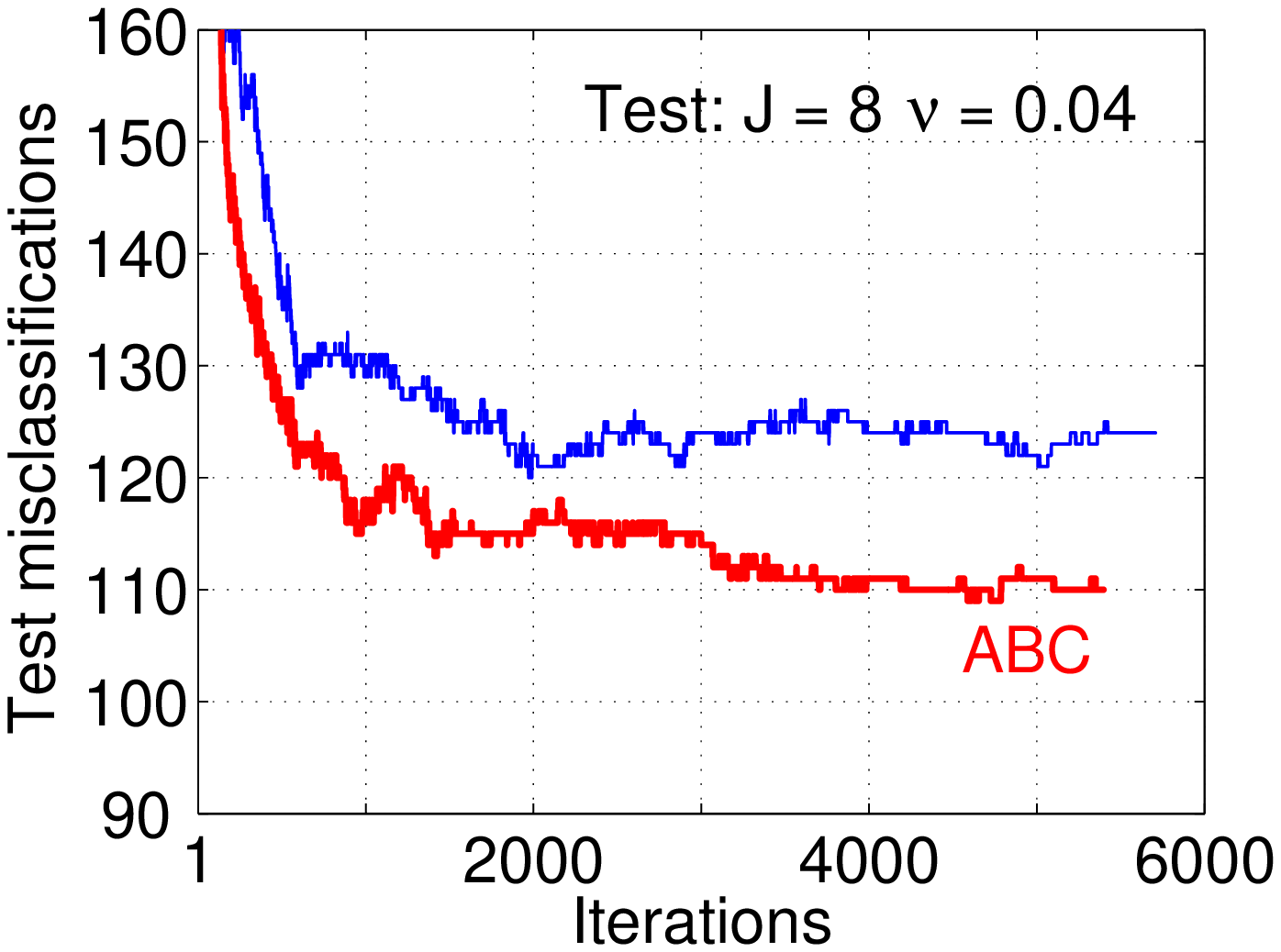}}}
\mbox{\includegraphics[width=2.0in]{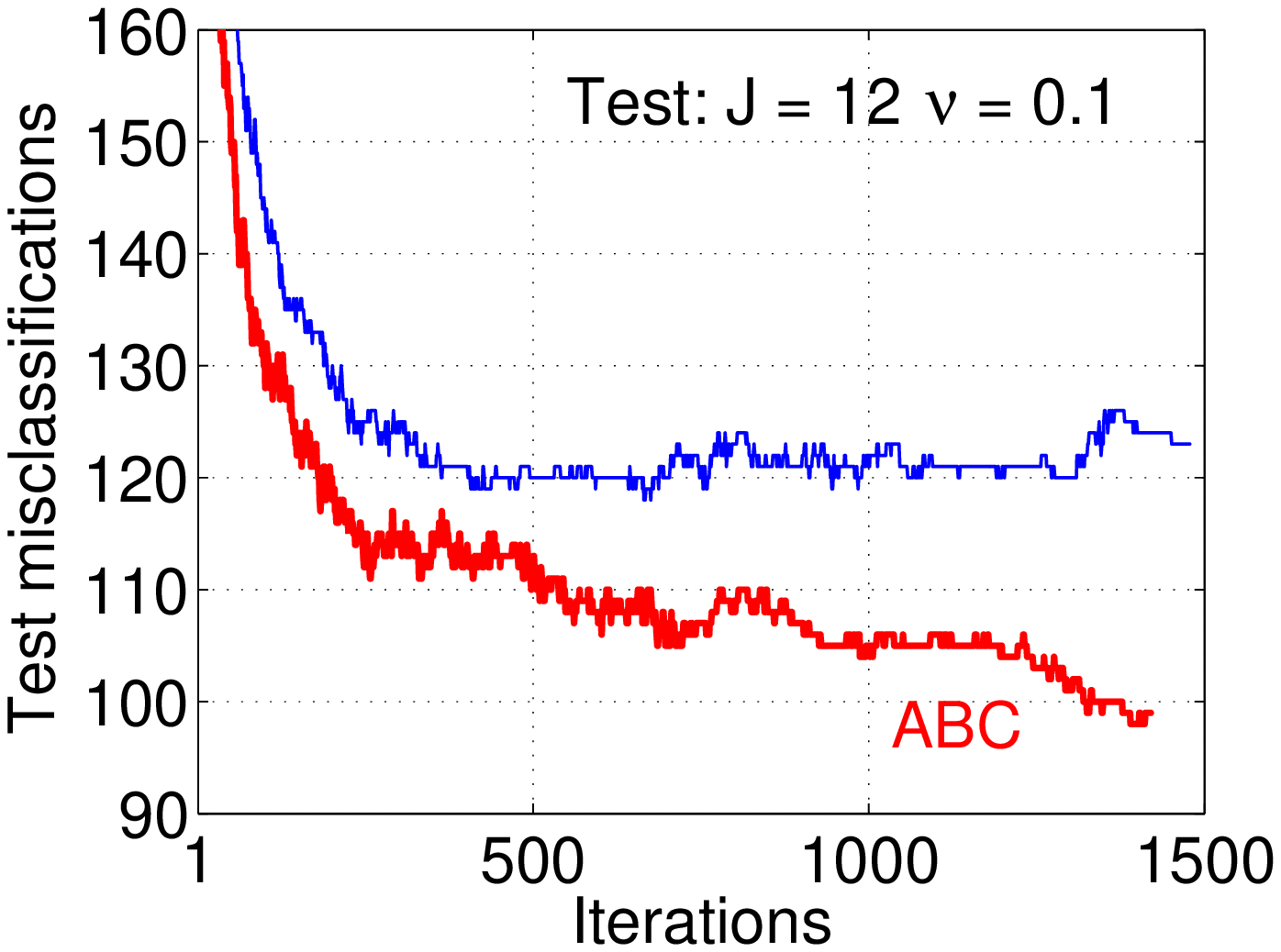}\hspace{0.1in}
{\includegraphics[width=2.0in]{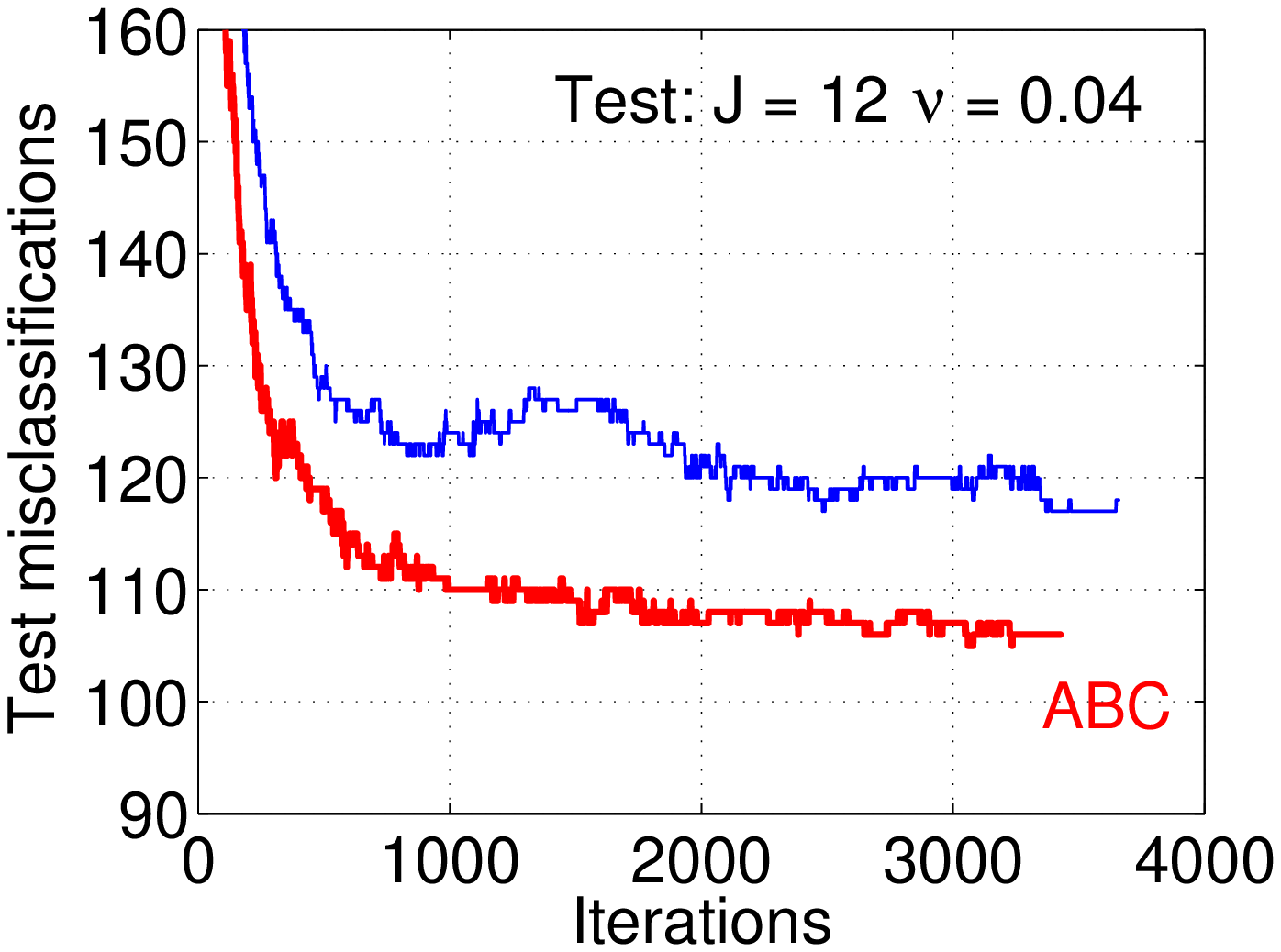}}}
\mbox{\includegraphics[width=2.0in]{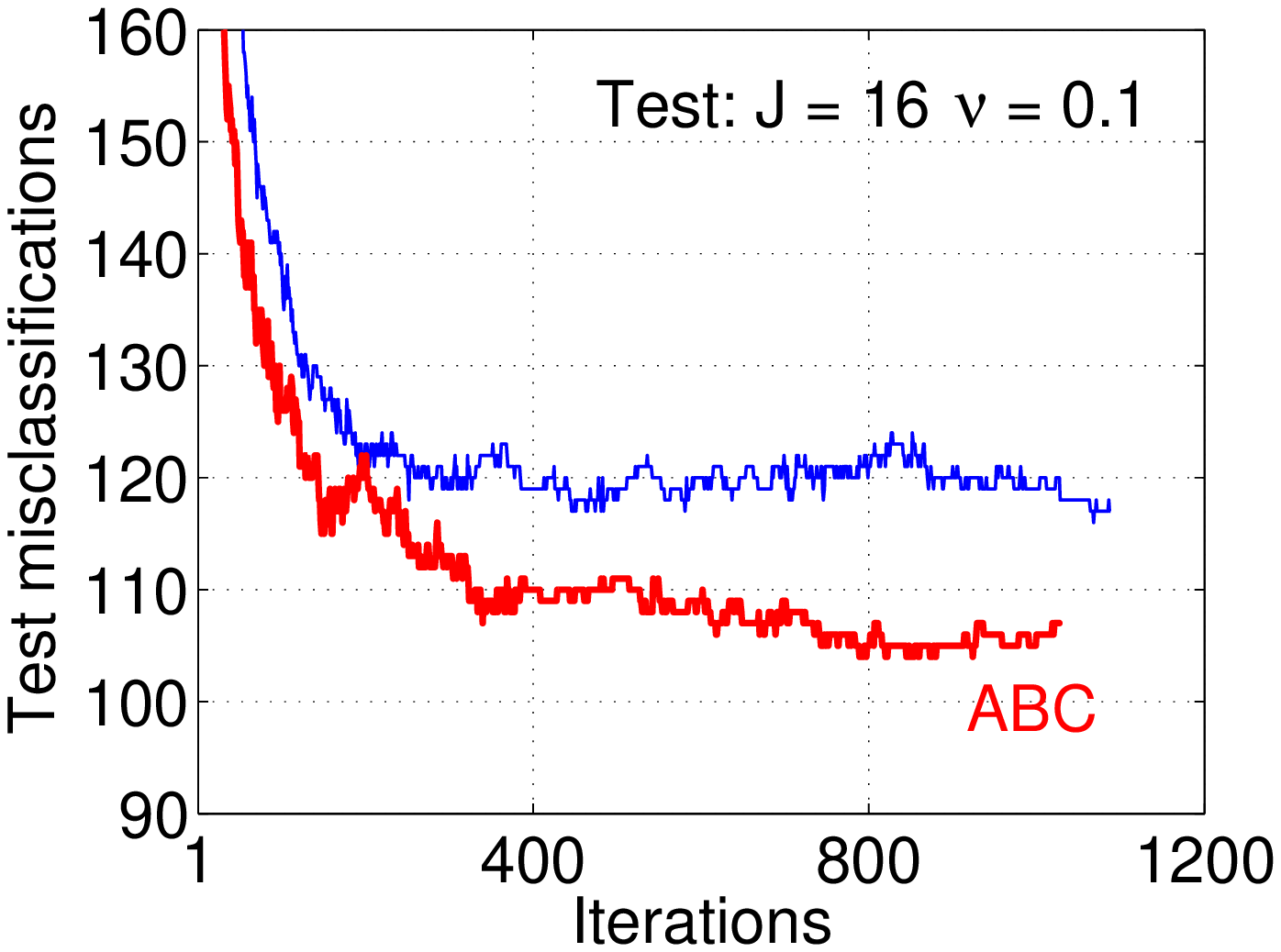}\hspace{0.1in}
{\includegraphics[width=2.0in]{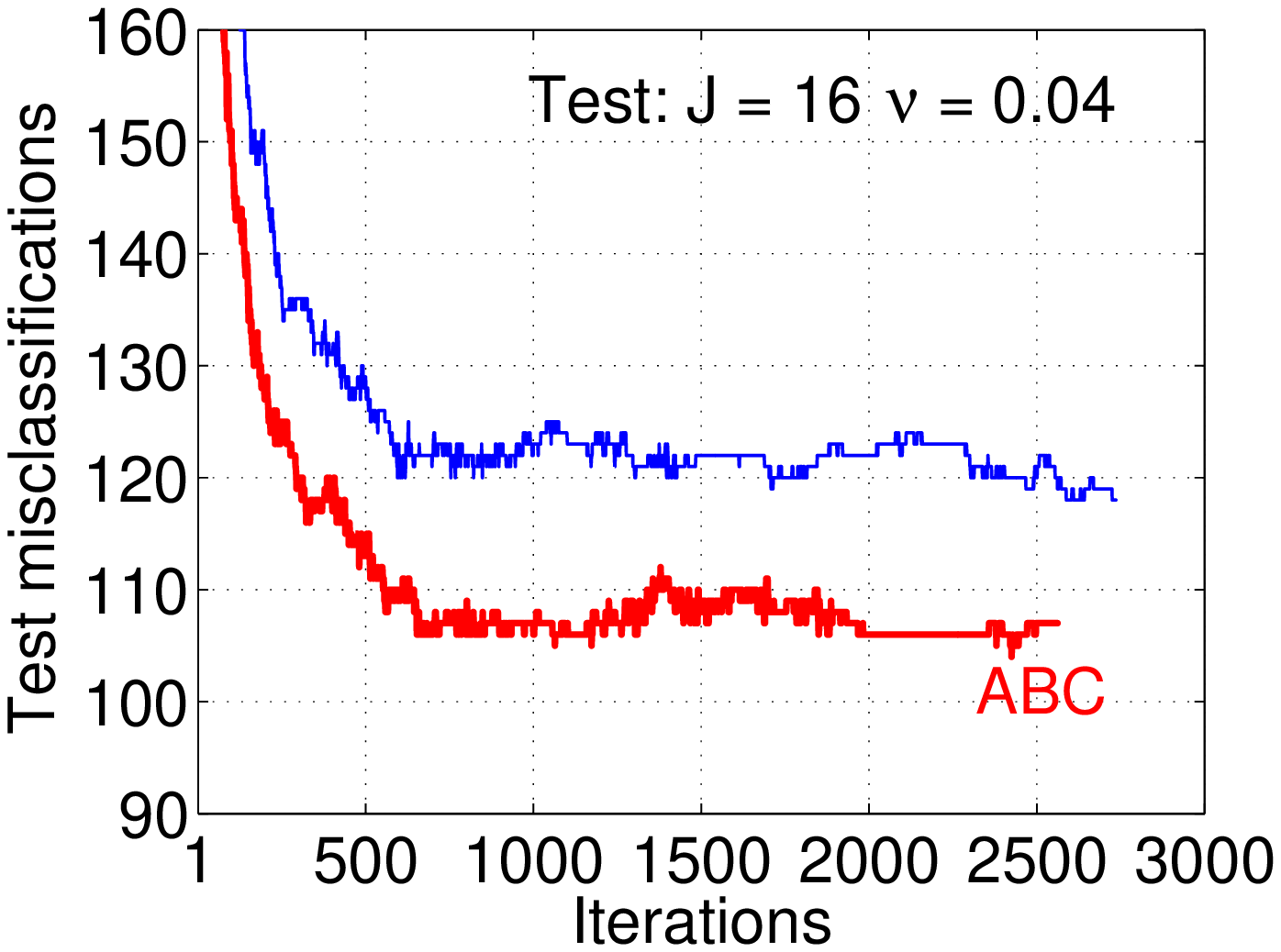}}}
\end{center}
\vspace{-0.2in}
\caption{\textbf{\em Zipcode}. The test mis-classification errors.}\label{fig_ZipcodeTest}
\end{figure}

\clearpage

\subsection{Experiments on the {\em Optdigits} Data Set}

\begin{table}[h]
\caption{\textbf{\em Optdigits}. The test mis-classification errors}.
\begin{center}
{
\subtable[MART]{\begin{tabular}{l l l l l }
\hline \hline
  &$\nu = 0.04$ &$\nu=0.06$ &$\nu=0.08$ &$\nu=0.1$ \\
\hline
$J=4$    &58    &57    &57    &59\\
$J=6$    &60    &57    &59    &57\\
$J=8$    &60    &60    &57    &61\\
$J=10$   &59    &59    &58    &59\\
$J=12$   &57    &58    &56    &61\\
$J=14$   &57    &59    &58    &56\\
$J=16$   &59    &58    &59    &57\\
\hline\hline
\end{tabular}}

\subtable[ABC-MART]{\begin{tabular}{l l l l l }
\hline \hline
  &$\nu = 0.04$ &$\nu=0.06$ &$\nu=0.08$ &$\nu=0.1$ \\
\hline
$J=4$    &45\   (22.4)   &41\   (28.1)   &44\   (22.8)   &42\   (28.8)\\
$J=6$    &47\   (21.7)   &48\   (15.8)   &47\   (20.3)   &45\   (21.1)\\
$J=8$    &49\   (18.3)   &49\   (18.3)   &49\   (14.0)   &46\   (24.6)\\
$J=10$   &55\   (6.8)   &54\   (8.5)   &50\   (13.8)   &49\   (16.9)\\
$J=12$    &56\   (1.8)   &55\   (5.2)   &50 \  (10.7)   &51\   (16.4)\\
$J=14$    &55\   (3.5)   &53\   (10.2)   &50\   (13.8)   &47\   (16.1)\\
$J=16$      &57\   (3.4)   &54\   (6.9)   &53\   (10.2)   &53\  (7.0)\\
\hline\hline
\end{tabular}}
}
\end{center}
\label{tab_Optdigits}
\end{table}

\begin{figure}[h]
\begin{center}
\mbox{\includegraphics[width=2.0in]{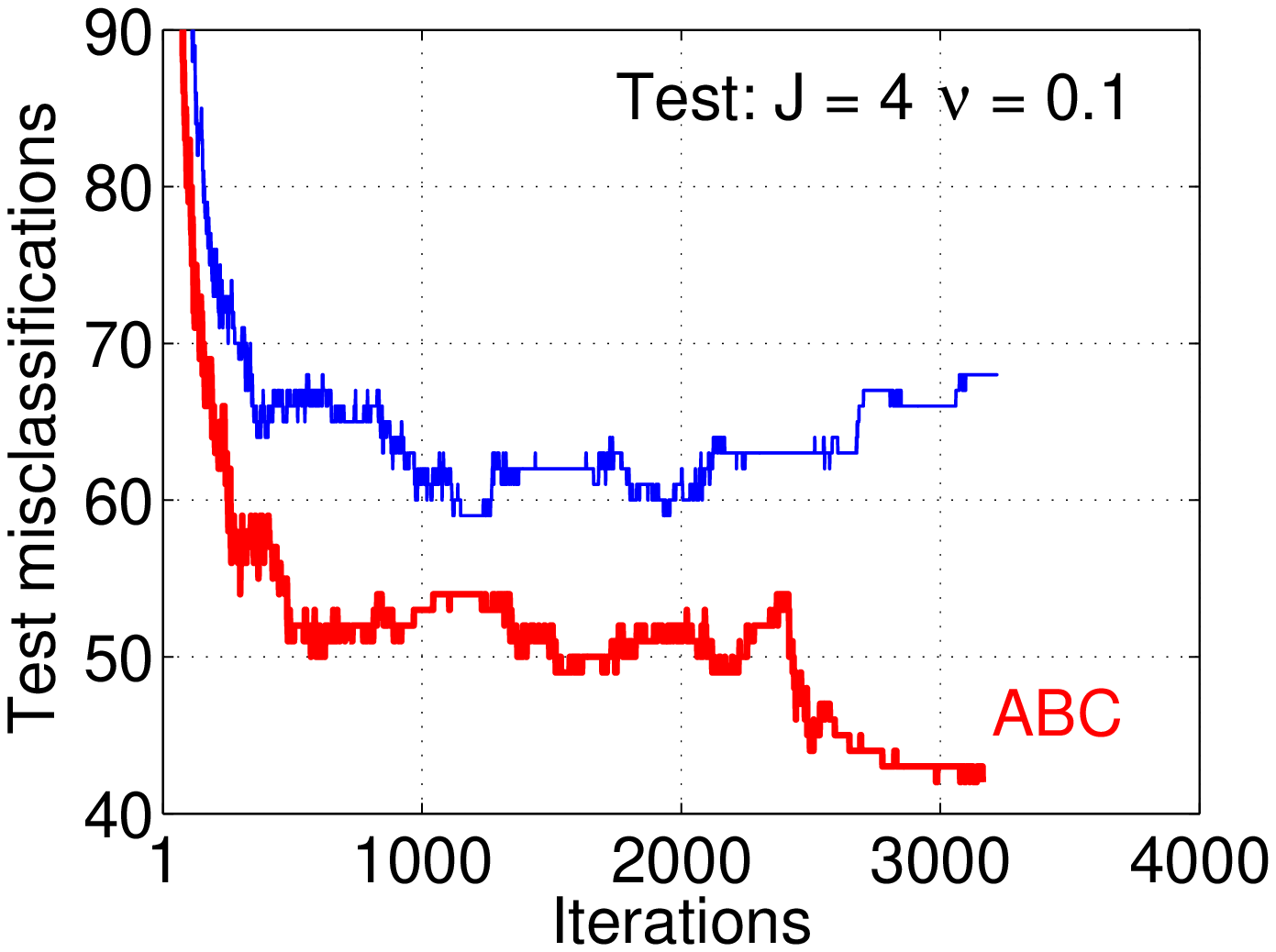}\hspace{0.1in}
{\includegraphics[width=2.0in]{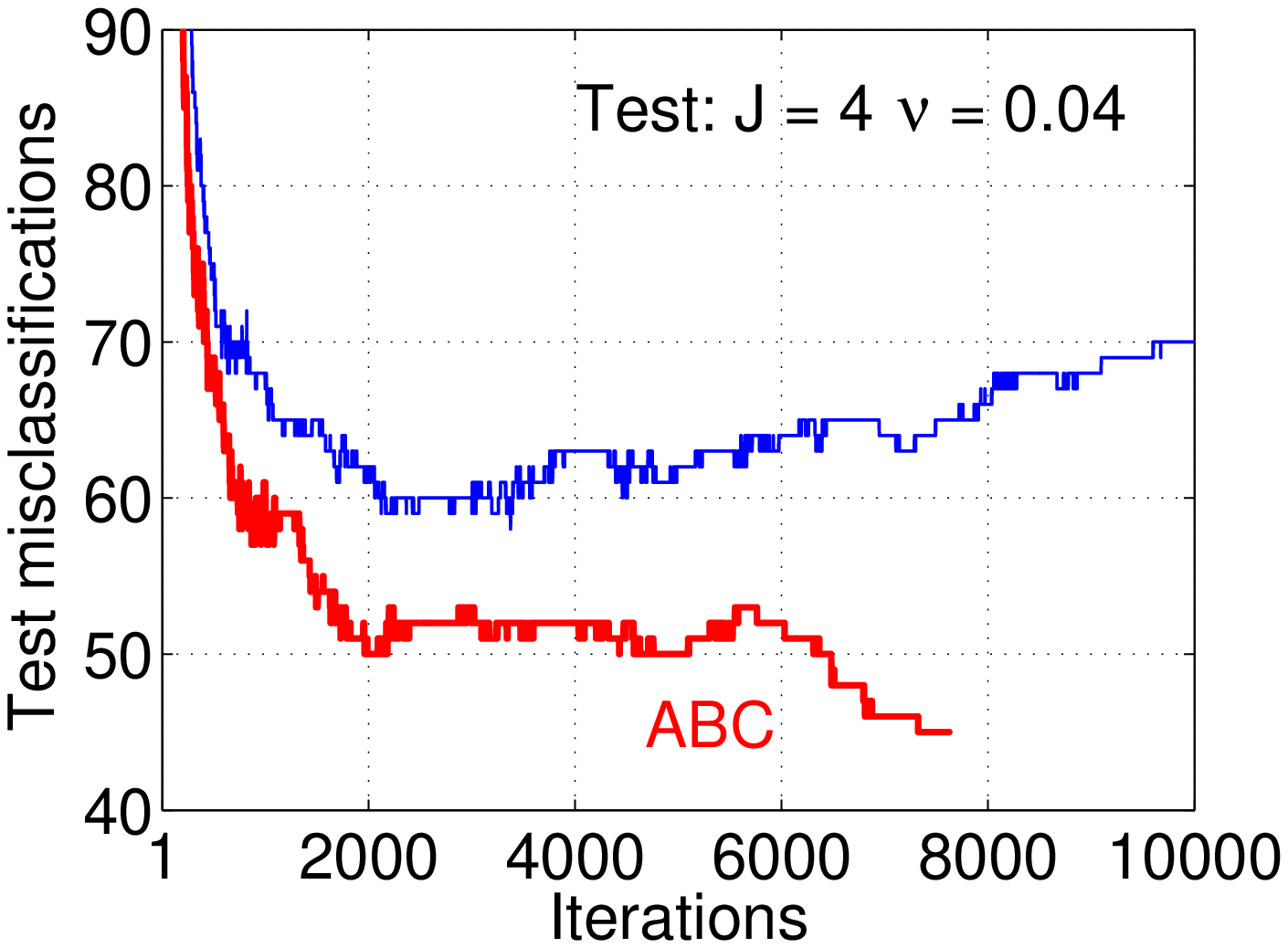}}}
\mbox{\includegraphics[width=2.0in]{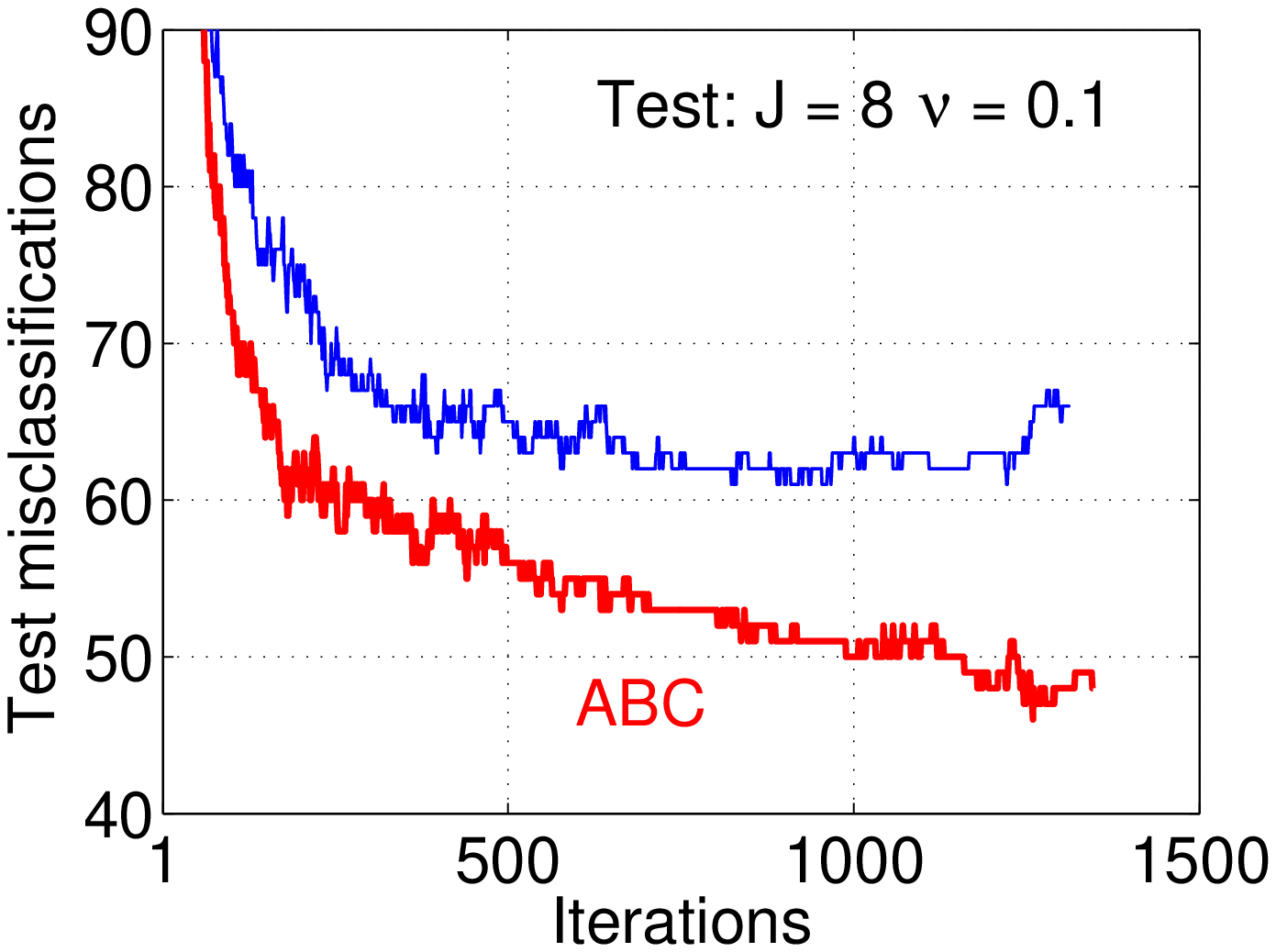}\hspace{0.1in}
{\includegraphics[width=2.0in]{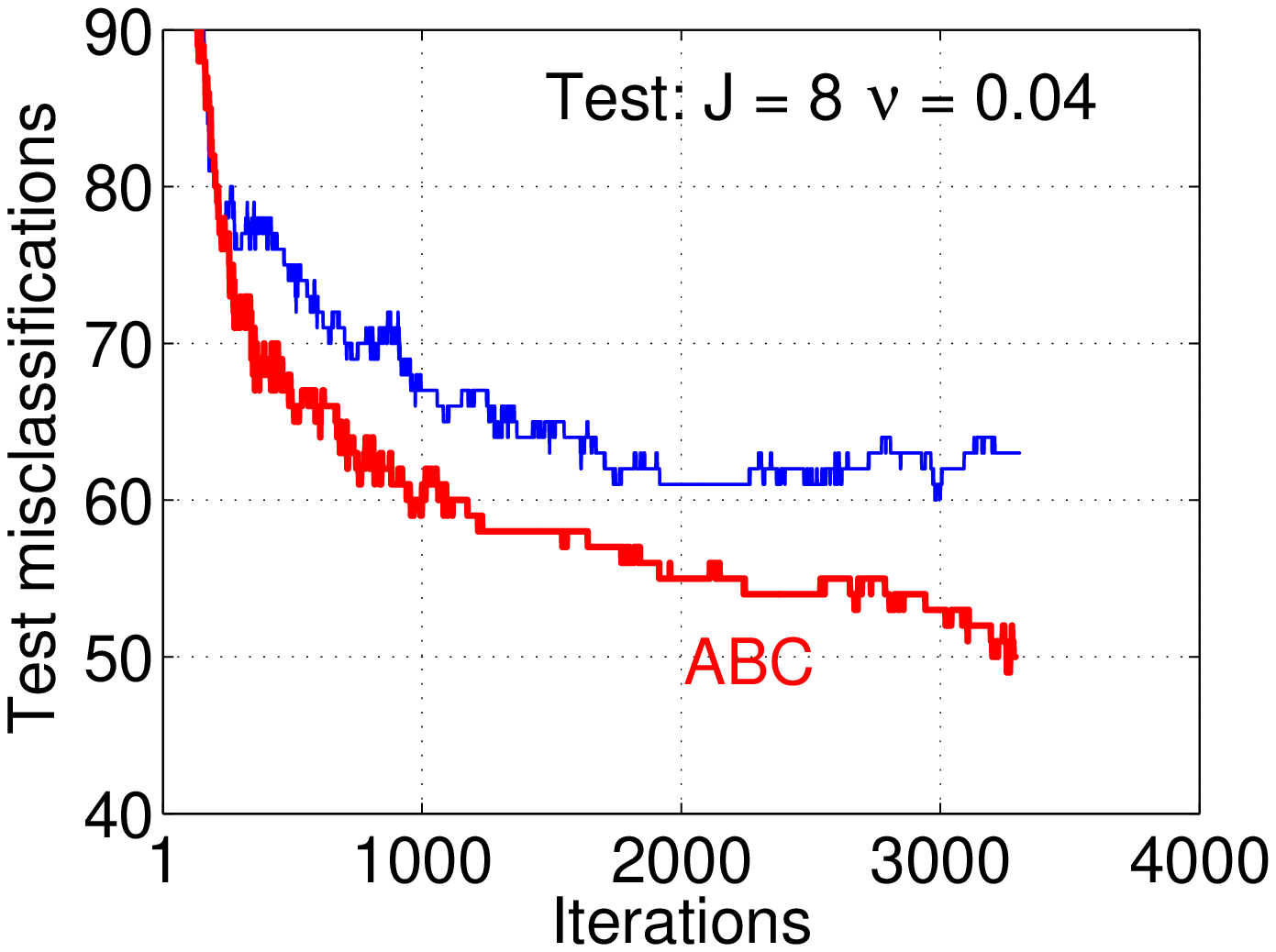}}}
\mbox{\includegraphics[width=2.0in]{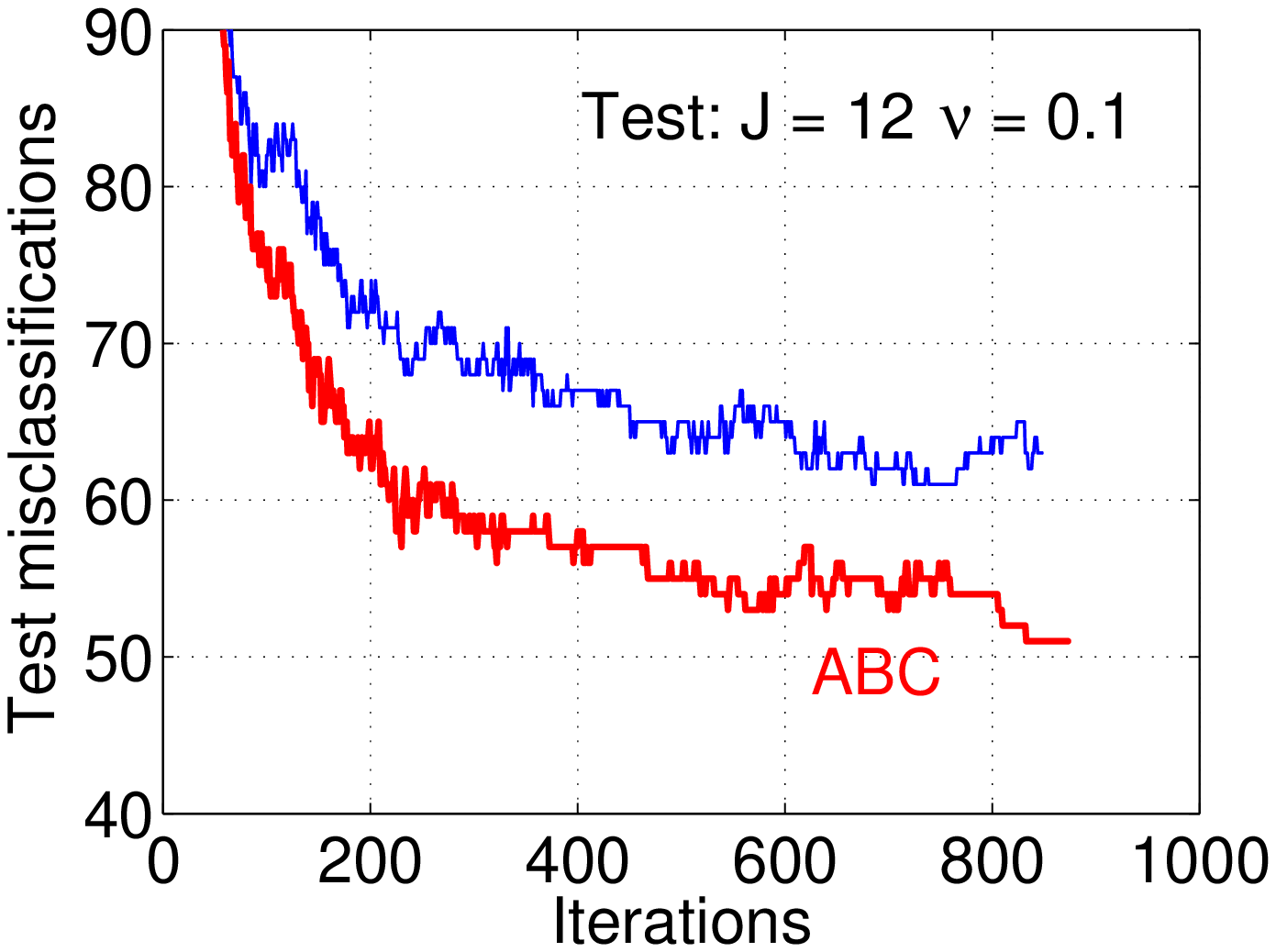}\hspace{0.1in}
{\includegraphics[width=2.0in]{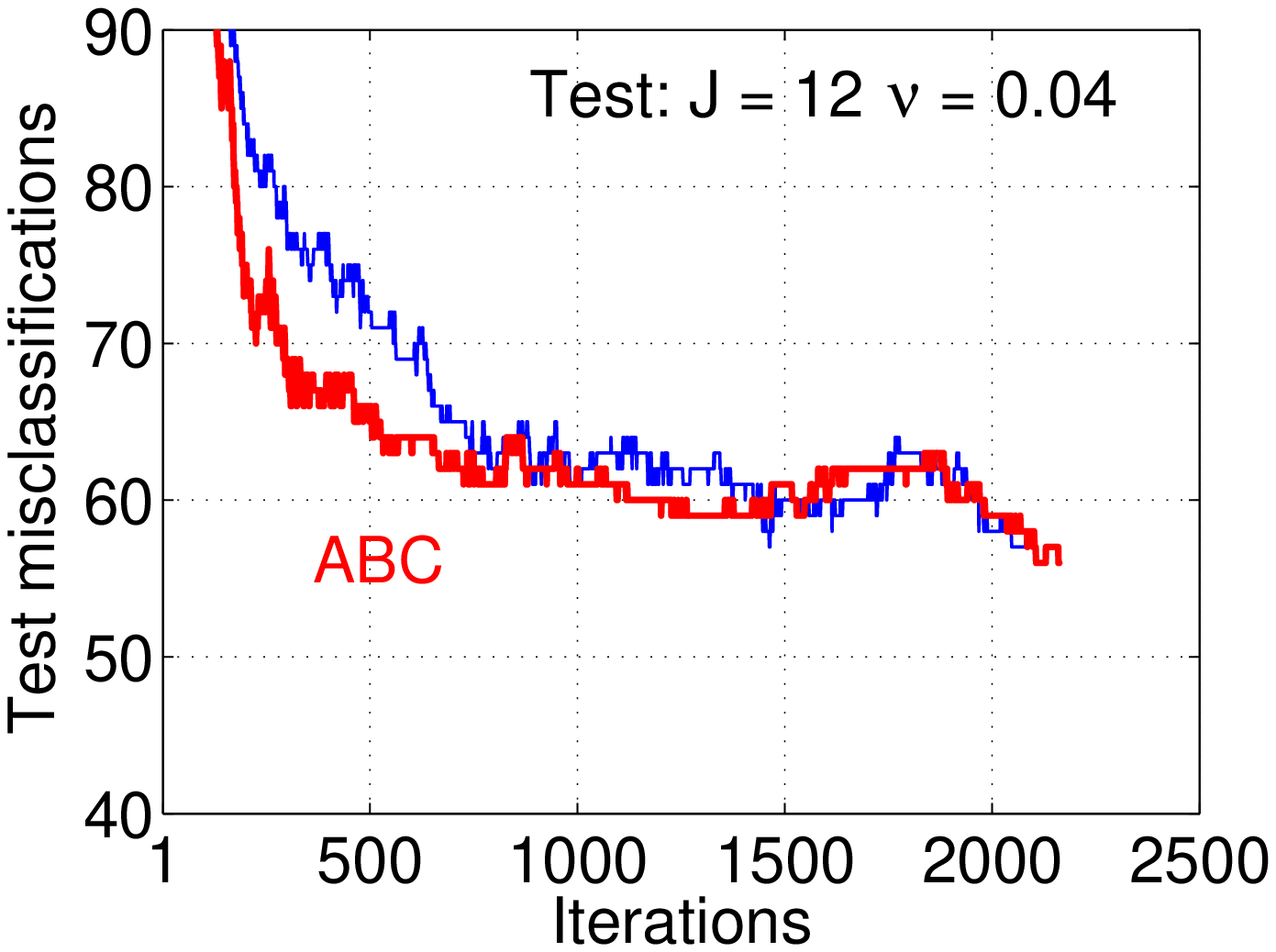}}}
\mbox{\includegraphics[width=2.0in]{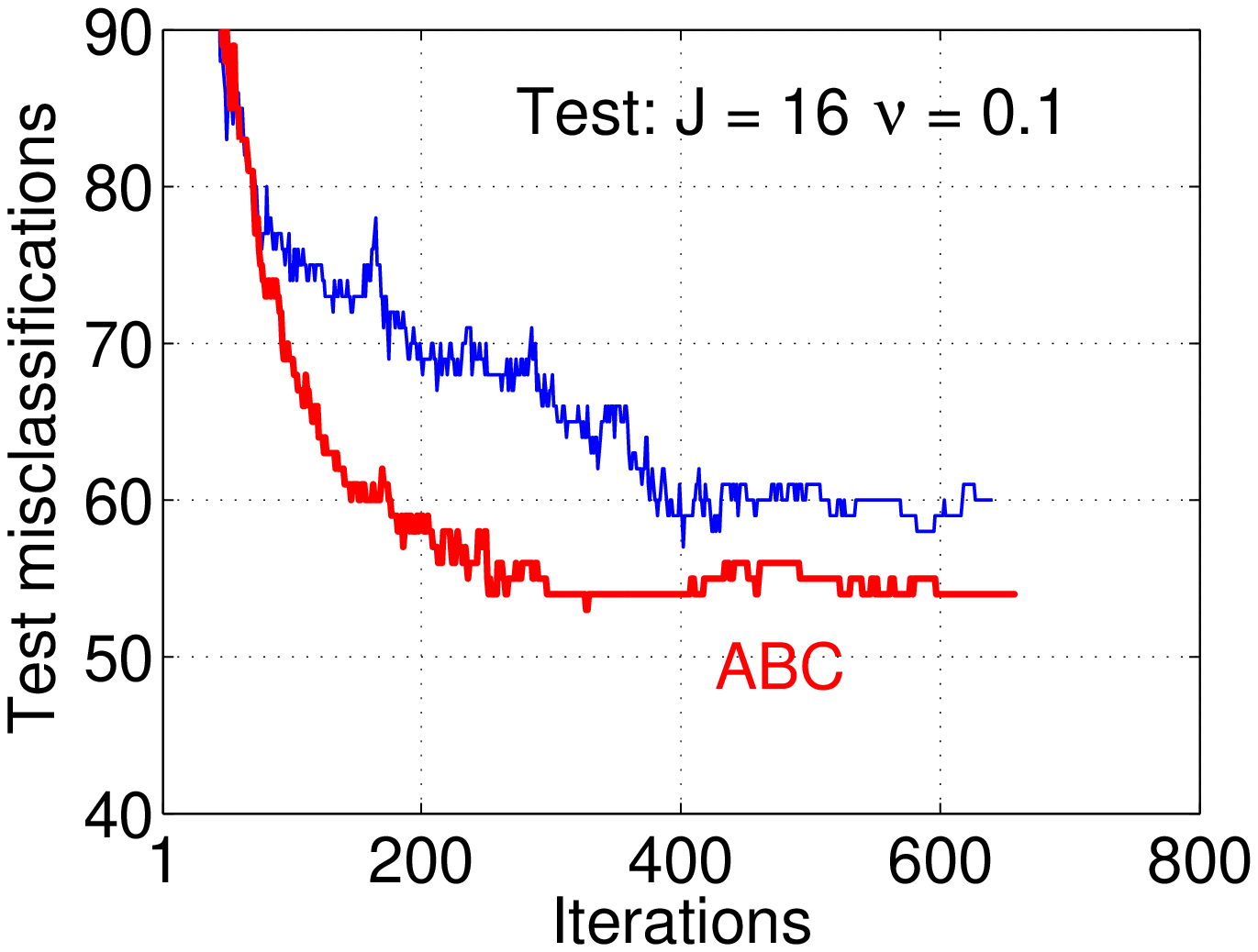}\hspace{0.1in}
{\includegraphics[width=2.0in]{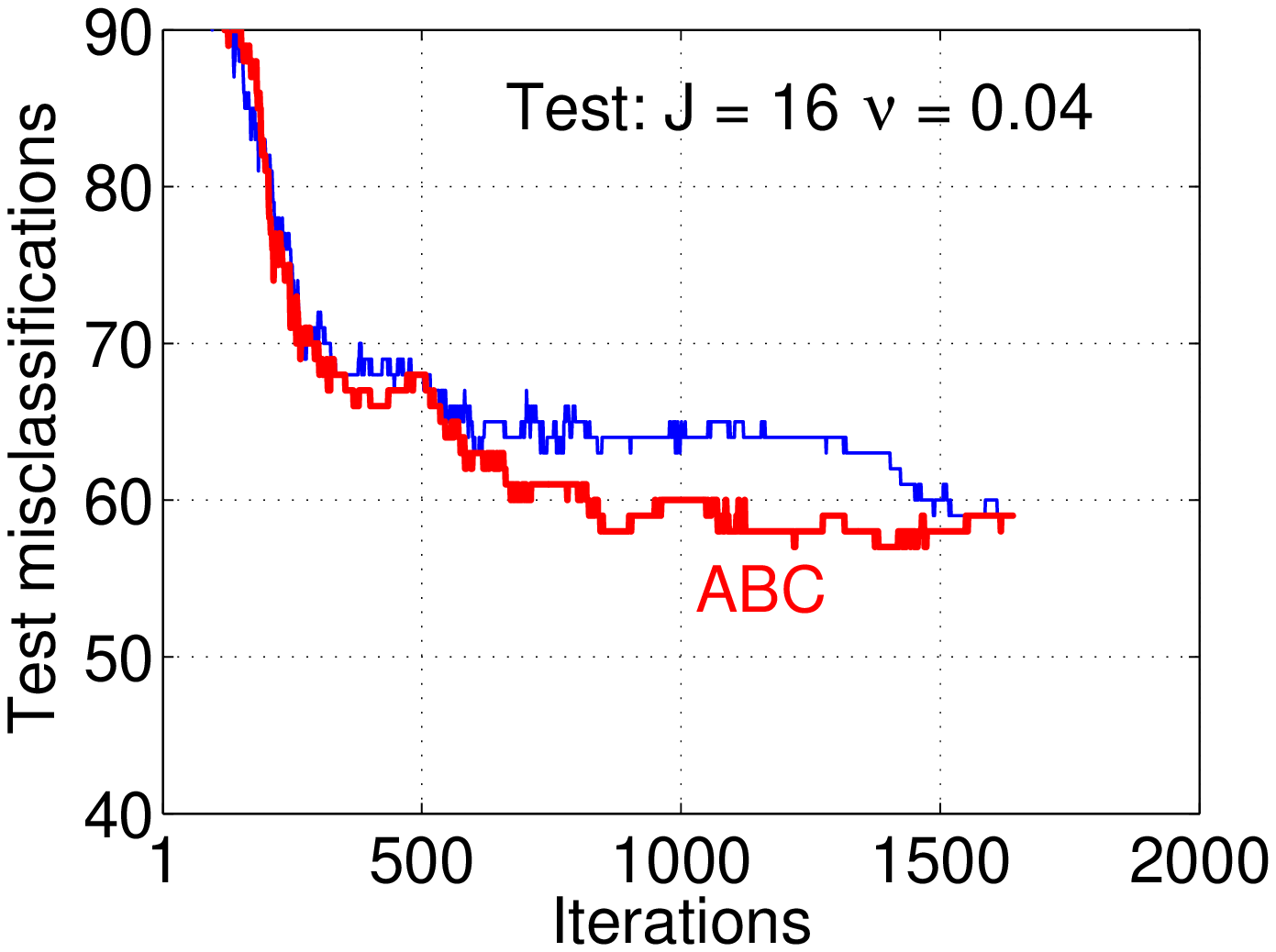}}}
\end{center}
\vspace{-0.2in}
\caption{\textbf{\em Optdigits}. The test mis-classification errors.}\label{fig_OptdigitsTest}
\end{figure}

\clearpage

\subsection{Experiments on the {\em Isolet} Data Set}\label{sec_Isolet}
\begin{table}[h]
\caption{\textbf{\em Isolet}. The test mis-classification errors.}
\begin{center}
{
\subtable[MART]{\begin{tabular}{l l l l l }
\hline \hline
  &$\nu = 0.04$ &$\nu=0.06$ &$\nu=0.08$ &$\nu=0.1$ \\
\hline
$J=4$    &84    &84    &85    &84\\
$J=6$    &84    &84    &85    &85\\
$J=8$    &87    &89    &86    &85\\
$J=10$   &88    &87    &84    &84\\
$J=12$   & 88   &91    &85    &86\\
$J=14$   &94    &93    &90    &91\\
$J=16$  & 93    &91    &90    &88\\
\hline\hline
\end{tabular}}

\subtable[ABC-MART]{\begin{tabular}{l l l l l }
\hline \hline
  &$\nu = 0.04$ &$\nu=0.06$ &$\nu=0.08$ &$\nu=0.1$ \\
\hline
$J=4$   &74\   (11.9)   &74\   (11.9)   &74\   (12.9)   &72\   (14.3)\\
$J=6$   &76\   ( 9.5)   &71\   (15.5)   &69\   (18.8)   &71\   (16.5)\\
$J=8$     &72\   (17.2)  &74\   (16.9)   &73\   (15.1)   &73\   (14.1)\\
$J=10$    &72\   (18.2)   &76\   (12.6)   &75\   (10.7)   &71\   (15.5)\\
$J=12$   &78\   (11.4)   &74\   (18.7)   &75\   (11.8)   &72\   (16.3)\\
$J=14$    &74\   (21.3)   &72\   (22.6)   &71\   (21.1)   &79\   (13.2)\\
$J=16$    &79\   (15.1)   &75\   (17.6)   &76\   (15.6)   &79\   (10.2)\\
\hline\hline
\end{tabular}}
}
\end{center}
\label{tab_Isolet}
\end{table}

\begin{figure}[h]
\begin{center}
\mbox{\includegraphics[width=2.0in]{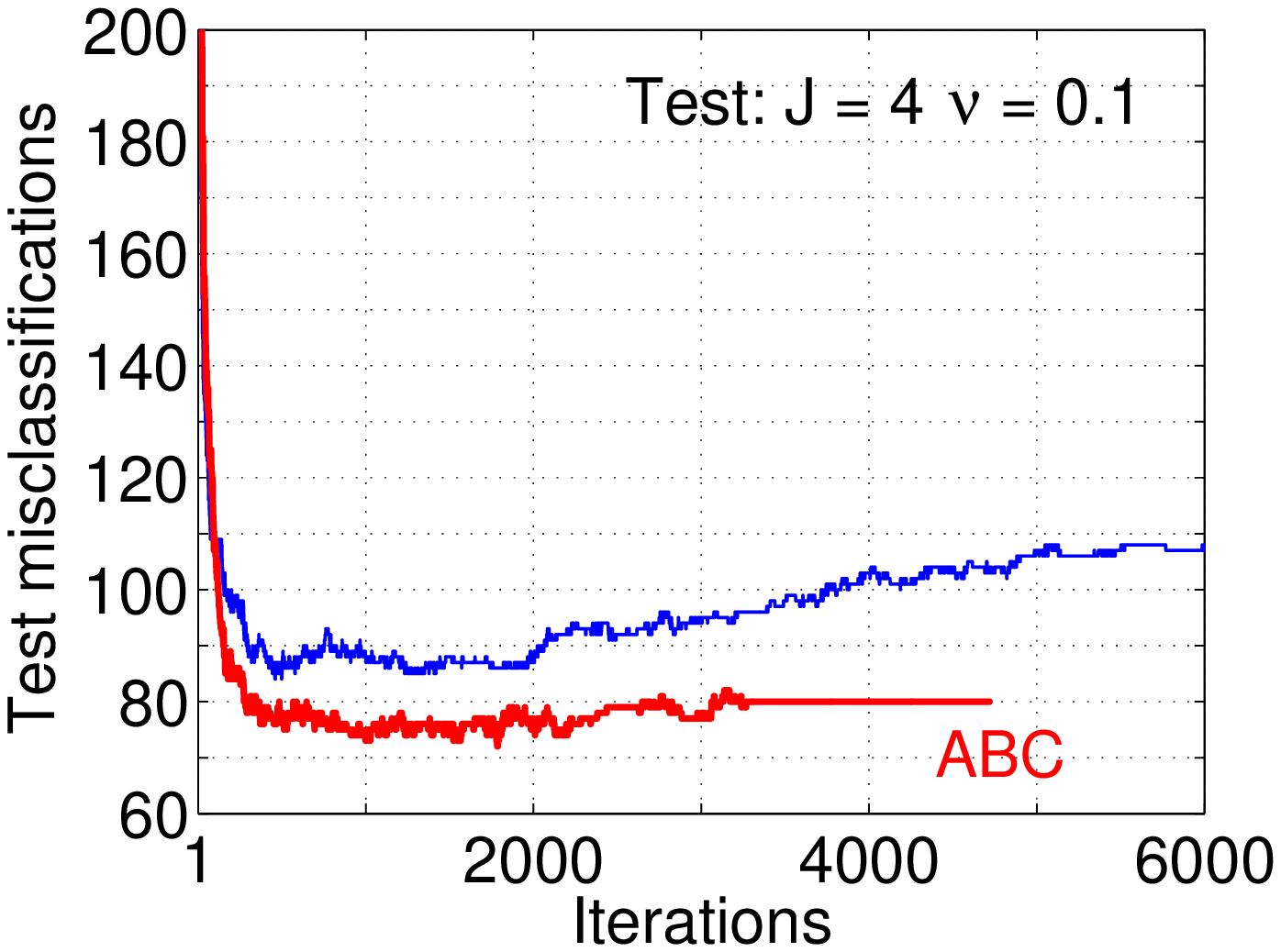}\hspace{0.1in}
{\includegraphics[width=2.0in]{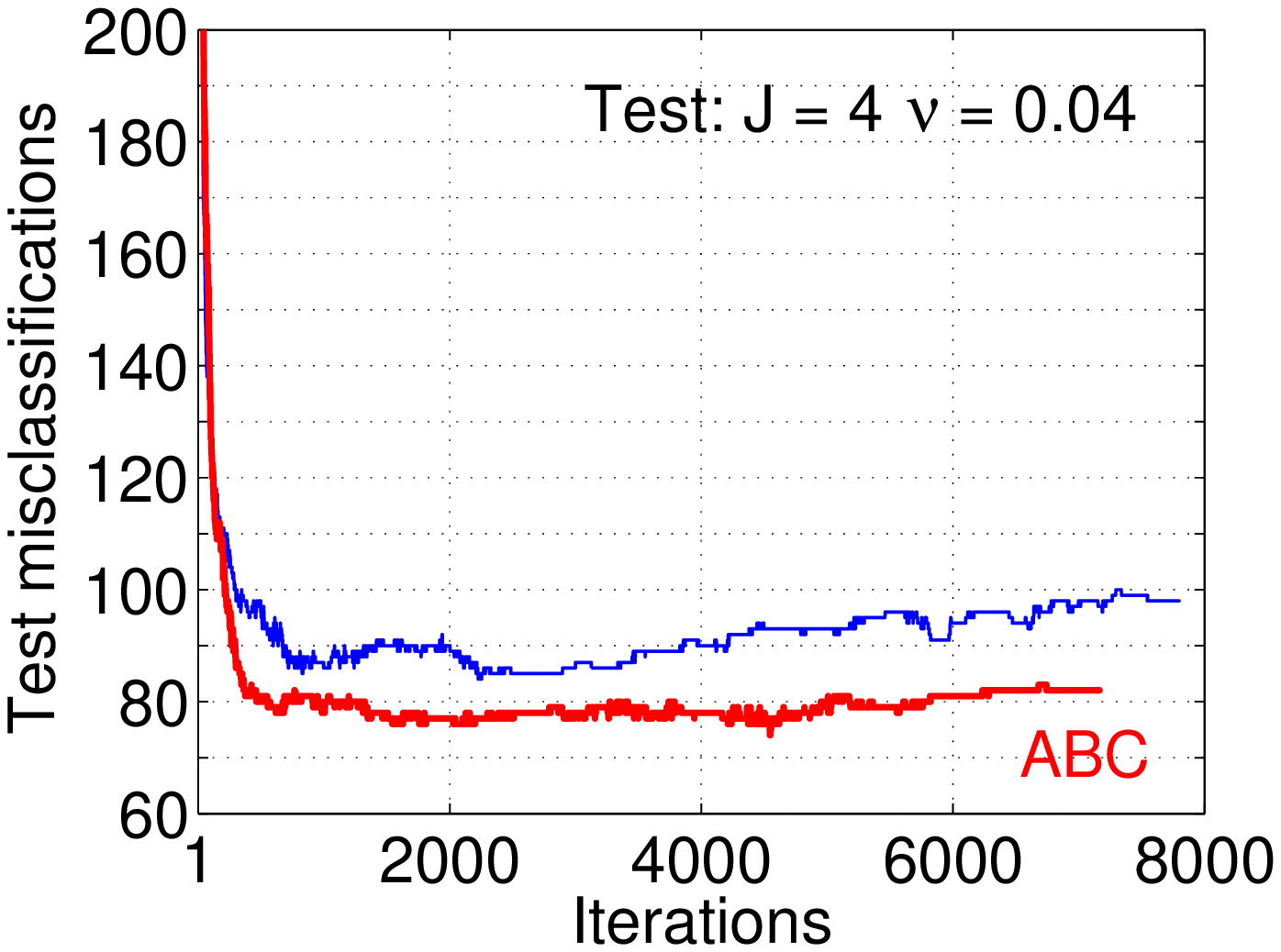}}}
\mbox{\includegraphics[width=2.0in]{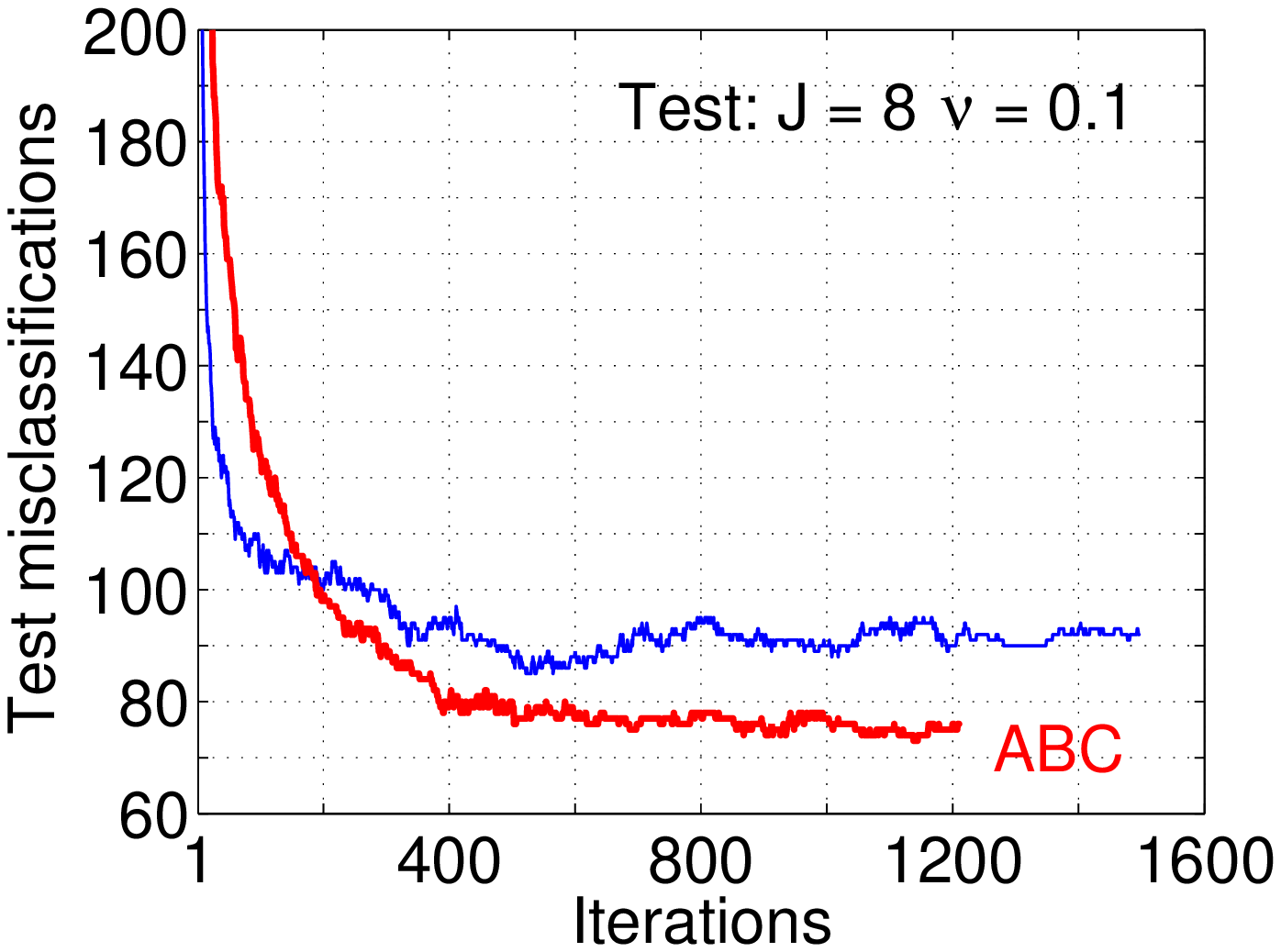}\hspace{0.1in}
{\includegraphics[width=2.0in]{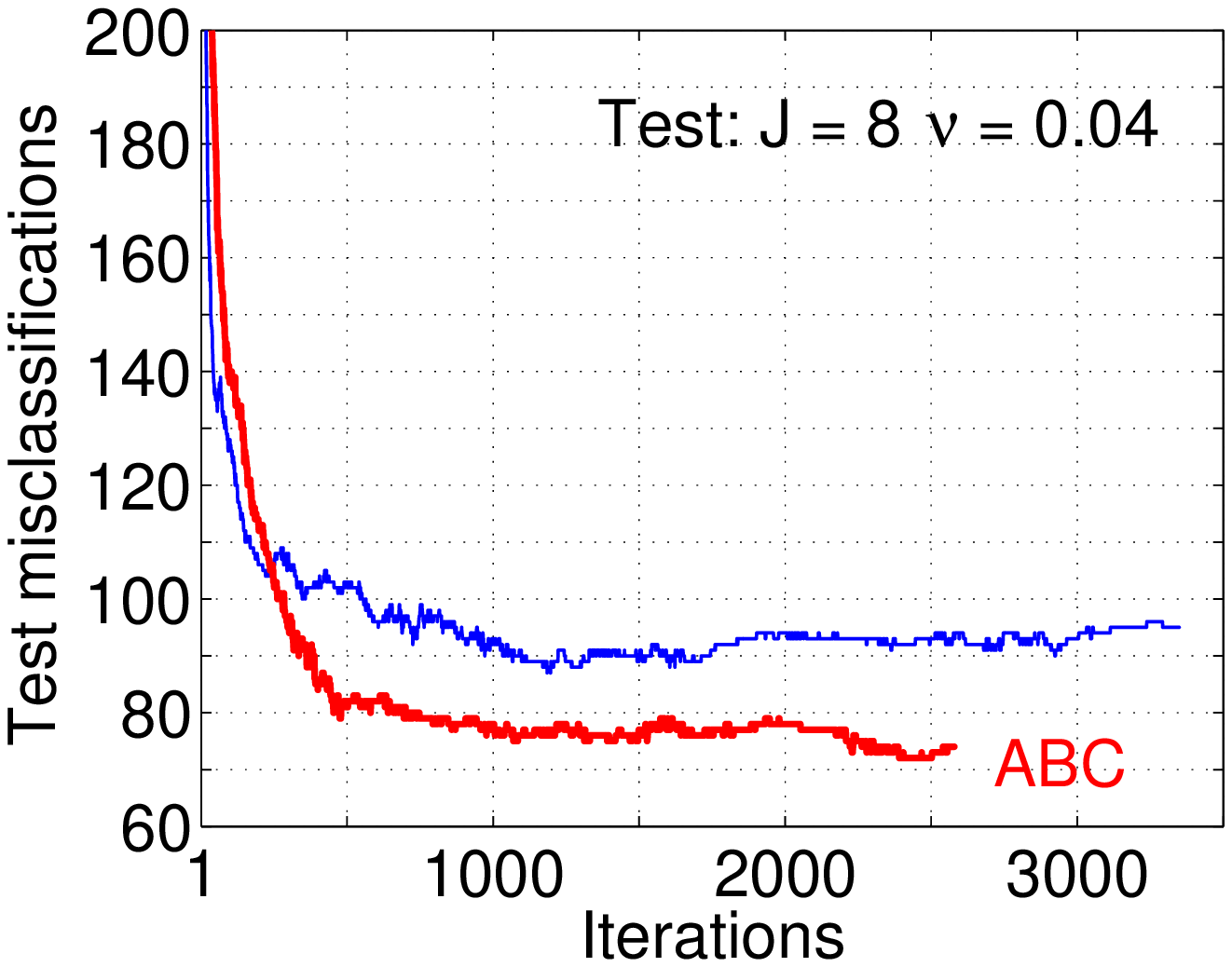}}}
\mbox{\includegraphics[width=2.0in]{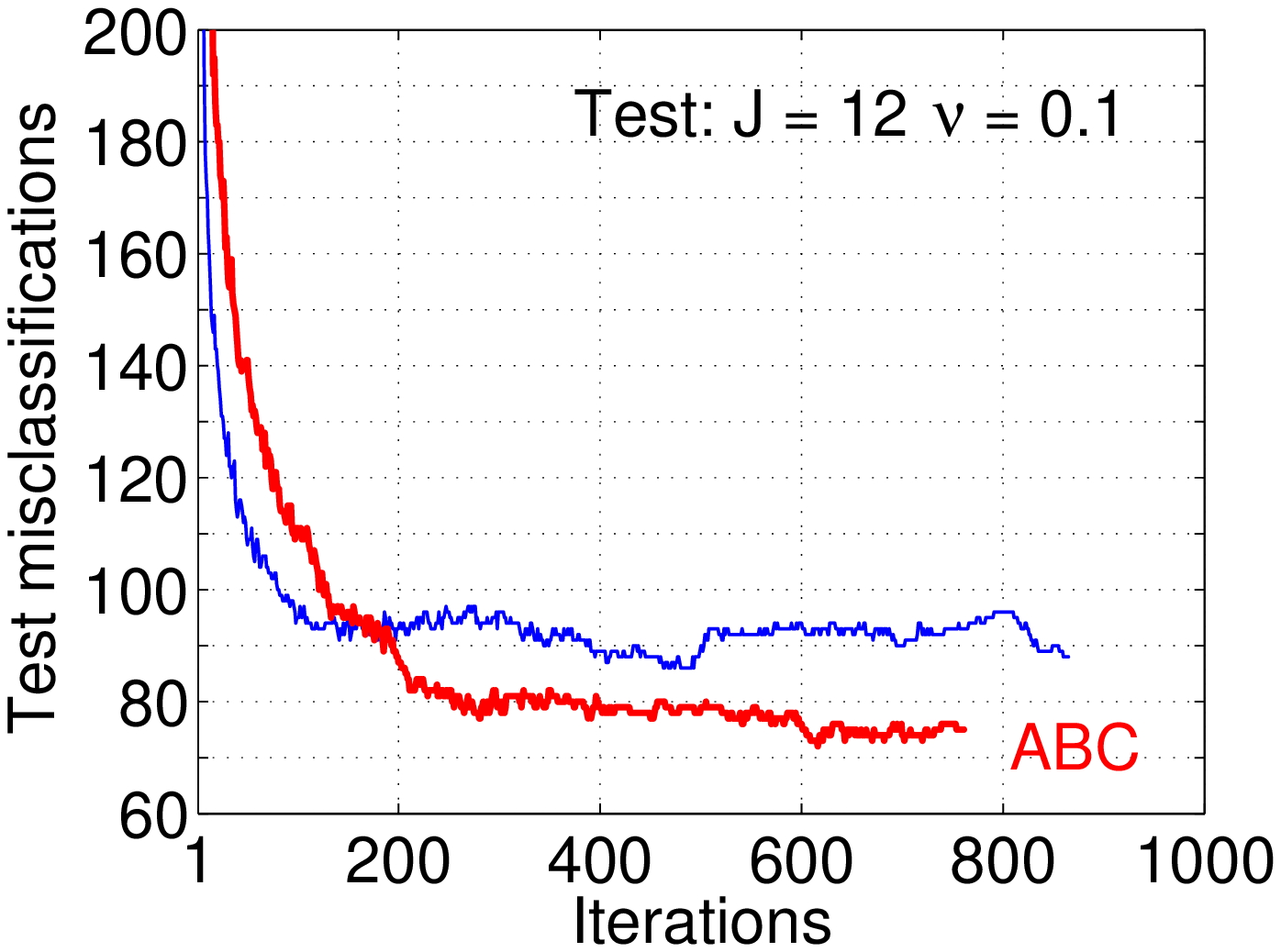}\hspace{0.1in}
{\includegraphics[width=2.0in]{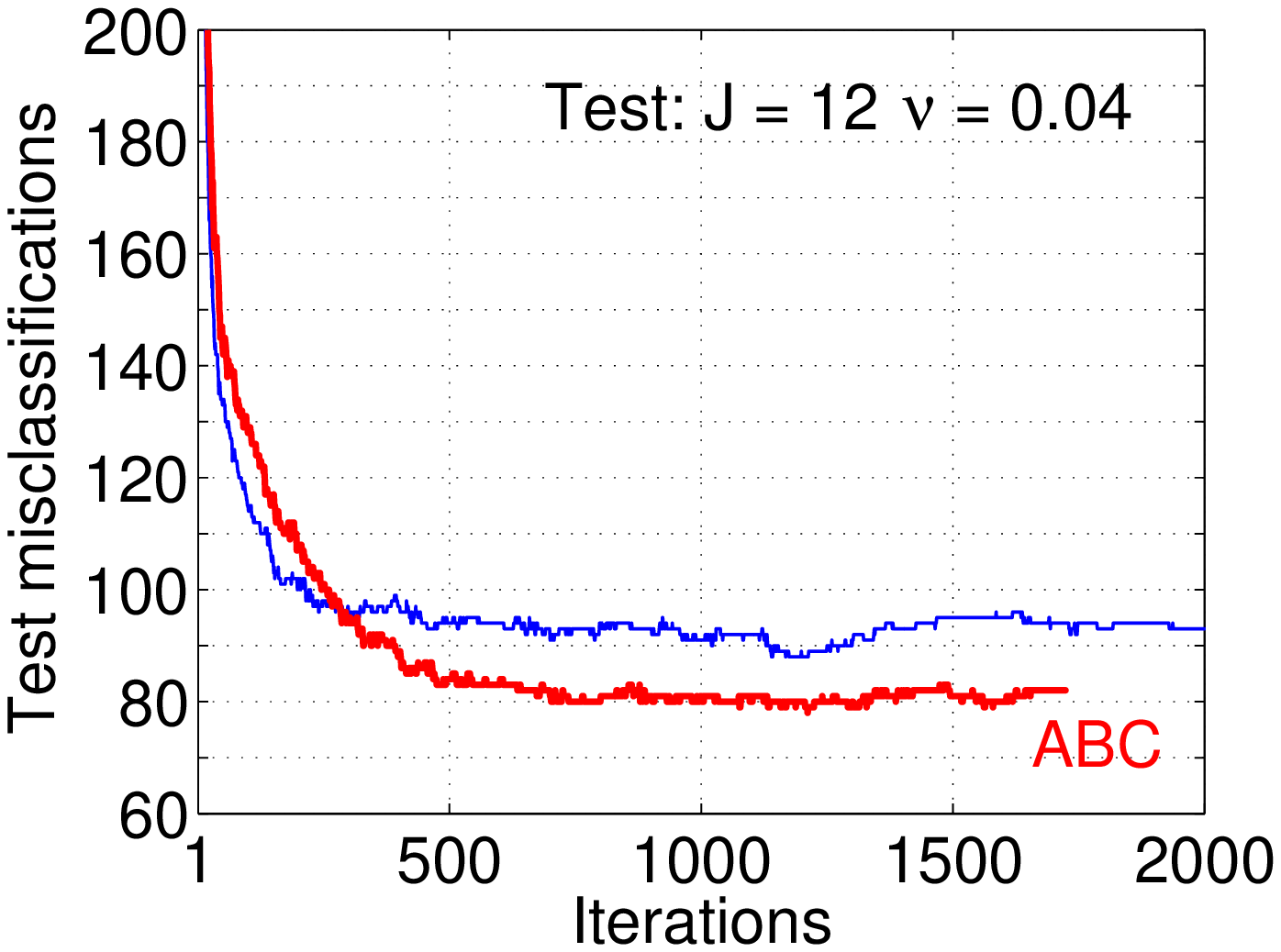}}}
\mbox{\includegraphics[width=2.0in]{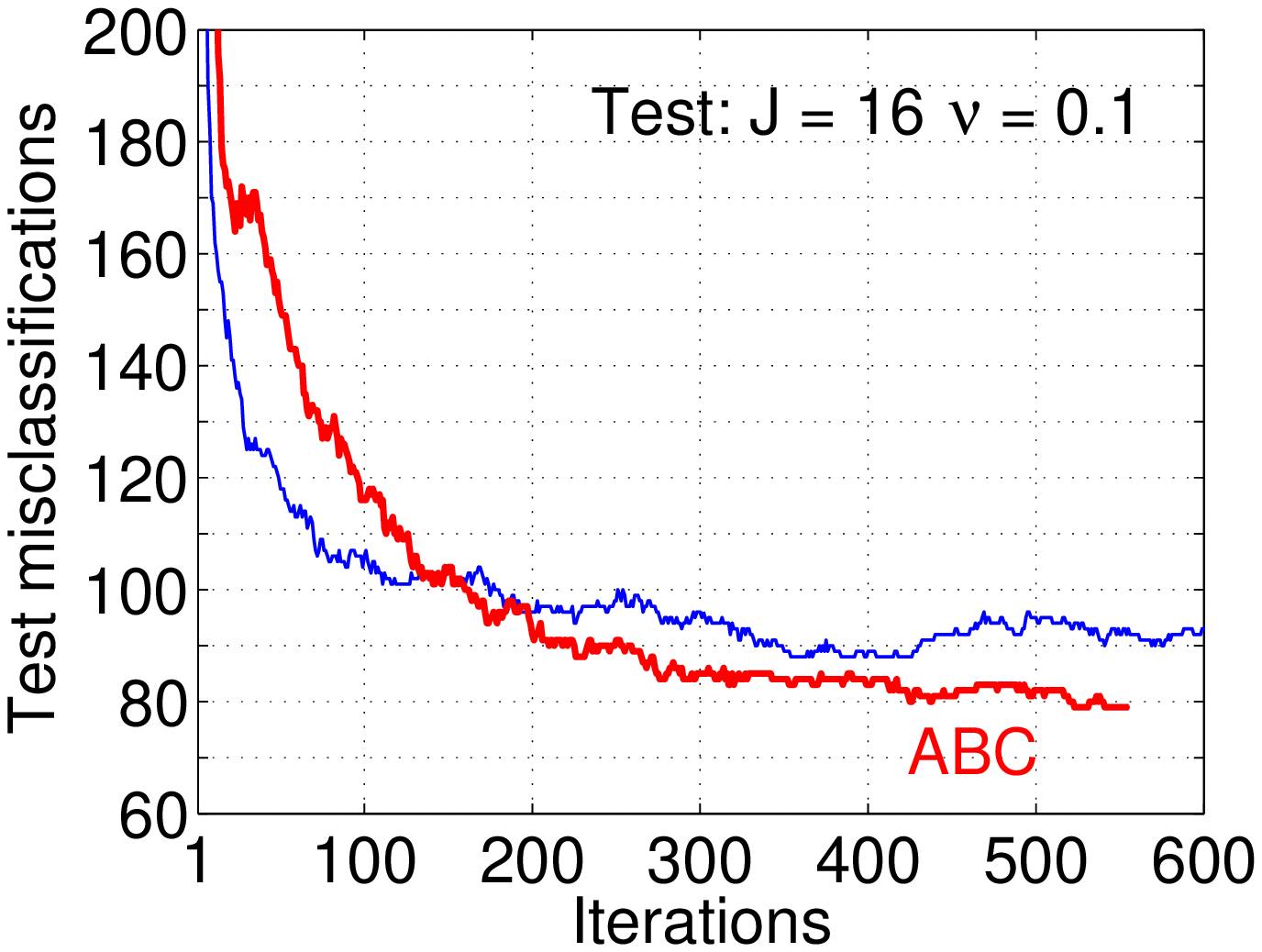}\hspace{0.1in}
{\includegraphics[width=2.0in]{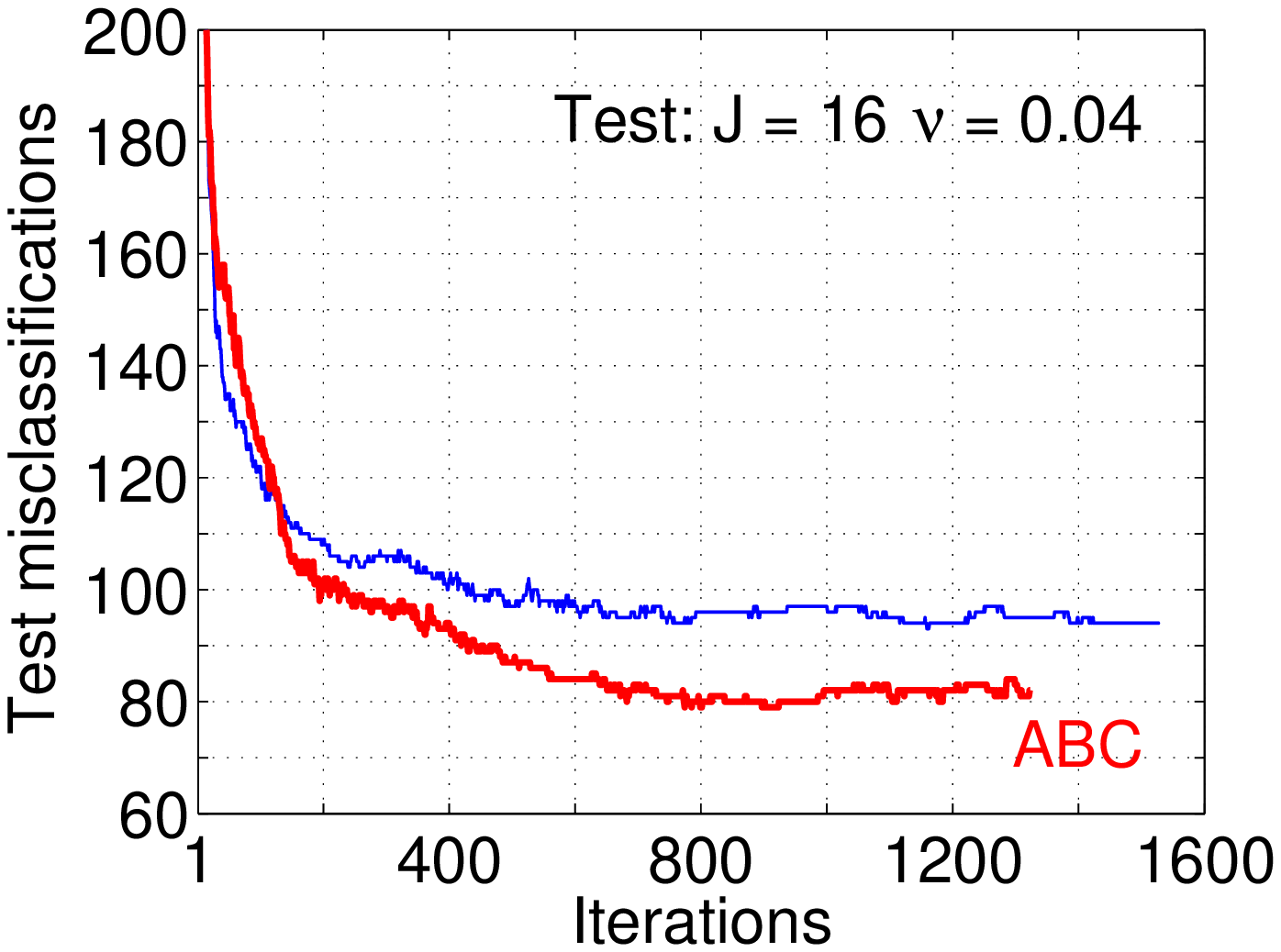}}}
\end{center}
\vspace{-0.2in}
\caption{\textbf{\em Isolet}. The test mis-classification errors.}\label{fig_IsoletTest}
\end{figure}

\clearpage

\section{Discussion}

Retrospectively, the ideas behind ABC-Boost and ABC-MART appear  simple and intuitive. Our experiments in Section \ref{sec_evaluations} have demonstrated the effectiveness of ABC-MART and its considerable improvements over MART.

The two key components of ABC-Boost are: (1) developing boosting algorithms by assuming one \textbf{base class}; (2) \textbf{adaptively} choosing the base class so that, at each boosting step, the ``worst'' class will be selected. We believe both components contribute critically to the good performance of ABC-MART.

Note that assuming the sum-to-zero constraint on the loss function and the base class is a ubiquitously adopted strategy\cite{Article:FHT_AS00,Article:Friedman_AS01,Article:Zhang_JMLR04}. Our contribution  in this part is the  different set of derivatives of the loss function (\ref{eqn_loss}), compared with the classical work\cite{Article:FHT_AS00,Article:Friedman_AS01}.

One may ask two questions. 1: Can we use the MART derivatives (\ref{eqn_MART_d1}) and (\ref{eqn_MART_d2}) and adaptively choose the base? 2: Can we use ABC-MART derivatives and  a fixed base? Neither will achieve a good performance.

\begin{figure}[h]
\begin{center}
\mbox{\includegraphics[width=3.0in]{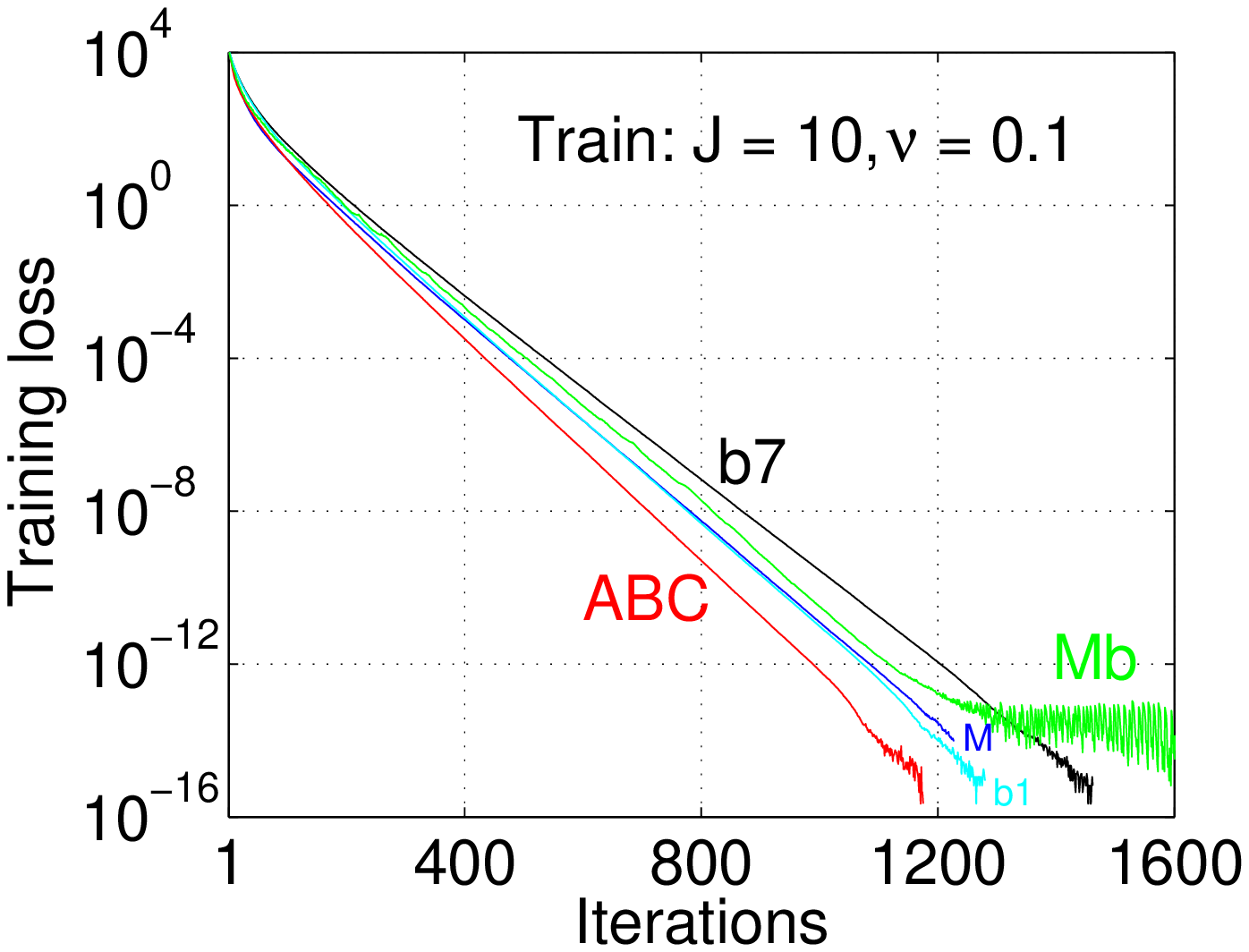}\hspace{0.1in}
{\includegraphics[width=3.0in]{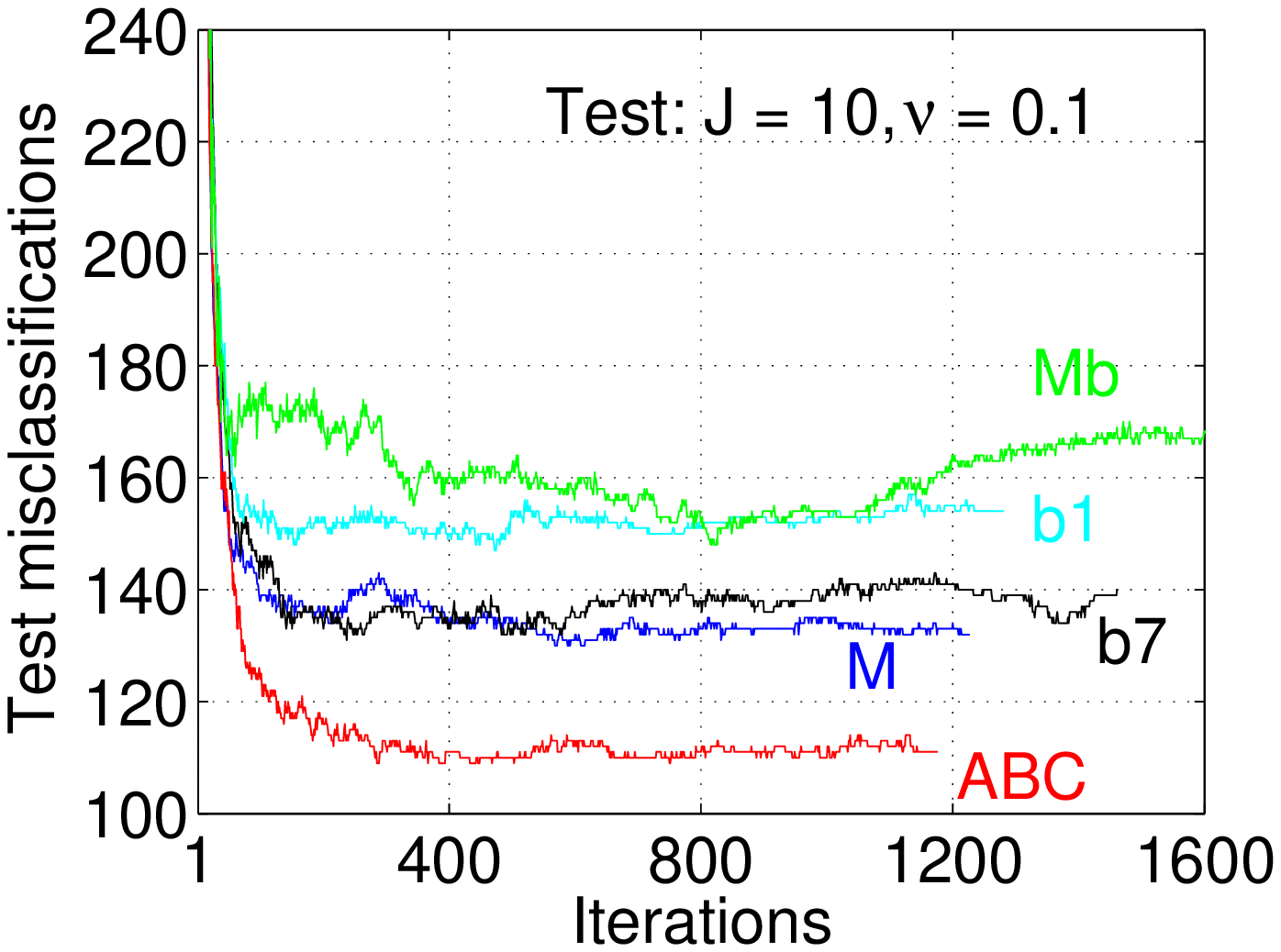}}}
\end{center}
\vspace{-0.2in}
\caption{\textbf{\em Pendigits}. The training loss and test mis-classification errors of three alternative boosting algorithms (labeled by ``Mb'', ``b1'', and ''b7''), together with the results of ABC-MART and MART (labeled by ``M''). }\label{fig_Pendigits_Comp}
\end{figure}

To demonstrate this point, we consider three alternative boosting algorithms and present their training and test results using the {\em Pendigits} data set ($K=10$).
\begin{enumerate}
\item {\em MART derivatives + adaptively choosing the worst base}.  It is the same as ABC-MART except it uses the  derivatives of MART. In Figure \ref{fig_Pendigits_Comp}, we label the corresponding curves by ``Mb.''
\item {\em ABC-MART derivatives + a fixed base class chosen according to MART training loss}. In the experiment with MART, we find  ``class 1'' exhibits the overall largest training loss. Thus, we fix  ``class 1'' as the base  and re-train  using the derivatives of ABC-MART. In Figure \ref{fig_Pendigits_Comp}, we label the corresponding curves by ``b1.''
\item  {\em ABC-MART derivatives + a fixed base class chosen according to MART test mis-classification errors}. In the experiment with MART, we find ``class 7'' exhibits the overall largest error. We fix  ``class 7'' as the base  and re-train using the derivatives of ABC-MART. In Figure \ref{fig_Pendigits_Comp}, we label the corresponding curves by ``b7.''
\end{enumerate}

Figure \ref{fig_Pendigits_Comp}  demonstrate that none of the alternative boosting algorithms could outperform ABC-MART.

\section{Conclusion}

We present the general idea of \textbf{ABC-Boost} and its concrete implementation named \textbf{ABC-MART}, for multi-class classification (with $K\geq 3$ classes). Two key components of ABC-Boost include: (1) By enforcing the (commonly used) constraint on the loss function, we can derive boosting algorithms for $K-1$ class using a \textbf{base class}; (2) We \textbf{adaptively} choose the current ``worst'' class as the base class, at each boosting step. Both components are critical.  Our experiments on a fairly large data set and five small data sets demonstrate that ABC-MART could considerably improves the original MART algorithm, which has already been highly successful in large-scale industry applications.

\section*{Acknowledgement}

The author thanks Tong Zhang for providing valuable comments to this manuscript. In late June 2008, the author visited Stanford University and Google in Mountain View. During that visit, the author had the opportunity of presenting to Professor Friedman and Professor Hastie the algorithm and some experiment results of ABC-MART; and the author highly appreciates their comments and encouragement. Also, the author would like to express his gratitude to Phil Long (Google) and Cun-Hui Zhang for the helpful discussions of this work, and to Rich Caruana who suggested the author to test the algorithm on the {\em Covertype} data set.

\end{document}